\documentclass[11pt,a4paper]{article}
\usepackage[T1]{fontenc}

\usepackage[a4paper,top=2cm,bottom=2cm,left=3cm,right=3cm,marginparwidth=1.75cm]{geometry}
 
\usepackage{amsmath}
\usepackage{graphicx}
\usepackage{hyperref}
\usepackage{subcaption}
\usepackage{longtable}
\usepackage{booktabs}
\usepackage{xcolor}
\usepackage[export]{adjustbox}
\usepackage{float}
\usepackage{comment}
\usepackage{cleveref}
\crefname{figure}{Figure}{Figures}
\crefdefaultlabelformat{#2#1#3}

\usepackage[mathlines]{lineno} 

\usepackage[affil-it]{authblk}

\usepackage{algorithm} 
\usepackage{algpseudocode}
\usepackage{amssymb}

\makeatletter
\renewcommand{\section}{\@startsection {section}{1}{\z@}%
              {24pt}{12pt} {\large\scshape\bfseries}}

\renewcommand{\subsection}{\@startsection {subsection}{2}{\z@}%
             {12pt}{12pt}  {\itshape\bfseries}}

\usepackage{natbib}

\title{\bfseries \normalsize Learning to Learn the Macroscopic Fundamental Diagram using Physics-Informed and Model Agnostic Machine Learning}

\author[1]{Amalie Roark}
\author[2]{Serio Agriesti\footnote{corresponding author: samaa@dtu.dk}}
\author[2]{Francisco Camara Pereira}
\author[2]{Guido Cantelmo}

\affil[1]{Technical University of Denmark, 2800 Kgs. Lyngby, Denmark}
\affil[2]{Department of Technology, Management and Economics, Division of Transportation Science, Technical University of Denmark, 2800 Kgs. Lyngby, Denmark}

\date{\vspace{-5ex}}

\begin{document}
\maketitle

\section*{Abstract}\small

The Macroscopic Fundamental Diagram is a popular tool used to describe traffic dynamics in an aggregated way, with applications ranging from traffic control to incident analysis. However, estimating the MFD for a given network requires large numbers of loop detectors, which is not always available in practice. This article proposes a framework to alleviate the data scarcity challenge harnessing
Meta-Learning, a subcategory of Machine Learning that trains models to understand and adapt to new tasks on their own. We use Meta-Learning to identify and exploit transferable patterns from data-rich cities to cities where not enough data is available to estimate the MFD. The developed model is trained and tested by leveraging data from multiple cities and exploiting it to model the MFD of other cities with different shares of detectors and topological structures. The proposed Meta-Learning framework is applied to an ad-hoc Multi-Task Physics-Informed Neural Network, specifically designed to estimate the MFD. Results show an average MAE improvement in flow prediction of around 50\% across cities (depending on the subset of loop detectors tested). The Meta-Learning framework thus successfully generalizes across diverse urban settings and improves performance on cities with limited data, demonstrating the potential of using Meta-Learning when a limited number of detectors is available. We directly test this assumption by applying the Meta-Learning outputs to unseen cities to simulate a real-life application scenario and the wide applicability of the proposed methodology. Finally, the proposed framework is validated against traditional Transfer Learning approaches and tested with FitFun, a model for FD estimation from the literature, to prove its transferability.

\textbf{Keywords}: MFD, Estimation, Multi-Task Physics-Informed Neural Network, Meta Learning, Transfer Learning, data scarcity, UTD19, Model-Agnostic Machine Learning, MAML.

\section{Introduction}

Traffic in cities is related to multiple topical issues and many negative impacts such as air pollution \citep{batista2022exploring, peng2023airpollution}, declining health \citep{levy2010publichealth} or lower economic output \citep{weisbrod2003congeffect}, phenomena strictly intertwined with the societal functioning of the city itself and thus affecting almost the whole share of the population. A significant amount of research has been done to understand the nature and dynamics of urban congestion and how it propagates in the attempt to reduce its impact on the economy, health, and the environment. One key element is the Fundamental Diagram (FD), which defines the relation between traffic flow, density and speed on a single link. The FD has seen widespread applications both in theory and in practice, as it allows to identify the state transition between free flow and congestion. The Macroscopic Fundamental Diagram (MFD) extends these principles to a network level, describing the relationship between the aggregated flow and density across an urban area \citep{daganzo2007urban}. This concept was first introduced to capture the collective traffic dynamics at a macroscopic scale and has since become a powerful tool for analyzing and managing urban traffic networks \citep{geroliminis2008existence}. Indeed, multiple applications of the MFD can be found in literature, a non-exhaustive list includes route guidance \citep{Hajiahmadi2013route, Yildirimoglu2015routeMFD, zhong2018gating, chen2024route}, gating \citep{Ekbatani2015gating, Guo2020gating, sirmatel2021gating}, assessment of automated vehicles' coordinating strategies \citep{Zhang2022MFDAV} and behavioural analyses \citep{batista2025activity}. A wider literature review that covers the many applications of the MFD is reported in \citep{Zhang2020review}
\\
Nevertheless, deriving the MFD from empirical data is a challenging exercise since (i) it requires data from a large number of loop detectors, (ii) detectors should cover the entirety of the network, and (iii) the data coming from the detectors is prone to various types of biases and inaccuracies, making its cleaning and processing challenging \citep{aghamohammadi2022parameter}. The problem is not trivial, as proved by \cite{Leclerq2014MFDestimation}, who conclude that the only way to estimate MFD without bias is to have complete information about vehicle trajectories, an unrealistic assumption even in the age of Big Data.

Several analytical and simulation-based methods have been proposed to address this problem and draw representative MFDs. The Method of Cuts \citep{daganzo2008analytical}, for example, allows estimating upper bounds for the MFD parameters. \cite{laval2015stochastic} and \cite{aghamohammadi2022parameter} extended the method to consider bias in the data and stochasticity. \cite{Tilg2023sioux} improve it to account for different spatial demand patterns and irregular network topologies. \cite{Shim2019bifurcation} focus instead on identifying the main sources of bifurcation trends emerging in the high density branch of the MFD, which causes its shape to continuously change over time. \cite{leclercq2025mfdvario} design methods to estimate less biased MFDs in areas of the network that are not covered by enough detectors while \cite{saeednabesg2021kalaman} apply an extended Kalaman filter algorithm to estimate the MFD in real-time. \cite{Saffari2020criticallinks} use probe vehicle data and apply Principal Component Analysis (PCA) to identify critical links, through whose recorded data they reconstruct the MFD. Probe data has also been coupled with loop detectors (LDs) for MFD estimation in \cite{Ambuhl2016FD&LD,Saffari2022Datafusion}. In \cite{bramich2023fitfun}, authors point out that the distribution of measurements' averages in a MFD is non-Gaussian and non symmetric, and proposes FitFun, a non-parametric tool based on Generalised Additive Models for Location, Scale, and Shape (GAMLSS) to fit the FD. The authors test FitFun on more than 200 different models, with different assumptions on the shape of the MFD as well as different assumptions on noise.

Data is not the only factor affecting the MFD estimation. \cite{Ambuhl2018resampling} assess the effects of network inhomogeneity and design a resampling methodology to estimate the critical density. \cite{Elbukhari2024PCA} use PCA to partition the network in more homogeneous zones, to maximize MFD quality. Clustering is also used for MFD partitioning, with some authors arguing that Gaussian Mixture Models is the most suited technique to ensure that basic MFD assumptions are met during the cluster creation \citep{batista2021partitioning}. \cite{Yuan2023hyst} study instead the relationship between departure times, user equilibrium and the phenomena of MFD hysteresis and missing congested branch. \cite{batista2025activity} further analyse the impact of activities and trip purpose on the MFD, showing the impact of heterogenous travel preferences on the user equilibrium in terms congestion and departure time. 

Another stream of work found in the MFD literature is the one of Machine Learning (ML), which generally tries to harness patterns within the data to generalize a MFD shape. In the work of \cite{yuan2021macroscopic}, for example, a physics-regularized Gaussian process (GP) model for macroscopic traffic flow modeling is proposed. This approach integrates traffic flow physics into the GP framework to improve prediction accuracy. Similarly, \cite{Shi2021PIML} present a physics-informed deep learning approach for estimating the MFD which incorporates traffic flow physics. \cite{Jin2023MLcities} compare multiple traditional ML techniques in their ability to reconstruct MFDs across different cities, while \cite{alayasreih2025ml} compare ML techniques in their ability to cluster traffic states in the MFDs based on traffic homogeneity and spatial connectivity. \cite{Lin2024lstm} use LSTM and GA algorithms to forecast MFDs on future short term time intervals. A physics-informed computational graph method is developed in \cite{Zhang2024PIML} to estimate free flow speed and critical density for the MFD. \cite{Jin2024preprint} use Graph Attention Networks and Gated Recurrent Units to enhance the precision of link-level traffic speed estimations based on MFD. \cite{bansal2023utd19} use causal statistical modeling and bayesian Machine Learning to empirically estimate a bias-free macroscopic fundamental relationship, also using the UTD19 dataset. Other authors proposed using Active Learning to sample a representative set of trips (either using simulation or using GPS data), to be used for MFD estimation \citep{batista2022gaussian}.  

Relevant for this work, an emerging stream of the ML literature focuses instead on ML approaches involving Meta-Learning to address the lack of data issue.  Meta-Learning - often referred to as ”learning to learn” - is a paradigm aimed at training ML models that can generalize across multiple tasks or datasets. It is widely applied on other fields and subjects, such as pattern recognition \citep{luo2022pattern}, drug design \citep{olier2018drugdesign} or cybersecurity \citep{Yang2023cyberspace}, for example. A comprehensive review of the methods is provided in \cite{Vettoruzzo2024mlreview}. Meta-Learning is also applied to traffic studies (e.g., \cite{Sao2023metacitta, Yang2024Koopman, Sun2024crosslight}), but, to the best of the authors' knowledge, only \cite{Sun2023meta} and \cite{Liu2023fewshot} (and now this work) harness it to tackle on the specific challenge of MFD estimations. In \cite{Sun2023meta}, a framework that leverages Meta-Learning to improve traffic prediction accuracy is presented. The work of \cite{Liu2023fewshot} demonstrates a cross-city application of few-shot learning for traffic forecasting. This study focuses instead on improving the predictions of MFDs in data scarce scenarios. It does so by leveraging Meta-Learning to learn generalizations across urban environments with varying detector distributions and network structures. It therefore differs from the work of \cite{Sun2023meta}, which applies Meta-Learning to variance reduction in Monte Carlo methods, and of \cite{Liu2023fewshot} which explore few-shot dataset distillation. 
\\
Overall, based on the research summarized above, it appears that Meta-Learning has not been widely applied to the presented problem. Like for the conventional FD \citep{bramich2022fitting}, the shape of the MFD depends on the available data, as well as the predominant demand patterns \citep{tilg2021application, ambuhl2021uppermfd}. As a consequence, the wider scientific literature does not yet agree on one functional shape for the MFD. For example, some authors suggest adopting a cubic polynomial to approximate the continuous shape of the MFD \citep{huang2023characterizing, geroliminis2008existence}, while other authors suggest instead using a bi-parabolic interpolation of the data used as an analytical approximation \citep{daganzo2008analytical, batista2019biparab}. To derive the actual MFD shape is mostly still a data-driven task carried out for each case study and/or implementation. This paper argues that, given our analytical and theoretical understanding of the MFD, as well as datasets from different cities, it should be possible to approximate the MFD in cities where only limited data is available. 
It is therefore reasonable to hypothesize that, while Machine Learning techniques can exploit the data-driven nature of the problem, Meta-Learning may increase the transferability of results, learning patterns hidden in data-rich datasets and transferring it to new scenarios where less data is available. This paper builds upon and expands these hypotheses.  Thus, as it will be described, Machine Learning holds the potential to identify a MFD shape fitting data when data is fully available and to estimate a similarly fitting MFD, through Meta-Learning, in settings of data scarcity. 
\\
The contributions of this research can be summarized as:
\begin{itemize}
    \item Developing a non-parametric neural network (MTPINN) for MFD estimation that embeds domain knowledge
    \item Designing a Meta-Learning framework that succeeds in leveraging data patterns from data-rich cities for the estimation in data-scarce cities
    \item Proving the applicability of the proposed methods in the transport domain, by estimating the MFD in settings with as little as 10 LDs and achieving a fit comparable to the one achievable with the full LD dataset available
\end{itemize}

With this work, we contribute to the current state of the art by both expanding the number of cities for which a MFD can be estimated (i.e., cities with less LDs) but also providing a solution for cities that, for any reason, need to reduce the number of operational LDs (e.g., due to maintenance or budget constraints).

The paper is structured as follows. In Section~\ref{sec:data}, the dataset used to train both the developed MTPINN and model agnostic Meta-Learning framework (MAML) is detailed; In Section~\ref{sec:methodology}, the algorithms behind MTPINN and MAML are reported, while a bi-parabolic hybrid model is designed as benchmark; In Section~\ref{sec:experimental}, the design of the experiments are described and; In Section~\ref{sec:results} the results of the experiments are reported and the superior performance of the MAML with limited LDs data is proved. Section~\ref{sec:validation} validates the Meta-Learning framework by applying it to another non parametric model: FitFun \citep{bramich2023fitfun}. It also compares the Meta-Learning performance with Transfer Learning methods and conventional Neural Networks (NNs). Finally, Section~\ref{sec:limitations} and~\ref{sec:conclusions} summarize the limitations and future research directions. 

\section{Data} \label{sec:data}

As the meta-Machine Learning model will require data from multiple cities, we first introduce the adopted dataset. This research uses the UTD19 dataset \citep{loder2020utd19}, which contains LD data from 39 cities around the world. These case studies include cities with different features and characteristics, reflected in the spatial distribution of the LDs and the recorded data. Indeed, the spatial distribution of LDs has a relevant impact on the biases in the MFD, as proved by \cite{Lee2023LDpositioning}. It is important to capture different dynamics and types of urban scenarios, as the aim of the study is to design a Meta-Learning algorithm able to generalize to different case studies. 

\begin{table}[ht]
\centering
\begin{tabular}{|p{4cm}|p{4cm}|}
\hline
\textbf{LD features} & \textbf{Recording} \\
\hline
Number and length \newline of lanes monitored & Interval of \newline aggregation \\
\hline
Type of road & Occupancy\\
\hline
Time of the day &  Flow\\
\hline
Speed limit & Speed \\
\hline
\end{tabular}
\caption{LD features and recording information} \label{Dataset}
\end{table}

The data captured in the dataset is summarized in Table~\ref{Dataset}. MFDs are defined through average densities, flows, and speeds. In practice, occupancy is often used to approximate density, as it is easier to measure. As occupancy is properly measured in the UTD19 data, in this research, we will represent the MFD as a function of the average flow versus the average occupancy. Data were pre-processed using the same steps discussed by \cite{aghamohammadi2022parameter}. Namely, the following patterns have been deemed unreasonable and deleted from the data:

\begin{itemize}
    \item Negative or non-numerical flows or occupancy measures
    \item Occupancy larger than 1
    \item Flow measurements larger than 2500 veh/hr/ln
    \item Flow measurements less than 10 veh/hr/ln for occupancy values between 0.2 and 0.75
    \item Flow measurements higher than 100 veh/hr/ln for occupancy above 0.95
\end{itemize}

As in \cite{aghamohammadi2022parameter}, we only kept in the database detectors with valid data above 80\% of the time, and similarly intervals with valid observation from above 80\% of the detectors. Cities with no or limited occupancy measurements were excluded from the dataset. After data cleaning, only 29 cities were kept out of the original 39. 

\begin{figure*}[h]
    \centering
    \includegraphics[width=0.80\linewidth]{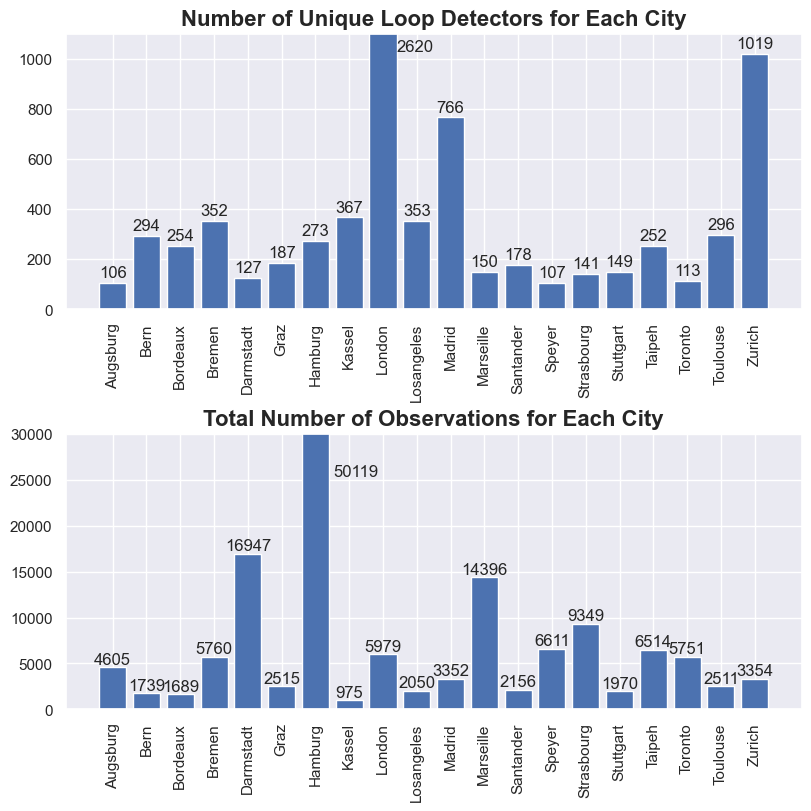}
    \caption{Overview of the number of LDs and the total number of observations for each considered city.}
    \label{fig:numberofLDs}
\end{figure*}

\begin{figure*}
    \centering
    \includegraphics[width=0.94\linewidth]{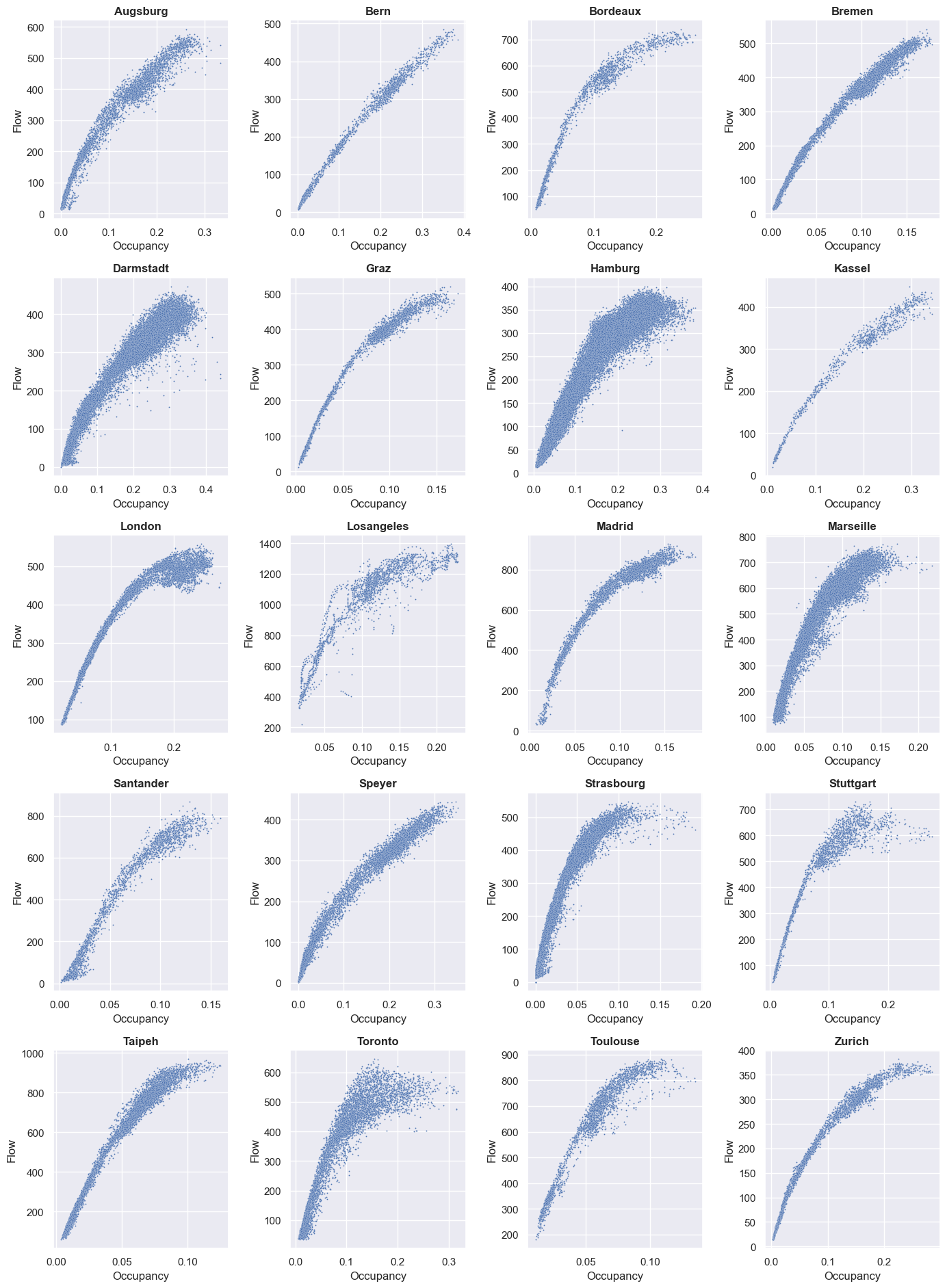}
    \caption{The plots show the average occupancy [\%] versus average flow [Veh/Hour/Lane] for the cities selected from UTD19 \citep{loder2020utd19} }.
    \label{fig:29_averages}
\end{figure*}

The number of LDs and of observations varies across cities, with the majority having between 100 and 350 detectors. Still, there are outliers, London for example stands out with 2620 LDs and, on the other side of the spectrum, Essen sees only 36 LDs. To ensure comparability between the cities included in the experiments, only those with at least 100 LDs available have been included in the study, leaving 21 appropriate cities. This has been done to ensure the sampling would be consistent across cities (e.g., it would not be possible to sample 75 LDs from Essen, as only 36 LDs are available). Furthermore, Luzern has been excluded as its number of observations exceed the city with the second highest number of observations (Hamburg) by more than 120.000. Thus, the number of considered cities is 20. The values for each city are reported in Fig.~\ref{fig:numberofLDs}. As it will be shown, this does not prevent us from testing the MAML in challenging configurations, as some scenarios will consider as little as 10 LDs. The resulting MFD plots, in terms of occupancy and flow, are instead depicted in Fig.~\ref{fig:29_averages}.

\begin{figure*}[h]
    \centering
    \includegraphics[width=1.04\linewidth]{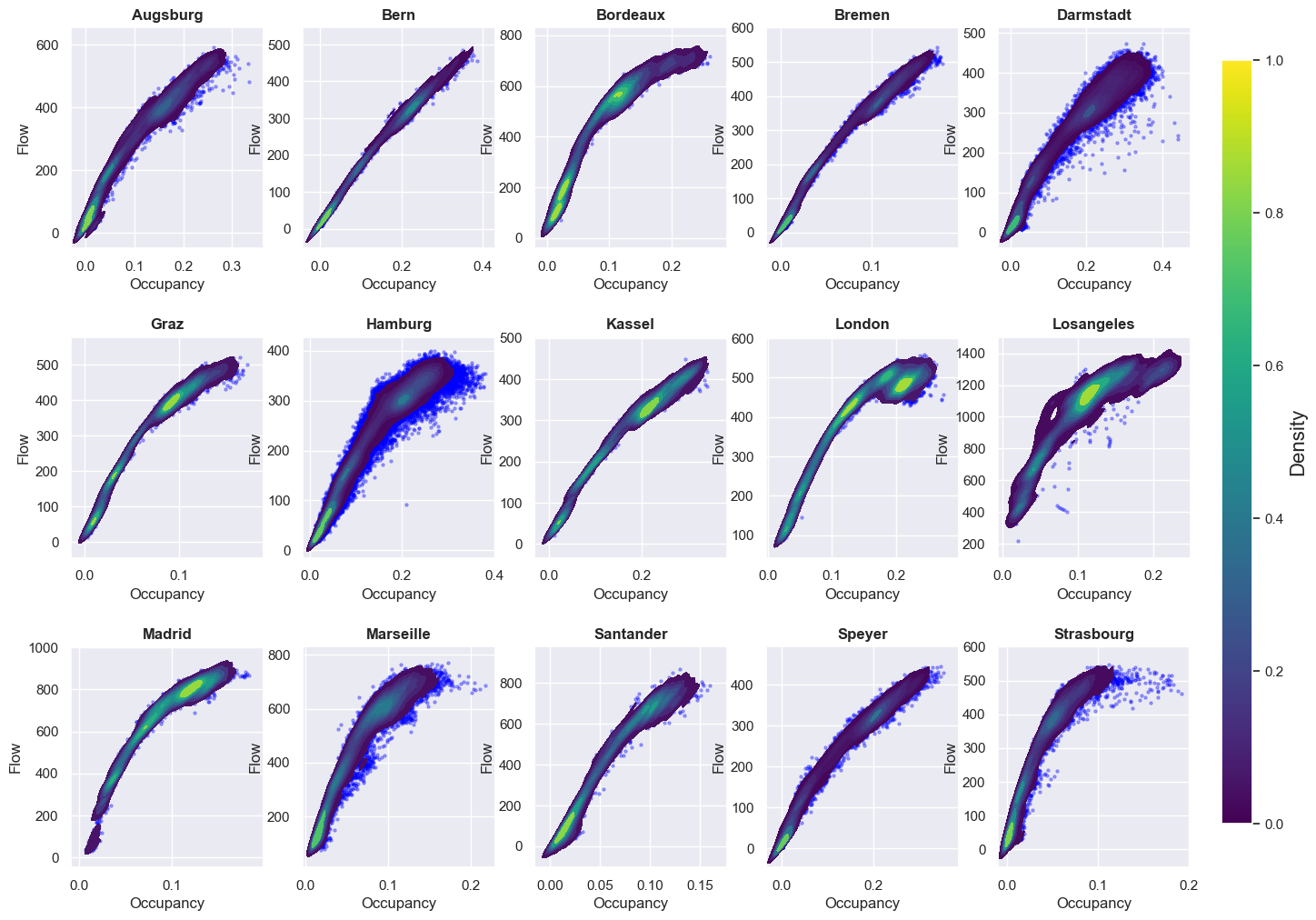}
    \caption{Distribution and density of observations of the average occupancy versus flow for a subset of the 20 cities with more than 100 LDs, based on data from UTD19 \citep{loder2020utd19}.}.
    \label{fig:densityplots}
\end{figure*}

The different distributions of LDs and the resulting wide range of values and patterns for the MFD are a desirable feature for the current aim, as a dataset framing different MFD trends is more suitable for training a Machine Learning model by the means of a Meta-Learning algorithm. Still, upon examination of the MFDs in Fig.~\ref{fig:29_averages}, some common trends emerge. For example, free-flow conditions are characterized by a quasi-linear behavior. Besides, in all the cities, the congestion branch (i.e., the portion of the theoretical fundamental diagram to the right of the critical density) is not fully observed. In some other aspects, different groups of cities see a different MFD shape arise from the recorded data. In some cities (e.g., Toronto), the maximum average flow values are observed for various average occupancy values,
resulting in a stretched-out top of the MFD curve. In other cities, e.g., Darmstadt, Hamburg and Marseille, a different pattern occurs
as the upper third of the flow values shows a higher spread in occupancy values, resulting in a stem-with-a-bulb shape. Interestingly, Darmstadt, Hamburg and Marseille are also the three cities with the the largest amount of observations.
Another relevant aspect to be analyzed and compared is how the density of observations varies for different datasets, namely which portion of the MFD contains more measurements. The density plots for a subset of cities are reported in Fig.~\ref{fig:densityplots}. Overall, it is observed that there is generally a very high density of observations in the lower left corner, where both flow and occupancy are low.
The variation in density along the curves then slightly varies between cities but it remains common to have a high density of observations towards
the top of the curve, suggesting that both free-flow and mildly congested regimes may be the most common.

Overall, while exploring the UTD19 dataset, several interesting characteristics of the MFDs were observed, including the shape, a time dependent aspect, and the presence of often well-defined MFDs for the separate road types in the dataset. On the overall shape of
the MFD, it was observed that the total amount of observations for a city appeared to relate to the amount of scatter or spread of the MFD. It was also noted that the shapes of the MFD from the cities with the most LDs did not differ notably. As biased datasets (i.e., datasets hiding portions of the available recordings to simulate different levels of data scarcity)
were constructed for the meta-models in the various experiments reported in Section~\ref{sec:MAMLexp}, it is important to note that selecting different subsets of the detectors for a given city gives rise to various shapes for the MFD, and it would appear that the fewer LDs are used, the more diffuse or spread out the MFD shape becomes.

\section{Methodology} \label{sec:methodology}

In this section we describe the algorithms adopted in this paper for estimating the MFD. The aim is for the models to estimate the shape of the MFD, so only the spatial dimension is considered across the paper. We first present a simple bi-parabolic model that will be used as a theoretical benchmark to validate the performance of other models in predicting the Critical Occupancy (CO) and the maximum flow of a MFD. Then, we introduce a non parametric Machine Learning model based on the Multi-Task Physics-
Informed Neural Network (MTPINN), which is specifically designed for MFD estimation on a single city based on LD data. Essentially, MTPINN represents a non-parametric alternative of the traditional bi-parabolic model. Finally, we discuss the algorithm behind the MAML framework and how it can be combined with the previous algorithms in order to estimate the parameters of the MFD in cities where only a limited number of LDs are available. The code to implement each of the 3 models is publicly available at https://github.com/s184227/MAML\_for\_MFD \\
The MFD is typically obtained by aggregating data from all road segments in the network, where the network-wide flow is the average flow across all segments in the network while the density is the weighted average density across all road segments. Therefore, in order to derive the MFD from occupancies and flows, one should first derive link-level density values from occupancy data. This conversion requires additional data, such as the length of the road segment on which the measurement from each loop detector is carried out and the average vehicle length for each city, to estimate the average detection zone length for each loop \citep{ bramich2023fitfun}. Fortunately, the link-level occupancy is approximately proportional to the density \citep{kim2004relationships, sun2014data, bramich2023fitfun}. Additionally, in this paper, flow and density values are normalized by the capacity flow and jam density. This entails that a linear relationship between the occupancy and density values can be assumed. Then, one can regard the occupancy values as the normalized density values \citep{aghamohammadi2022parameter}. This also applies to this study as in all these cases, the key variables used in the MFD are the aggregated flow and occupancy. In the equations presented below, the average occupancy is the independent variable $x$, while the average flow is the dependent variable $f(x)$.

\subsection{Bi-parabolic hybrid model}\label{sec:bi-parabolic_model}
The bi-parabolic hybrid model is designed to use two parabolas that share the same vertex to approximate the MFD in each city. These functions correspond to a shape founded in theory and allow a physical interpretation of their parameters. The model encodes this relationship as two piecewise parabolic functions that meet at a shared vertex, representing the CO at which traffic flow reaches its maximum. It is implemented as a fully interpretable white-box model, with its parameters estimated using backpropagation. It was implemented using PyTorch \citep{Paszke2019pytorch}, as it optimizes learnable
parameters directly via gradient-based methods, utilizing the specialized
components of PyTorch to combine physical intuition with
data-driven learning efficiently. A bi-parabolic model was chosen as theoretical benchmark to evaluate the performances of the models introduced in the following sub-sections (i.e., the MTPINN and the MAML), as its structure is perfectly known and thus the results are fully interpretable. It is acknowledged that the bi-parabolic model may stray away from the MFD predicted flow when the bi-parabolic functional form strongly differs from the one of the MFD, still the benchmarking is not carried on the average flows for any value of density but is instead carried out on the CO and maximum flow, where the bi-parabolic model proves more reliable across the 20 selected cities. Besides, the bi-parabolic shape is the most commonly used in the MFD literature \citep{batista2022exploring} and, as it will be shown, the described non parametric model actually tackles those cases in which the average flow predicted through the MFDs has a different functional form.

Using two parabolas, the bi-parabolic model predicts
traffic flow $f(x)$ as a function of $x$. The first parabola describing the uncongested regime is required to pass through the origin, $(0,0)$, and then ends at the vertex $(x_{co},f_{vertex})$, and is defined as:
\begin{equation}\label{eq:bi-parabolic_fit1}
    f_1(x) = -a_1 (x-x_{co})^2 + f_{vertex}, \; x \leq x_{co}
\end{equation}

\begin{figure*}
    \centering
    \includegraphics[width=0.70\linewidth]{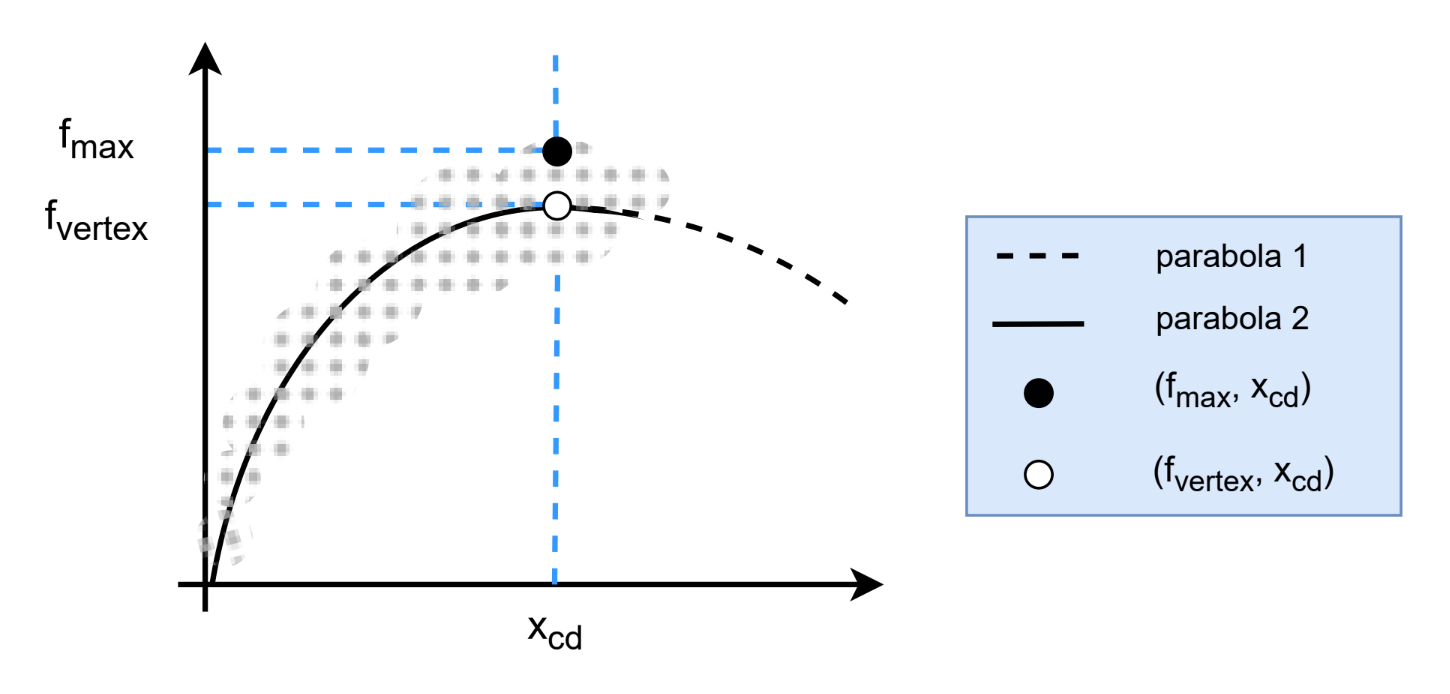}
    \caption{Relationship between the parameters of the bi-parabolic model and the MFD plot.}
    \label{fig:biparabolic}
\end{figure*}

Where $a_1 = \frac{f_{vertex}}{x_{co}^2}$ ensures that the parabola passes through the origin. The second parabola describes the congested regime and extends from the vertex to a second intersection with the x-axis:
\begin{equation}\label{eq:bi-parabolic_fit2}
    f_2(x) = -a_2(x-x_{co})^2 + f_{vertex}, \; x \geq x_{co} 
\end{equation}

where $a_2 > 0$. The negativity of $a_2$ is enforced through a logit transformation, ensuring that the parabola opens downward. The model is trained using a composite loss function, which can be summarized by:
\begin{equation}
    \mathcal{L}_{total} = ( \mathcal{L}_{1} + \mathcal{L}_{2} ) + \alpha \mathcal{L}_{\lambda}
\end{equation}
In this expression, $\mathcal{L}_{1}$ and $\mathcal{L}_{2}$ are the Mean Squared Error (MSE) losses for the two regimes given the two fits described in Eq.~\ref{eq:bi-parabolic_fit1} and \ref{eq:bi-parabolic_fit2}. As mentioned,
the model learns the optimal parameters used in the two parabolic
fits while training, given the loss functions. The intersection between
the two fits is their common vertex, and the appropriate location
for this is determined as a compromise between the quality
of the fits, i.e., their contribution to the total loss, and a penalty
that encourages the location to be consistent with theory. The
penalty is designed to encourage the model to determine the point
of critical occupancy to correspond to one of the max flow values
observed for the city. This is implemented by letting the penalty
term scale with the number of observations of flow values greater
than the flow value of the current chosen max flow, ($x_{co}$, $f_{max}$).
As the plot contains some scattering, it is desirable to allow some
flexibility by not forcing it to select the max flow value but simply
a value near the max flow. $\alpha \mathcal{L}_{\lambda}$ is a penalty function that ensures that the estimated CO fits the data while being consistent with the theory. The weight assigned to the qualities of
the fits versus the regularization through the penalty terms can be
balanced by tuning the hyperparameter $\alpha$.
The location of the common vertex for the parabolic fits is defined
to be the critical occupancy $x_{co}$ and a value $f_{vertex}$ that provides
a good fit. A visual representation is given in Fig.~\ref{fig:biparabolic}. The main feature of this model is that the parameters of the two parabolas and the critical occupancy are jointly estimated. 
$x_{co}$ is represented as a weighted softmax of the occupancy values and is calculated through gradient descent:
\begin{equation}
    x_{co} = \sum_{i = 1}^Nw_i \cdot x_i \hspace{2cm} w_{i} = \frac{exp(l_i)}{\sum_{j=1}^Nexp(l_j)}
\end{equation}

Where $l_i$ are learnable logits. Finally, weights are applied to compensate for the uneven data distribution. This step is needed, as the data distribution between the two parts of the fit is very
uneven, as the uncongested branch is fully observed, while only a few data points are available for the congested branch, often
corresponding to 5-10 \% of the total observations. Additionally, these observations are often concentrated around the CO point. To handle the
imbalanced data and ensure that the model prioritizes the quality
of both fits, the MSE losses L1 and L2 were weighted through Inverse Frequency Weighting, in which weights are assigned inversely proportional to the sample of each regime.

\subsection{Multi-Task Physics-Informed NN} \label{sec:mtpinn}

Similarly to the Bi-parabolic approach, the MTPINN is a non parametric model designed to estimate three key parameters of the MFD: the critical occupancy, the maximum flow, and the flow-occupancy relationship. MTPINN integrates Multi-Task Learning (MTL) and Physics-Informed Neural Networks. More specifically, Physics-Informed regularization techniques are used to introduce domain knowledge into the MTL architecture. Since the architecture is composed of two main components, we first introduce the MTL and its structure, and then its Physics-Informed constraints.  \\

\begin{figure*}
    \centering
    \includegraphics[width=0.54\linewidth]{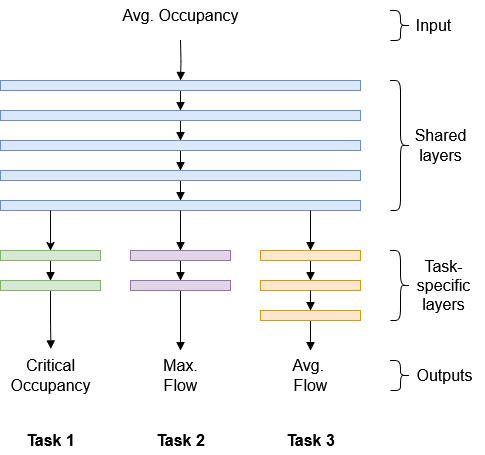}
    \caption{The figure shows a basic structure for a multi-task model. The shared layers learn a common representation of the input data, X, which is then fed into the separate output branches. These consist of separate, task-specific layers which are trained to predict different variables.}
    \label{fig:multitask}
\end{figure*}

\hspace{0.5cm}\textbf{\textit{Multi-Task Learning:}} MTL is a type of Machine Learning architecture where multiple related tasks are learned simultaneously within a shared framework. By leveraging shared representations, MTL can exploit
commonalities between tasks to improve learning efficiency
and generalization \citep{zhang2018MTL}. The structure of an MTL model typically consists of a shared base
and task-specific output branches, as shown in the example on
Fig.~\ref{fig:multitask}. The shared base includes common layers that learn a
unified representation of the input data, capturing features relevant
across all tasks. In the case of the MFD, it captures common features between critical occupancy, maximum flow, and flow-occupancy relationship. Each task-specific output branch comprises
separate layers tailored to predict each specific target task separately. This
modular design  improves learning efficiency and generalization. The specific network used in this research consist of:

\begin{enumerate}
    \item Feature extraction layers: a series of fully connected layers with ReLU  activations and dropout for regularization. These layers map input occupancy values to a latent feature representation and are shared between the prediction tasks.
    \item Output branches: as mentioned above, there are three output branches and each learn a separate task: flow predictor, critical occupancy predictor and maximum flow predictor
    \item Learnable parameters: Two learnable parameters are defined to help inform the loss. One is the offset (namely, a parameter used to define how much can the maximum flow and the flow at CO differ), and the other is the occupancy scaler (i.e., a parameter used to regulate the wideness proportion between the uncongested and the congested parabolas).
\end{enumerate}
The architecture is implemented in PyTorch \citep{Paszke2019pytorch}, a widely used
deep learning framework for Python. Note that, as shown in Fig.~\ref{fig:multitask}, the model requires prior knowledge of the critical occupancy, the max flow, and the average flow during training to minimize the loss. However, some of this information is not available before training (e.g., critical occupancy). Hence, the next section introduce the Physics-Informed Loss function that allows us to overcome this limitation.
\\

\hspace{0.5cm}\textbf{\textit{Physics-Informed Regularization:}}
While powerful, MTL is unlikely to capture the difference between the congested and the uncongested branch, especially with limited data in the congested branch. Therefore, we apply a Physics-Informed regularization to the Loss Function to force the model to respect analytical properties of the MFD. Hence, the loss function is composed of two components:

\begin{equation}
    \mathcal{L}_{total} = \mathcal{L}_{MSE} + \mathcal{\alpha L}_{Physics}
\end{equation}

$\mathcal{L}_{MSE}$ represents the loss of the purely data driven model, while $\mathcal{L}_{Physics}$ computes the Physics-Informed loss, designed to encourage the predictions to follow a bi-parabolic shape. $\alpha$ is a hyperparameter controlling the relative strength of the two loss components.
$\mathcal{L}_{MSE}$ is formulated as following, with $f_i$ and $\hat{f_i}$  being the predicted and estimated average flow, respectively:
\begin{equation}
    \mathcal{L}_{MSE} = \frac{1}{N}\sum_{i=1}^N (f_i - \hat{f_i})^2
\end{equation}

$\mathcal{L}_{Physics}$ is comprised of the following terms:
\begin{equation}
    \mathcal{L}_{Physics} = w_1 \cdot \mathcal{L}_1 + w_2 \cdot \mathcal{L}_2 + \lambda_{offset} + \lambda_{scale} + \lambda_{max}
\end{equation}

That is, two loss terms $\mathcal{L}_1$ and $\mathcal{L}_2$, weighted using $w_1$ and $w_2$, and three penalty terms for refining the prediction. Because the model is designed to predict some of the key values
for the MFD, namely the maximum flow, the critical occupancy,
and predicting average flow given occupancy, the physics loss is designed
to regulate how these parameters are learned and which values
they can take or are likely to take. $\mathcal{L}_1$ and $\mathcal{L}_2$ guide the model predictions and are designed so that the two parabolas share the vertex. The first parabola describes the observations from 0 to the critical occupancy, $x_{co}$. It is defined as follow:
\begin{equation}
    f_1(x) = a_1 + (x - x_{co})^2 + (f_{max} -  f_{offset})
\end{equation}
where the value $a_1$ is given by:
\begin{equation}
    a_1 = - \frac{(f_{max} - f_{offset})}{x_{co}^2}
\end{equation}
The second parabola describes the observations to the right of the critical occupancy and is defined as:
\begin{equation}
    f_2(x) = a_2 + (x - x_{co})^2 + (f_{max} -  f_{offset})
\end{equation}
where the value $a_2$ is given by:
\begin{equation}
    a_2 = - \frac{(f_{max} - f_{offset})}{((x_{scaler} - 1) \cdot x_{co})^2}
\end{equation}
The denominators in the expressions for $a_1$ and $a_2$ control the width of the parabolas, i.e., the horizontal distance between either intersection with the x-axis and the vertex point. For $a_1$, the desired distance is $x_{co}$, whereas for the second parabola, it
is expected to be different, and $x_{scaler}$ lets it be defined as some
multiplex of the $x_{co}$ distance. The shape of
the parabolas is controlled by the predictions made by the model
for the dataset, i.e., the shape parameter $a_1$ is calculated based
on the height of the parabolas’ vertices. This is determined by
$f_{max}$ - $f_{offset}$, as the best value for the fit will typically go through the mean of the spread of the flow data at $x_{co}$ and not the peak value. The three penalty terms each guide the model towards certain behaviors of the predicted MFD. $\lambda_{offset}$ encourages the location of the
model-determined critical occupancy to align with the plot region containing the max flow observation(s), $\lambda_{scale}$ encourages the width of the second parabola to be between 1 and 4 times that of the first one by penalizing widths outside this range. Finally, $\lambda_{max}$ penalizes flow values larger than the model-determined $f_{max}$.

\subsection{The MAML model} \label{sec:MAML}

This section describes the embedding of MTPINN within the Meta-Learning architecture. The underlying logic is for the MAML to learn a set of model parameters ($\theta$) that serve as good initialization for fine-tuning possible new tasks in the MTPINN. To do so, the MAML operates in two main loops:

\begin{itemize}
    \item \textbf{Inner Loop:} For each task, the model’s parameters are updated using a small number of gradient steps on task-specific data. These updates generate task-specific parameters $\theta^{\prime}_{t}$.
    \item \textbf{Outer Loop:} The meta-objective is computed by evaluating the task-specific parameters ($\theta^{\prime}_{t}$) on a separate validation set for each task. In the conventional MAML, a second-order derivative of the meta-objective is used to update the shared initialization parameters ($\theta$) \citep{finn2017model}.
\end{itemize}

\begin{figure*}
    \centering
    \includegraphics[width=0.94\linewidth]{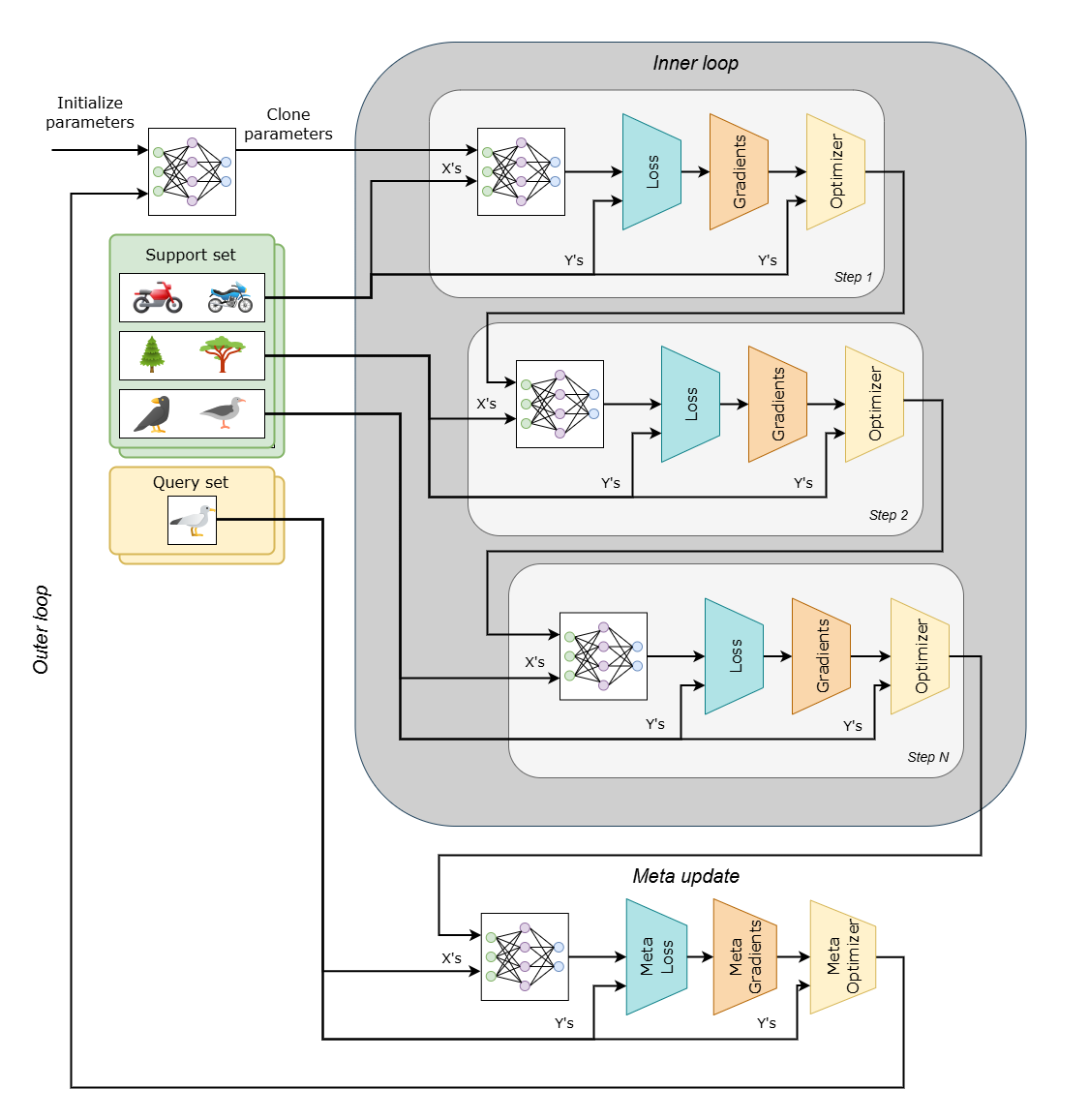}
    \caption{This figure illustrates the fundamental design of Meta-Learning models. A model is created and its parameters are initialized. These are then cloned for the inner loop of the meta-training, where the model is trained on a task-specific support set containing  $N_{samples}$ samples, including $h$ sub-tasks (or target classes) consisting of $k$ examples each (so $k \cdot h = N_{samples}$). This is also known as h-way, k-shot training tasks. In the example (green and yellow boxes), this corresponds to a 3-way, 2-shot task — i.e., 3 classes with 2 examples each. After training the model on the support set in $N_{ite}$ steps of the inner loop and updating the cloned model parameters accordingly, the model is tested on a query set containing $M$ observations from the $h$ sub-tasks included in the support set. The loss is stored, and the whole process is repeated - i.e., the original model parameters are cloned, and the inner loop trains the model on a new training task. This is known as an episode of meta-training, and $p$ episodes constitute an epoch of meta-training. At the end of each epoch, the stored losses from each episode constitute the meta-loss and are used to calculate the meta-gradients and optimize the meta-parameters of the model. This is then back-propagated through the model, and the model with updated parameters is used in the next epoch.}
    \label{fig:metalearning}
\end{figure*}

An overview of the Meta-Learning process is given in Fig.~\ref{fig:metalearning}. To implement Meta-Learning, first it is necessary to define a \textit{Task}. In our case, a task consists in using MTPINN to estimate the MFD parameters in a specific city, given a number of LDs. As mentioned, if data from multiple cities  is available, we can use this data to pre-train MTPINN using MAML. This can be done by generating a number of artificial Tasks for each city in the UTD19 database. Each artificial Task corresponds to the averaged network occupancies and flows - generated using a sub-set of randomly selected detectors. For each Task the model will attempt to estimate the target MFD parameters (i.e., those that would have been obtained using data from \textit{all} LDs). In the inner loop, MTPINN will attempt to locally adjust the model parameters ($\theta$) in a few steps, avoiding overfitting. In the Outer Loop, MAML evaluates how close the estimate is to the target MFD, which is the MFD that would be obtained if all links were equipped with LD. The Meta-Learning framework is based on the MAML formulation \citep{finn2017model} and, as it will be shown, succeeds in estimating the MFD diagram when only a limited number of links is equipped with detectors. The proposed implementation of MAML differs from the one from \cite{finn2017model}, as the original method cannot be directly applied to the MFD without explicitly accounting for the data aggregation step necessary to generate a MFD. The learn2learn software library \citep{Arnold2019learn2learn}, which builds on top of PyTorch, provides support for Meta-Learning in Python and is used
to implement the model.\\

\hspace{0.5cm}\textbf{\textit{Meta-Training Procedure:}}
Traffic data is organized into city-wise groups, with separate
training (support set) and validation (query set) datasets for each
city. The way these two sets are created in this paper is different from the way tasks are usually generated, to take into account that the MFD is an aggregated representation of LD data. For Meta-Learning, tasks are defined at the city level, where each Task corresponds to predicting the average flow based on occupancy data for a specific city. The support set is used for inner-loop adaptation, and uses only a subset of LDs to compute the aggregated flows and occupancies. The query set includes instead the aggregated flows and occupancy computed using all LDs in the network and evaluates the performance of the adapted version of the model resulting from the inner loop. A class has been implemented to handle task sampling to ensure the modularity of the code. The cities to be used in the support and query sets are sampled randomly from the training and validation data of the respective city for each task, ensuring that the spatial distribution of LDs does not impose a bias on performance of the models. A consistent random seed ensures reproducibility. The data used for validation is the flow and occupancy averages created from a given city’s full dataset. For training, the data
consists of averages calculated from a subset of randomly selected detectors for each city for the entire observation period available for a given city. The chosen number of detectors for sampling is n = 75, 50, 25, and 10. For each city, this number of detectors is randomly sampled 30 times, resulting in four datasets - one for each of the numbers of sampled detectors - of 30 biased averages for each city. To ensure that the constructed biased MFD datasets are significantly different from the full MFD datasets, only cities with more than 100 detectors are used. Furthermore, only cities with more than one day of observations are included, leaving 20 cities.
\\
The structure of the meta-training is built around two optimization objectives. The objective of the inner loop is to minimize the task-specific loss $\mathcal{L}_{task}$ by adapting the model parameters $\theta$ for a given Task $\mathcal{T}_i$:
\begin{equation}
    \theta'_i = \theta - \alpha \nabla_{\theta} \mathcal{L}_{task} (\mathcal{D}_i^{support};\theta)
\end{equation}
where $\alpha$ is the inner loop learning rate, $\mathcal{D}_i$ is the support set for $\mathcal{T}_i$ (the sampled task $i$), $\theta$s are the shared meta-parameters that are cloned for each new task and $\theta'_i$ are the task-adapted parameters for $\mathcal{T}_i$. While each $\theta'_i$ represents the learned NN weights to estimate the MFD in one specific city (i.e., Task), $\theta$ represent the best identified shared representation, result of the MAML "learning to learn" how to tackle a new MFD estimation problem. For the MTPINN model, this shared representation is captured by the set of weights $\theta$ (namely the weights of the multiple linear layers - feature extractor, flow predictor, critical occupancy predictor and max flow predictor - introduced with the MTPINN). The meta-objective is to minimize the expected loss over the query sets of all tasks after performing the inner-loop adaptation:
\begin{equation}
    \min_{\theta} \mathbb{E}_{\mathcal{T}_i \sim p(\mathcal{T})} [\mathcal{L}_{task} (\mathcal{D}_i^{query};\theta'_i) ]
\end{equation}
where $\mathcal{D}_i^{query}$ is the query set for $\mathcal{T}_i$, $\theta'_i$ is computed from the inner loop, and $p(\mathcal{T})$ is the distribution over tasks.\\
By differentiating through the inner-loop updates, the meta-gradient is computed to optimize $\theta$ using the outer-loop learning rate $\beta$:
\begin{equation}
    \theta \leftarrow \theta - \beta \nabla_{\theta} \sum_i \mathcal{L}_{tassk} (\mathcal{D}_i^{query};\theta'_i)
\end{equation}

\begin{algorithm}[h!]
\caption{Adapted MAML for Few-Shot Supervised Learning}\label{alg:maml-algo}
\begin{algorithmic}[]
\Require{$p(\mathcal{T})$: distribution over tasks}
\Require{$\alpha, \beta$: step size hyperparameters (learning rates)}
\State randomly initialize $\theta$ 
\While{not done}
    \State Sample batch of tasks $\{\mathcal{T}_i\} \sim p(\mathcal{T})$
    \For{\textbf{all} $\mathcal{T}_i$}
        \State Sample $K\cdot N_{ite}$ obs. $\mathcal{D}^\text{support}_i = \{x^{(j)}, y^{(j)}\}_{j=1}^K$ from $\mathcal{T}_i$
        \State Sample $M$ obs. $\mathcal{D}^\text{query}_i = \{x^{(j)}, y^{(j)}\}_{j=1}^M$ from $\mathcal{T}_i$
        \State Initialize task-specific parameters $\theta'_i \gets \theta$
        \For{\textbf{$N_{ite}$ inner loop steps}}
            \State Compute task-specific loss:
            \State $\mathcal{L}_\text{task}(\mathcal{D}
            ^\text{support}_i; \theta'_i)$
            \State Update task-specific parameters:
            \State $\theta'_i \gets \theta'_i - \alpha \nabla_{\theta'_i} \mathcal{L}_\text{task}(\mathcal{D}^\text{support}_i; \theta'_i)$
        \EndFor
        \State Compute query loss: $\mathcal{L}_\text{task}(\mathcal{D}^\text{query}_i; \theta'_i)$
    \EndFor
    \State Accumulate meta-gradient: $\nabla_\theta \sum_i \mathcal{L}_\text{task}(\mathcal{D}^\text{query}_i; \theta'_i)$
    \State Update meta-parameters: $\theta \gets \theta - \beta \nabla_\theta \sum_i \mathcal{L}_\text{task}(\mathcal{D}^\text{query}_i; \theta'_i)$
\EndWhile
\State \textbf{return} $\theta$ 
\end{algorithmic}
\end{algorithm}

Algorithm \ref{alg:maml-algo} describes the steps in the model's training procedure in pseudo-code and is closely inspired by \cite{finn2017model}. However, several differences with the original design exist. 
The differences arise as a result of adapting the algorithm to more efficiently address the specific type of data scarcity typically faced when working with MFD estimation, namely not few observations over a short period but biased observations due to limited and unevenly distributed LDs. Therefore, each task corresponds to the MFD of a given city resulting from the specific sampled subset of all detectors. Put differently, the available data points for a task are many but biased. Whearas the original MAML framework assumes very few observations, $K$, are available, and thus uses e.g., one observation (in the case of the one-shot learning) for each gradient step of the inner loop, this adaptation uses all available information $N$ at each step, to leverage the number of observations available across all cities in the support set. $N$ represents, therefore, all available data points for the MFD, as plotted in Fig.~\ref{fig:29_averages}. As the algorithm imposes that $K \cdot N_{ite} = M$ and the number of query points $M$ is a hyperparameter, the total number of observations $N$ represents an upper bound for the data used in the Meta-Training step, therefore $M\leq N$. In the inner set, these points are generated using a subset of detectors, while in the query set using all available LDs.
\\

\hspace{0.5cm}\textbf{\textit{Meta-Testing Procedure:}}
Every time a dataset is constructed, three random cities out of the 20 available are randomly selected for testing and separated from the training data (support set) so that the data used for testing is unseen for the model. The Meta-testing phase is designed to imitate the situation in which we use the MAML model pre-trained on multiple cities to estimate the MFD for a city where only a limited number $n$ of detectors are equipped with sensors. After the model has been trained, it is tested on three random cities, and the following statistics are stored for later evaluation: MAE, MSE, root relative squared error (RRSE) and the correlation coefficient $r$. This coefficient measures the strength and direction of the linear relationship between the true and predicted values, it is calculated by:

\begin{equation}
    r = \frac{\sum_{i = 1}^N(y_{i}-\overline{y_{i}})(\hat{y_i}-\overline{\hat{y_i}})}{\sqrt{\sum_{i=1}^N{(y_{i}-\overline{y_i})^2}}\sqrt{\sum_{i=1}^N(\hat{y_{i}}-\overline{\hat{y_{i}}})^2}}
\end{equation}

The meta-testing procedure mimics the inner loop in that the trained MAML model takes the same number of $N_{ite}$ gradient steps as the number of inner steps and uses the same number, $K$, of observations from the support test set for each of these in the inner loop. The learning rate for the inner loop is used, and as
such, the model follows the same procedure as in the training
phase when learning to adapt to a new task, i.e., sampled data
set for a city before making predictions for the query set.
When the model has adapted to the test dataset through the $N_{ite}$ gradient steps, predictions are made for the entire query set for
the city in question.

\section{Experimental setting}\label{sec:experimental}

This section introduces the experimental settings for each of the models. 
\subsection{The bi-parabolic hybrid model}
The purpose of the experiments for the bi-parabolic hybrid model
is to get robust fit of the MFD using the full dataset, to be later used as benchmark. In other words,
using all detectors available for each city, the goal is to have
the model determine the parabolic functions that best describe the MFD and determine the critical occupancy and
maximum flow in accordance with theoretical descriptions. Two hyperparameters have to be tuned, based on the formulation reported in Section~\ref{sec:bi-parabolic_model}: $\alpha$ and $\beta$. To determine the optimal parameter values, all
combinations of [0.01, 0.1, 1.0, 10.0] for the two parameters are
executed, and the learning rate is also varied between 0.001 and
0.01. These combinations are applied to all cities. The optimal combination is $\alpha$ = 1.0, $\beta$ = 0.1, and a learning rate
of 0.01. The optimal learning rate is determined by studying
the trace plots for the training loss, and the optimal values for
$\alpha$ and $\beta$ are determined by how well the parabolas describe the MFD curves, if the location of the maximum flow is suitably closely aligned with the actual maximum flow values and if the location of the critical
occupancy aligns well with the limit between the uncongested and
congested regime.

\subsection{MTPINN}
Like the bi-parabolic model, MTPINN has been trained on normalized occupancy and flow data. Additionally, an hold-out based approach is used during training. The dataset is
divided into training, validation, and test subsets to ensure robust
evaluation.
To determine the combination of parameters that result in the
optimal fit, an experiment is carried out where one parameter is
varied at a time to give an intuition on the influence of each parameter
on the fit. Based on this, a subset of the values are selected for
each parameter, and they are then all tested out in combinations
in nested for-loops. For each city the goal is for the model to adapt well to all
cities and make adequate predictions without tuning the parameters
to each specific city. 3 values for each of the 4 hyperparameters are evaluated. This means that $3^4 = 81$ combinations are tested for 20 cities, i.e., 2349 models are trained. The resulting set of hyperparameters is reported in Table ~\ref{tab:hyperparams}.

\begin{table}[ht]
\centering
\begin{tabular}{|p{4cm}|p{4cm}|}
\hline
\textbf{Parameter} & \textbf{Value} \\
\hline
$\alpha$ & 1.0 \\
\hline
Learning rate & 0.001\\
\hline
Batch size & 32\\
\hline
Dropout & 0.0 \\
\hline
\end{tabular}
\caption{Set of hyperparameters providing the best fit across the 20 cities} \label{tab:hyperparams}
\end{table}

\subsection{MAML} \label{sec:MAMLexp}
The training for the MAML approaches the problem of MFD from a slightly different angle. Indeed, while the previous models have been trained to estimate the MFD observed from the complete dataset for each city, the meta-model is trained to learn common patterns across the different cities and crucially how to transfer these (e.g., in conditions of data scarcity). Therefore, datasets have been created to simulate the scenarios of having only 75, 50, 25, and 10 randomly chosen detectors available in each city, resulting in biased MFDs. The MTPINN model is chosen to be the ’learner’ model used in the Meta-Learning algorithm, meaning that the goal will be to find the optimal initialization of the parameters and weights for this model through the MAML algorithm described in Section~\ref{sec:MAML}. The first steps include defining the learning rates for the inner and outer
loops and initializing the model parameters. The MTPINN model has two learnable parameters, the offset $o$ and the occupancy scaler $s$, that relate to the penalty terms in the physics-informed loss. They are initialized to 0.0 and 3.0, respectively. An offset of 0.0 means that the biparabolic
fits used to inform the physics loss term will pass through the vertex specified by the critical occupancy and the maximum flow value, as opposed to being ’offset’ to a lower flow value. Initializing the occupancy scaler to 3.0 corresponds to assuming the width of the second parabola the loss term being twice as wide as the first parabola, which is not uncommon. In addition to the learning rates, several other hyperparameters can be adjusted for the model to facilitate Meta-Learning. This includes the number of inner loop steps per task $N_{ite}$, the number
of meta-iterations, the number of tasks sampled per meta-iteration, the number of support observations $K$, used for each inner loop step, the number of support observations $M$, used for
meta-testing, and the dropout rates used in the meta-training and the meta-testing phases. The different values tested for each parameter have been limited to those shown in Table~\ref{tab:hyperparamsMAML} with the chosen optimal values indicated in bold.

\begin{table}[ht]
\centering
\begin{tabular}{|p{6cm}|p{6cm}|}
\hline
\textbf{Parameter} & \textbf{Values tested} \\
\hline
$\alpha$, inner learning rate & 0.01, \textbf{0.02}, 0.05 \\
\hline
$\beta$, outer learning rate & 0.001, \textbf{0.005}, 0.01\\
\hline
$N_{ite}$, inner loop steps & \textbf{5}\\
\hline
Number of meta-iterations & \textbf{150}, 300 \\
\hline
Number of Tasks $|\mathcal{T}_i|$ per meta-iteration & 1, \textbf{3}, 5\\
\hline
$K$, support obs. per inner step & 50, \textbf{150}\\
\hline
$M$, support obs. for meta-test & 250, \textbf{750}\\
\hline
Dropout rate in meta-training & \textbf{0.0},0.1\\
\hline
Dropout rate in meta-testing & \textbf{0.0},0.1\\
\hline

\end{tabular}
\caption{Hyperparameter values for the experiments with the metamodel,
the ones selected are indicated in bold.} \label{tab:hyperparamsMAML}
\end{table}

The model parameters were tuned using the dataset containing the biased MFD datasets from
75 randomly chosen detectors and then applied to the datasets created from 50, 25, and 10 randomly chosen detectors for comparison. 30 datasets were constructed from this number of cities (i.e., 30 different combinations of 75 - or 50, 25, 10 - LDs among the eligible ones), and sampling a task for the support set in the training corresponds to sampling $K \cdot N_{ite}$  observations from one of these datasets (with $K$ being the number of observations sampled in the sub-task). To better account for variability in the datasets within and between cities, the model is trained 50 times on differently initialized task sets on each of the datasets created from the 75, 50, 25, and 10 randomly chosen detectors. The performance of the MAML for each city in the query set is invariant to the order by which different cities are considered in the training process, as it can be noticed by comparing the outputs of the MAML across all the experiments (the extended results can be found in the Github repository). Indeed, most cities end up in the query set multiple times as the model is trained 50 times. Still, the results reported in the mentioned tables remain similar despite different training sets.

\begin{figure*}[h]
    \centerline{\includegraphics[width=0.9\linewidth]{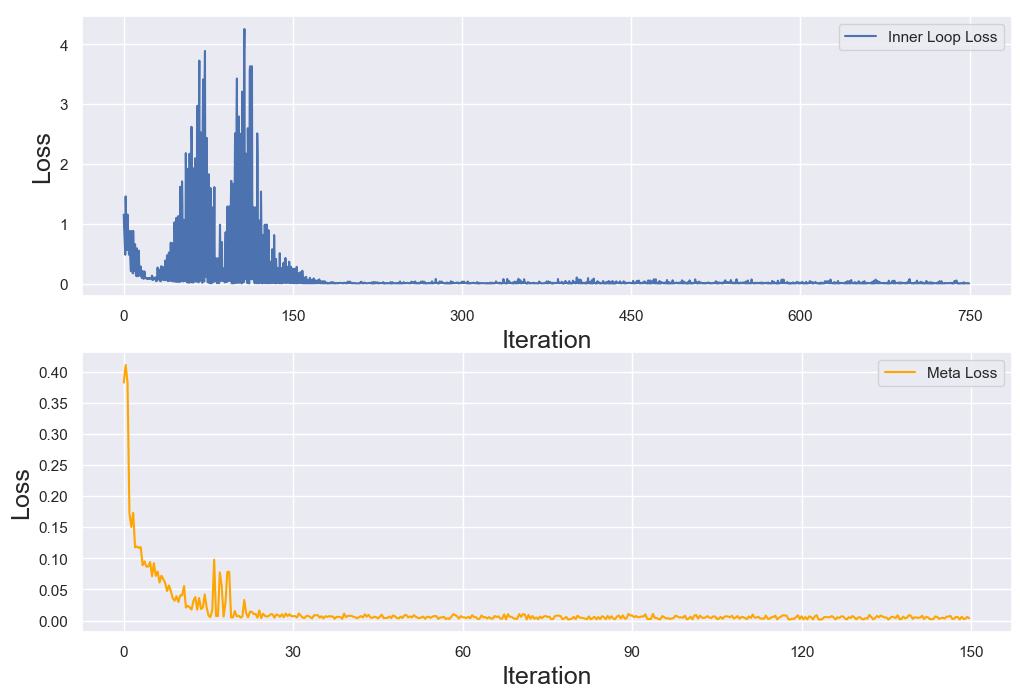}}
    \caption{Meta-losses for the inner and outer loop during meta-training of the model using datasets consisting of averages based on observations from 75 randomly selected detectors in a city. The training loss corresponds to one experiment, which consists of training the model for 150 meta-iterations, i.e., on 150 randomly sampled tasks (city datasets). For each meta-iteration, the model parameters are cloned and updated in 5 gradient steps in the inner loop, allowing the model to optimize its parameters for the current task. This is seen in the upper plot in the oscillations, where a new spike occurs each time the model is introduced to a new task. The meta-parameters, i.e., the model parameters cloned in the inner loop, are used to compute the meta-loss (bottom plot) based on the results of the inner-loop performed on 3 different tasks. The meta-loss is then used to update model parameters before the next meta-iteration.}
    \label{fig:MAMLtraininglosses}
\end{figure*}

An example of the development of the loss during the 150 meta-iterations of meta-training is shown in Fig.~\ref{fig:MAMLtraininglosses} for the model trained on a task set from the 75 detectors dataset. Fig.~\ref{A3} plots instead the same results for the 10 detector datasets.

In each task set, three out of 20 cities (see list in Fig.~\ref{fig:numberofLDs}) are selected for metatesting,
meaning that the meta-model is trained on the 18 other
cities and then tested using the support and query sets for the
three held-out cities. 
To carry out the comparison reported in the following Section, an MTPINN model is also
trained from scratch on the $M$ observations (obtained with averaging observations from the correspondent sub-set of LDs) used by the metamodel
during the test phase to adapt to a new task, i.e., city.
They both make predictions that are evaluated on the query sets,
the predicted maximum flow and critical occupancy are stored for
later comparison, and the calculated performance statistics for the
flow predictions are also stored. The parameters of the MTPINN
have been updated from the ones used in the experiments on the
full dataset reported in Section~\ref{sec:results} to achieve the best performance on this dataset. It is trained on batches of size 10 for 100 epochs, and a dropout rate of 0.1 has been found to improve performance.

\section{Results and discussion}\label{sec:results}
In this section, we assess the performance of the developed models, first with all of the available LDs available and then in conditions of data scarcity (75/50/25/10 LDs). It will be showed how, without Meta-Learning, the reduction of the number of LDs prevents from estimating a realistic MFD (i.e., average flow, critical occupancy and maximum flow in the target city). When Meta-Learning is implemented, it is instead possible to estimate said parameters with less measurements. As it will be discussed, the results prove that MAML improves the estimated MFD when only a subset of the LDs are available, thus allowing to extend results to cities equipped with fewer detectors or, possibly, to actually reduce the number of LDs in more thoroughly equipped cities (e.g., due to large scale maintenance or to reduce costs). 

\subsection{Full exploitation of the LDs dataset}\label{sec:fullLD}
Here, we present the results of the bi-parabolic model and of the MTPINN, focusing on how they perform in estimating the MFD for different cities, when all the set of detectors are available (and thus, in less challenging conditions that are not transferable to cities where the amount of LDs is a problem). Still, it is especially important to assess how the MTPINN fares as meta-model for the UTD19 dataset when the full set of LD recordings is available, as the performance of the developed MTPINN strongly affects the capabilities of MAML, which are discussed later in this Section.

\begin{figure*}[h]
    \centering
    \includegraphics[width=0.94\linewidth]{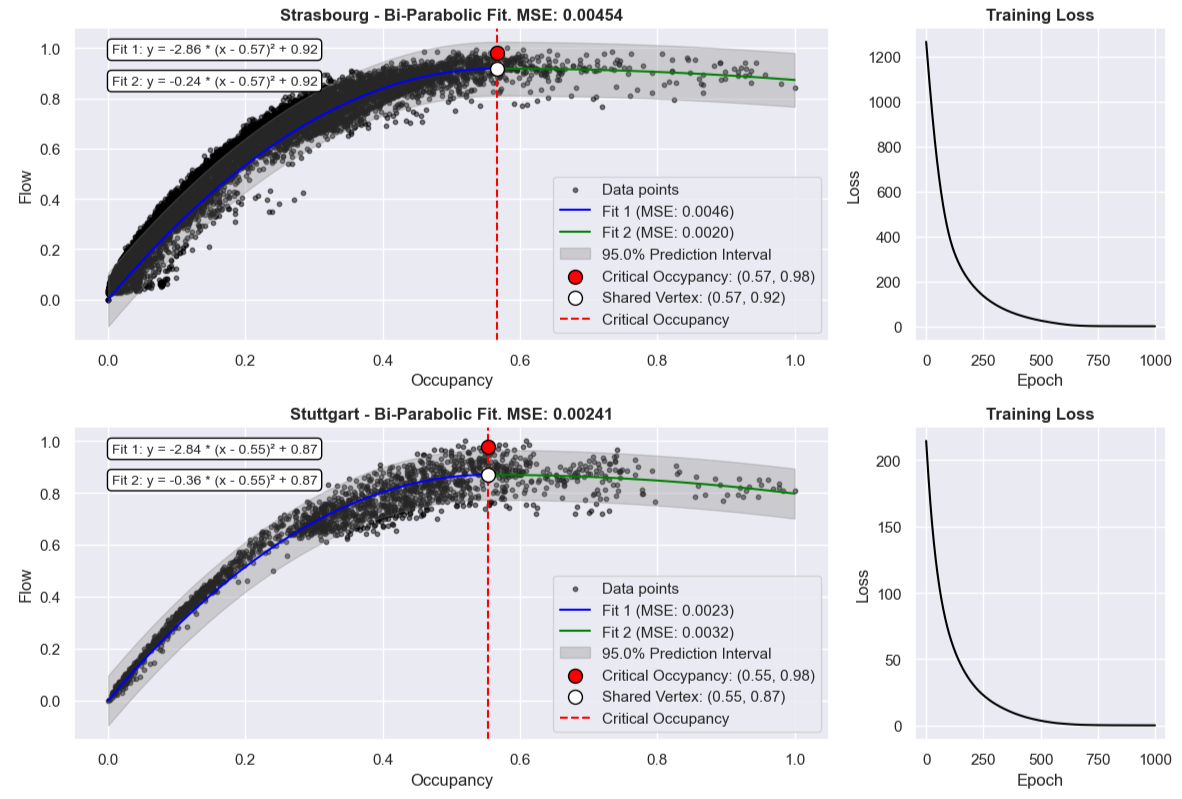}
    \caption{Results of the bi-parabolic hybrid model for two different cities - Strasbourg and Stuttgart - with differently shaped MFD diagrams and varying amounts of available observations. The model determines the point of critical occupancy and maximum flow, and the two parabolas meet at a shared vertex determined by the corresponding occupancy value, while the flow value for the vertex is determined by the best fit of the parabolas. The shaded area shows the 95\% prediction interval. The plots to the right show the training loss as a function of epochs, i.e., the number of gradient updates for the parametric model.}
    \label{fig:biperformance}
\end{figure*}

The bi-parabolic model performs reasonably well regardless the type of MFD, as shown in Fig.~\ref{fig:biperformance}. For all cases, the total loss decreases rapidly at first but eventually evens out, indicating that the model improves and eventually converges. A subset of the results is reported in Fig.~\ref{fig:biperformance}, the full set being available at https://github.com/s184227/MAML\_for\_MFD. The considered cities indeed represent typical characteristics for the MFD plots described in Section~\ref{sec:data} and reported in Fig.~\ref{fig:29_averages}, such as increased scattering associated with a large amount of observations (e.g., Toronto) and the stem-with-a-bulb shape (e.g., Stuttgart) where the MFD curve is narrow for the lower occupancy values but broadens at the top. These scenarios reflect different traffic patterns across the networks (e.g., different road types or prevalent lightly congested regimes) and ensure the models are able to capture MFDs with different characteristics. The observations result of the bi-parabolic hybrid model mainly lie within the shaded area, indicating a 95\% prediction interval, generally a fair benchmark to compare the results of the MTPINN and of the MAML. It should be noted though that there are few outliers. Many of the measurements for Santander, for example, fall below the 95\% range, especially near the origin (Fig.~\ref{fig:biperformance_santanders}), a behavior Santanders share only with Bern and Bremen out of the 20 considered cities. As mentioned in Section~\ref{sec:methodology}, this is to be expected when the real shape of the MFD is very different from the bi-parabolic functional form.

\begin{figure*}
    \centering
    \includegraphics[width=1.\linewidth]{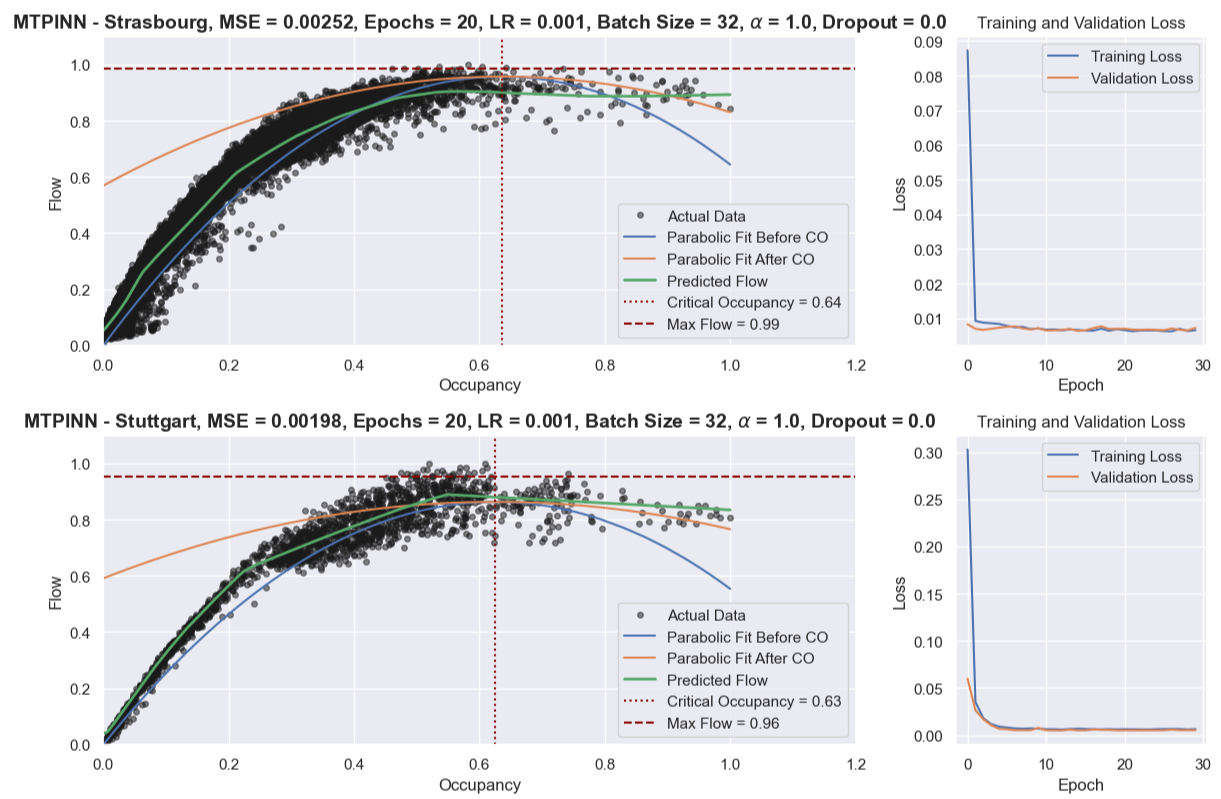}
    \caption{Results of the MTPINN model for two different cities - Strasbourg and Stuttgart, - with differently shaped MFD diagrams and varying amounts of available observations.}
    \label{fig:MTPINNperformance}
\end{figure*}

The MTPINN also reaches a satisfactory performance when the data it is trained against has enough entries both in the not congested and in the congested sides of the parabola, as in the subset in Fig.~\ref{fig:MTPINNperformance}. Still, the prediction of the MFD is less solid when data in the congested regime is scarce. This is what emerges from Fig.~\ref{fig:MTPINNspeyerstdrs}, which shows again a subset of the predictions. As mentioned in Section~\ref{sec:methodology}, one of the penalties embedded in the loss function guides the model towards a parabola for the congested regime (orange in Fig.~\ref{fig:MTPINNperformance}) between 1 and 4 times the width of the first blue parabola (uncongested regime). Still, this condition is not satisfied for example for Speyer and Santanders (Fig.~\ref{fig:MTPINNspeyerstdrs}). This is due to the fact that, for Speyer, it is not clear that a relevant portion of the measurements belongs to the congested branch while, for Santanders, while the beginning of the congested regime is observed, it does not include the expected decreasing flow, which the observations for Strasbourg and Stuttgart capture part of. 

This extends to the other cities; when the observations include part of
the characteristic behavior for the congested branch, the physics-informed
loss term successfully helps guide the model to predict a downward trend, but when none or very little of the congested branch is observed the model assigns very little importance to this compared to the overall trend of increasing flow. Besides, for all the considered cities, the green line of the model average flow prediction has a shape similar to the blue parabola but diverges a bit to provide a closer fit to the pattern of the distribution of observations for the city. This changes roughly at the point of the critical occupancy, where the orange parabola provides the MSE loss term for the physics-informed loss
function.

\subsection{Performance with biased and limited LD data}\label{sec:biasedLD}
\begin{figure*}[h]
    \centering
    \includegraphics[width=1.\linewidth]{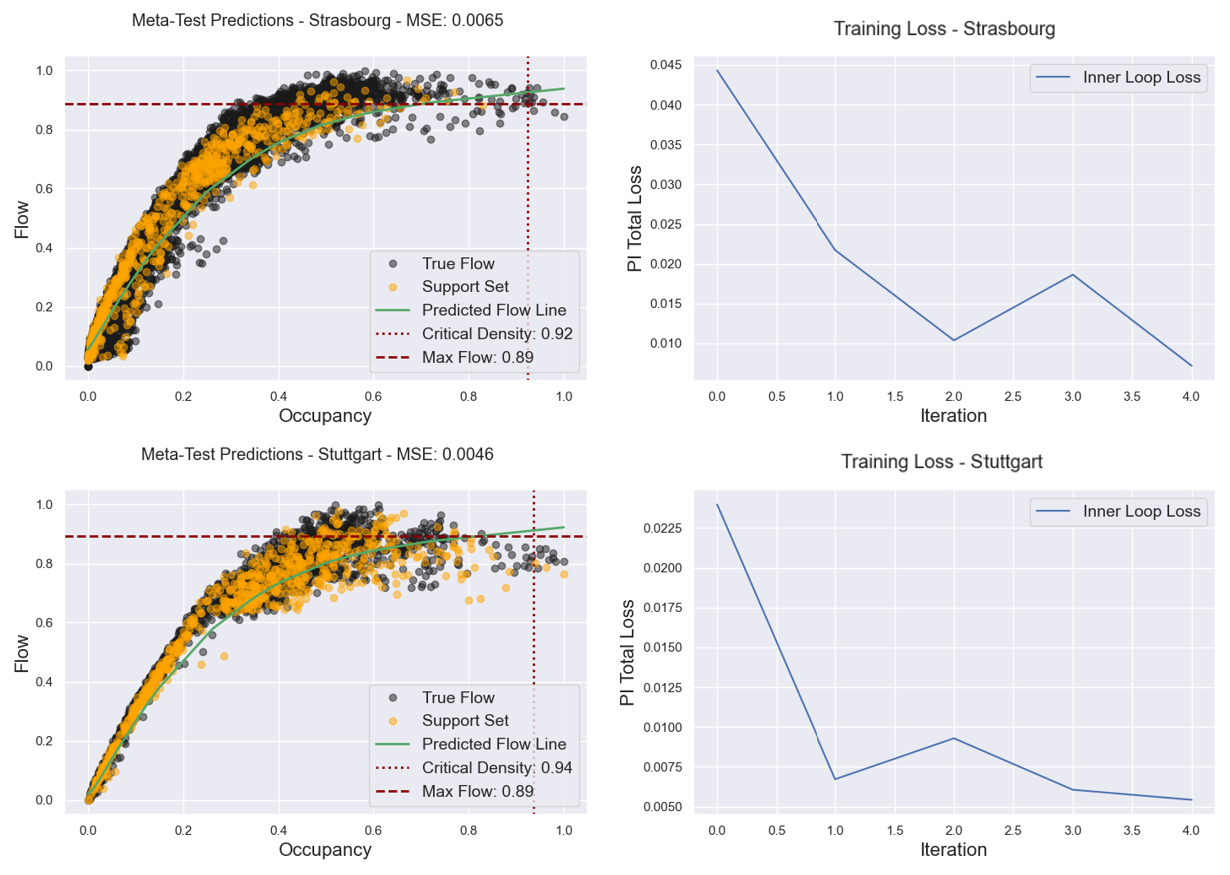}
    \caption{Meta-test sample of a meta-model trained on 75 detector datasets - The black dots are the observations from the full dataset using all available detectors, and the yellow dots show the $K \cdot  N_{ite} = M$ sampled observations from the biased MFD dataset that the model has used to adapt to the previously unseen task before making predictions, i.e., the green line.}
    \label{fig:MAML75performance}
\end{figure*}

\begin{figure*}[h]
    \centering
    \includegraphics[width=0.85\linewidth]{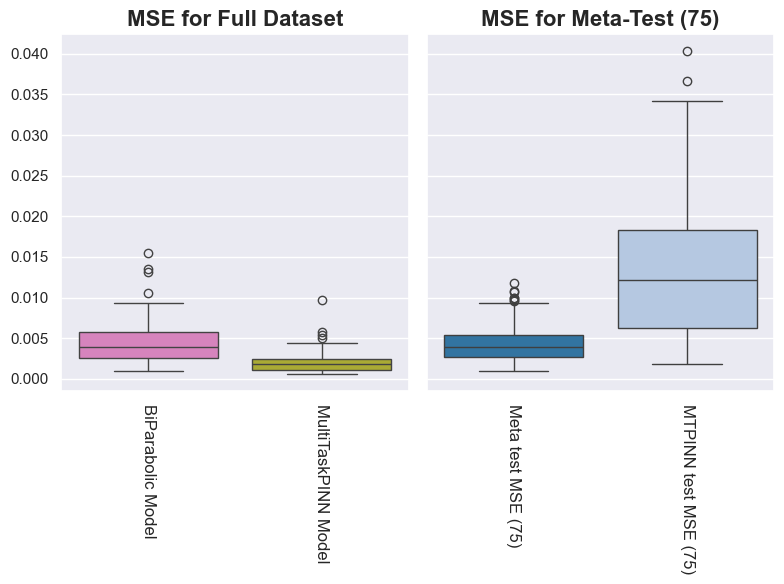}
    \caption{The boxplots show the distributions of the MSE calculated for the bi-parabolic model and the MTPINN model trained on the full datasets. The MSE for the meta-model and the MTPINN comparison model have been included as a reference for scale.}
    \label{fig:msecomparison}
\end{figure*}

From the results in Section~\ref{sec:fullLD}, it becomes clear how predicting the MFD for a city, even while exploiting the full set of available detectors, is no trivial task as each set of LDs rarely captures all the portions of the fundamental diagram. The problem quickly becomes intractable for a traditional Machine Learning model, when fewer LD recordings (and thus data points) are available. Still, this is a common occurrence in real life case studies, as many cities do not have full LD coverage. To overcome this challenge, the developed framework exploits Meta-Learning to learn a shared representations that allows the NN model to exploit the generalizable patterns and predict MFDs in cases with limited LD data.
As mentioned in Section~\ref{sec:MAMLexp}, different shares of randomly selected detectors are adopted for training, with varying values of 75, 50, 25 and 10 LDs. It should again be stressed how this is not a trivial hurdle, as the average number of available detectors being around 250 (Fig.~\ref{fig:29_averages}), with peaks of up to 2620. The experiments are designed to capture data scarcity conditions in different settings, as LDs are randomly sampled and the order of the cities is modified across 50 experiments (which ensures that different cities are considered as "data scarce" across experiments). 
Starting from a subset of 75 LDs, MAML accomplishes so (an extract of the results is provided in Fig.~\ref{fig:MAML75performance}). 
As mentioned in Section~\ref{sec:methodology}, MFD parameters and results have been normalized. For the MSE, the meta-model trained on 75 detector datasets has the lowest median and the narrowest distribution (as it can be seen in Fig.~\ref{fig:msecomparison} and the values reported in Table~\ref{tab:numericalresults}), indicating that it performs the best and is consistent in predictions, to a level rivaling the performance of the bi-parabolic model, as shown in Fig.~\ref{fig:msecomparison}.

Similar results are achieved for 50, 25 and 10 LDs, the full normalized set is reported in the Github repository at https://github.com/s184227/MAML\_for\_MFD. Fig.~\ref{fig:MAMLotherperf} shows how the MAML performs on the most challenging of scenarios, with only 10 LDs per city.

\begin{longtable}[h]{|p{1.8cm}|p{1.2cm}|p{1.2cm}|p{1.2cm}|p{1.2cm}|p{1.2cm}|p{1.2cm}|p{1.2cm}|p{1.2cm}|}\caption{Predicted critical occupancy and maximum flow by the different models.}\label{tab:numericalresults} \\


        \hline

        \textbf{City} & \textbf{Bipara\-bolic \newline CO} & \textbf{BiPara\-bolic \newline Max Flow} &  \textbf{MT\-PINN CO} & \textbf{MT\-PINN Max Flow} & \textbf{MAML Test (75) CO} &  \textbf{MAML Test (75) Max Flow} &  \textbf{MAML Test (10) CO} &  \textbf{MAML Test (10) Max Flow}
         \\
    \hline
Augsburg & 0,768 & 0,965 & 0,887 & 0,974 & 0,921 & 0,907 & 0,887 & 0,919 \\ \hline
        Bern & 0,921 & 0,975 & 0,950 & 0,983 & 0,936 & 0,905 & 0,915 & 0,895 \\ \hline
        Bordeaux & 0,879 & 0,983 & 0,806 & 0,978 & 0,922 & 0,921 & 0,920 & 0,887 \\ \hline
        Bremen & 0,919 & 0,962 & 0,950 & 0,988 & 0,924 & 0,911 & 0,886 & 0,867 \\ \hline
        Darmstadt & 0,724 & 0,960 & 0,884 & 0,952 & 0,911 & 0,902 & 0,922 & 0,886 \\ \hline
        Graz & 0,891 & 0,972 & 0,921 & 0,974 & 0,938 & 0,912 & 0,925 & 0,889 \\ \hline
        Hamburg & 0,701 & 0,964 & 0,833 & 0,945 & 0,930 & 0,903 & 0,925 & 0,891 \\ \hline
        Kassel & 0,858 & 0,972 & 0,950 & 0,960 & 0,945 & 0,909 & 0,898 & 0,902 \\ \hline
        London & 0,852 & 0,979 & 0,897 & 0,973 & 0,941 & 0,905 & 0,934 & 0,894 \\ \hline
        Losangeles & 0,913 & 0,983 & 0,794 & 0,966 & 0,910 & 0,923 & 0,925 & 0,907 \\ \hline
        Madrid & 0,825 & 0,983 & 0,858 & 0,978 & 0,941 & 0,913 & 0,941 & 0,899 \\ \hline
        Marseille & 0,698 & 0,972 & 0,688 & 0,960 & 0,935 & 0,902 & 0,926 & 0,906 \\ \hline
        Santander & 0,790 & 0,962 & 0,950 & 0,954 & 0,935 & 0,897 & 0,919 & 0,896 \\ \hline
        Speyer & 0,925 & 0,974 & 0,950 & 0,983 & 0,907 & 0,905 & 0,914 & 0,908 \\ \hline
        Strasbourg & 0,566 & 0,979 & 0,568 & 0,969 & 0,922 & 0,910 & 0,886 & 0,885 \\ \hline
        Stuttgart & 0,553 & 0,979 & 0,628 & 0,955 & 0,940 & 0,917 & 0,926 & 0,892 \\ \hline
        Taipeh & 0,797 & 0,978 & 0,942 & 0,989 & 0,936 & 0,894 & 0,924 & 0,879 \\ \hline
        Toronto & 0,486 & 0,965 & 0,617 & 0,918 & 0,922 & 0,911 & 0,922 & 0,901 \\ \hline
        Toulouse & 0,758 & 0,987 & 0,748 & 0,985 & 0,916 & 0,918 & 0,906 & 0,904 \\ \hline
        Zurich & 0,785 & 0,977 & 0,848 & 0,974 & 0,914 & 0,918 & 0,914 & 0,879 \\ \hline
\end{longtable}

The metamodels trained on the 50, 25, and 10 detectors dataset have very similar means and distributions. Still, as it could be expected, the mean and variability increase slightly as the MFDs become more biased, and the outliers become more spread out. This indicates that the model’s predictive performance decreases slowly the more biased the data it was trained on.
Still, it is worth reporting the average, not normalized results to appreciate the improved performance of the MAML. We report these values in Table~\ref{tab:moeavg}. For example, it can be seen how, with 75 LDs, the MAE calculated over the average predicted flow for MAML is of 30.34 averaged across cities, against a MAE of 60.34 for the MTPINN. Since averaging the MAE across cities also flattens its values in a way that may hide bigger errors, we report the MAE for each city in the Appendix (Table~\ref{tab:maepercity}). As mentioned in Section~\ref{sec:MAMLexp}, the experiments have been carried out in a similar setting also for the MTPINN, so that the performance could be compared. From the results, it appears that the MTPINN is not able on its own to predict the shape of the MFD with 75 LDs as input, let alone 10 LDs. As reported in Table~\ref{tab:moeavg}, the MSE and RSPP in these cases are equal to 6139.60 and 0.4 or to 11490.71 and 0.57 for 75 and 10 LDs respectively, averaged across cities. 

\begin{longtable}[]{|l|l|l|l|l|l|}\caption{MSE, MAE, RRSE and correlation averaged across cities} \label{tab:moeavg} \\

\hline
             & Mean                & Median              & Max                & Min                & Std                  \\
\hline
MAML MSE 75  & 1767.91   & 1481.87    & 9859.22    & 168.49    & 2058.84\\
\hline
MAML MSE 50  & 1968.18   & 1477.40   & 12365.09      & 222.76   & 2573.69    \\
\hline
MAML MSE 25  & 2340.81  & 1712.91    & 12016.34       & 332.62 & 2502.68     \\
\hline
MAML MSE 10  & 3081.56   & 1901.64    & 11151.25       & 320.59  & 2978.44   \\
\hline
MTPINN MSE 75  & 6139.60  & 4296.33    & 25457.05      & 1091.16  & 5859.19   \\
\hline
MTPINN MSE 50  & 6712.78  & 5395.83        & 25083.73       & 1139.31   & 5927.66   \\
\hline
MTPINN MSE 25  & 6628.45  & 5099.16      & 19378.00       & 1234.94     & 4857.38    \\
\hline
MTPINN MSE 10  & 11490.71   & 11039.83      & 31837.62      & 1717.62     & 8687.88    \\
\hline MAML MAE 75  & 30.34   & 31.16    & 71.87    & 10.41    & 13.35\\
\hline
MAML MAE 50  & 31.91   & 30.97   & 82.66      & 11.67   & 15.11    \\
\hline
MAML MAE 25  & 36.27  & 33.67    & 80.16       & 14.19 & 15.21     \\
\hline
MAML MAE 10  & 41.01   & 36.47    & 94.72       & 13.99  & 19.94   \\
\hline
MTPINN MAE 75  & 60.34  & 55.78    & 137.14      & 24.66  & 29.57   \\
\hline
MTPINN MAE 50  & 63.16  & 61.84        & 134.72       & 25.83   & 30.22   \\
\hline
MTPINN MAE 25  & 65.17  & 57.08      & 117.16       & 27.47     & 25.71    \\
\hline
MTPINN MAE 10  & 85.75   & 86.69      & 157.72      & 34.08     & 36.40    \\
\hline
MAML RRSE 75 & 0.23 & 0.23 & 0.47 & 0.11 & 0.07  \\
\hline
MAML RRSE 50 & 0.23 & 0.23 & 0.49 & 0.12 & 0.07  \\
\hline
MAML RRSE 25 & 0.28 & 0.27  & 0.55 & 0.10 & 0.09   \\
\hline
MAML RRSE 10 & 0.33  & 0.30  & 0.76 & 0.12 & 0.13  \\
\hline
MTPINN RRSE 75 & 0.40         & 0.39 & 0.78          & 0.14         & 0.16  \\
\hline
MTPINN RRSE 50 & 0.40  & 0.35         & 0.73         & 0.19         & 0.15  \\
\hline
MTPINN RRSE 25 & 0.45 & 0.44          & 0.81          & 0.18          & 0.14  \\
\hline
MTPINN RRSE 10 & 0.57  & 0.59  & 1.05          & 0.17          & 0.20   \\
\hline
MAML Corr 75 & 0.98  & 0.98  & 1 & 0.94 & 0.01 \\
\hline
MAML Corr 50 & 0.98   & 0.98  & 1   & 0.94 & 0.01  \\
\hline
MAML Corr 25 & 0.98  & 0.98  & 1  & 0.94  & 0.01  \\
\hline
MAML Corr 10 & 0.98  & 0.98  & 1 & 0.92 & 0.02   \\
\hline
MTPINN Corr 75 & 0.97  & 0.98  & 1 & 0.89 & 0.03  \\
\hline
MTPINN Corr 50 & 0.98  & 0.98  & 1 & 0.88 & 0.02 \\
\hline
MTPINN Corr 25 & 0.97  & 0.98  & 1 & 0.89 & 0.02 \\
\hline
MTPINN Corr 10 & 0.97  & 0.98  & 1 & 0.69 & 0.03 \\
\hline
\end{longtable}

The medians of the MAE and MSE errors for the MTPINN comparison model are significantly higher than the ones obtained with MAML. The variance is also larger, indicating a poorer and less consistent performance. Interestingly, the performance of the meta-models and the MTPINN models are very similar when measured with the RRSE for a given dataset. The mean is almost identical for the two models on the 75 and 50 detector datasets, although the variance is larger for the 50 detector dataset. Overall, an increasing variance is observed, indicating a less consistent performance in the more biased dataset. Still, there is no clear trend in the median before the 10 detector dataset, where both models had poorer performance. The mean of the correlation coefficients is 0.98 or higher for both models for the 75, 50, and 25 detector datasets, and the variance in performance is more significant for the MTPINN comparison model, indicating a somewhat inferior performance.

\begin{figure*}[h!]
    \centering
    \includegraphics[width=0.44\linewidth]{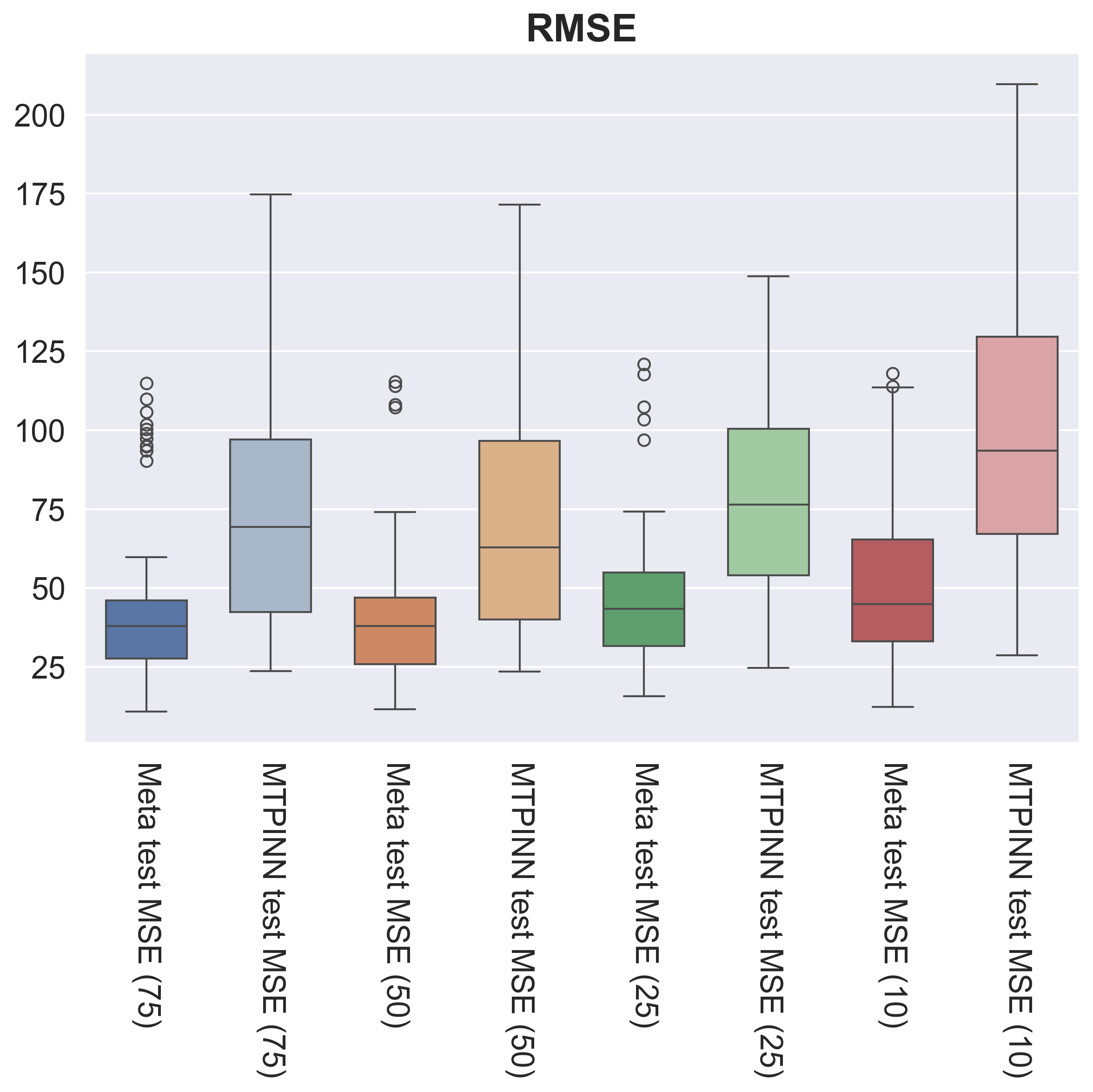}
    \includegraphics[width=0.45\linewidth]{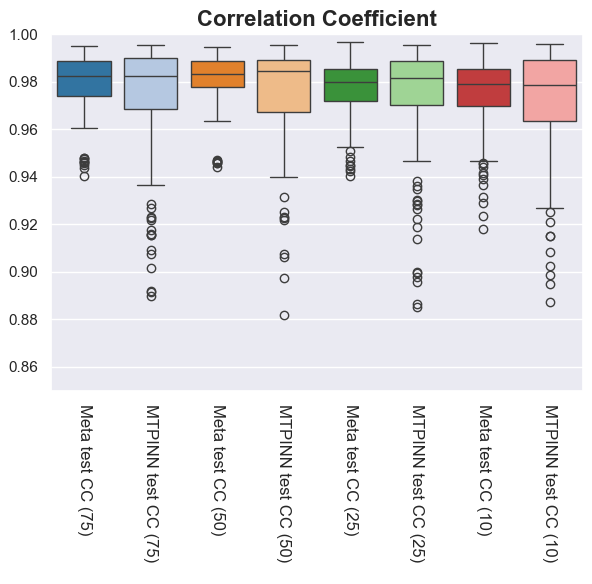}
    \caption{Performance metrics for the meta-model trained and tested on different types of datasets and the MTPINN comparison model.}
    \label{fig:meta-test-mse-rrse-cc}
\end{figure*}

Results from Fig.~\ref{fig:meta-test-mse-rrse-cc} clearly show that the error (RMSE) obtained using the MAML is systematically lower than the one using MTPINN, for the same set of detectors. The correlation coefficients, showing the correlation between estimated and target average occupancy and flow values, are also higher. Target in this case refers to the MFD shape, characterized through the bi-parabolic hybrid model, that would be estimated using all available detectors. This suggests that MAML successfully facilitates the learning of general MFD properties across the cities, as the pre-trained MAML model provides overall better predictions for unseen cities with limited detector data. 

The benefits arising from the MAML framework are particularly evident when evaluating the maximum flow prediction across experiments. As shown in Table~\ref{tab:maxflow}, MAML allows the MTPINN model to maintain a robust estimation (MAML 75 and MAML 10 are both very close to the biparabolic benchmark). The MTPINN model as standalone does instead suffer from a degradation of performance. This means errors of up to 289 vehicles per time interval for the standalone MTPINN and 10 LDs (Marseille) while the MAML shows an error of 48 vehicles per time interval for the same number of LDs. Even with 75 LDs, the error may reach values of 174 for standalone MTPINN versus 83 (Los Angeles). 

\begin{longtable}[]{|l|l|l|l|l|l|}
\caption{Denormalized values of maximum estimated flow, averaged across experiments, for the biparabolic benchmark and the extreme cases with 75 and 10 LDs} \label{tab:maxflow} \\
\hline
\textbf{City} & \footnotesize\textbf{\begin{tabular}[c]{@{}l@{}}Biparabolic\end{tabular}} & \footnotesize\textbf{\begin{tabular}[c]{@{}l@{}}MAML \\ 75\end{tabular}} & \footnotesize\textbf{\begin{tabular}[c]{@{}l@{}}MTPINN \\ 75\end{tabular}} & \footnotesize\textbf{\begin{tabular}[c]{@{}l@{}}MAML \\ 10\end{tabular}} & \footnotesize\textbf{\begin{tabular}[c]{@{}l@{}}MTPINN \\ 10\end{tabular}} \\  \hline
Augsburg      & 572.16               & 542.20           & 458.00             & 535.81           & 456.42             \\ \hline
Bern          & 473.41               & 433.15           & 377.21             & 438.63           & 295.81             \\ \hline
Bordeaux      & 727.62               & 679.23           & 628.66             & 668.40           & 520.41             \\ \hline
Bremen        & 521.83               & 495.74           & 447.79             & 491.36           & 396.12             \\ \hline
Darmstadt     & 453.36               & 430.61           & 366.35             & 427.02           & 231.54             \\ \hline
Graz          & 505.72               & 475.71           & 445.75             & 468.67           & 423.13             \\ \hline
Hamburg       & 385.53               & 362.88           & 265.43             & 362.28           & 193.56             \\ \hline
Kassel        & 436.41               & 413.25           & 388.90             & 402.47           & 364.19             \\ \hline
London        & 548.06               & 520.25           & 488.95             & 507.09           & 445.06             \\ \hline
Losangeles    & 1374.91              & 1291.77          & 1201.54            & 1302.28          & 1143.57            \\ \hline
Madrid        & 912.75               & 851.35           & 779.08             & 845.79           & 648.40             \\ \hline
Marseille     & 751.31               & 707.76           & 645.89             & 703.06           & 462.36             \\ \hline
Santander     & 834.73               & 780.58           & 712.04             & 785.84           & 580.74             \\ \hline
Speyer        & 432.99               & 405.15           & 357.53             & 402.01           & 293.13             \\ \hline
Strasbourg    & 535.74               & 496.15           & 428.12             & 496.76           & 306.71             \\ \hline
Stuttgart     & 715.56               & 672.52           & 609.52             & 666.84           & 598.09             \\ \hline
Taipeh        & 948.95               & 897.89           & 840.35             & 878.28           & 749.80             \\ \hline
Toronto       & 623.60               & 597.64           & 468.68             & 601.55           & 427.18             \\ \hline
Toulouse      & 873.92               & 821.08           & 772.31             & 825.15           & 721.69             \\ \hline
Zurich        & 374.07               & 348.27           & 323.44             & 353.70           & 259.53             \\ \hline
\end{longtable}

Still, the MFD prediction of MAML in data scarcity is not perfect. First, the predictions accurately capture the shape of the MFD, but not the expected downward trend towards the end of the MFD curve. Besides, the dashed red lines in Fig.~\ref{fig:MAML75performance} indicate the estimated maximum flow and critical occupancy for the cities, and it is noteworthy that, as seen in this figure, the maximum flow is generally slightly underestimated, while the critical occupancy is estimated at a significantly higher value than the reference value estimated by the bi-parabolic model.

Nevertheless, the above limitations are not comparable with the degradation in performance suffered by the single MTPINN and are direct result of the severe LDs shortage the MAML is subject to. It is evident from the results that pre-training the model on tasks from different cities allows the model to generalize and make better predictions when tested on a new city using only a limited amount of observations from the biased MFD for the new city, as compared to a model that has only seen the biased observations 
from the city in question. This is also seen for the models trained on the datasets calculated from 50 and 25 randomly selected detectors. A visual representation of the MFDs in all these scenarios is provided in the Github repository.

\subsection{Application to a transport problem}
\begin{figure*}[h!]
    \centering
    \includegraphics[width=0.8\linewidth]{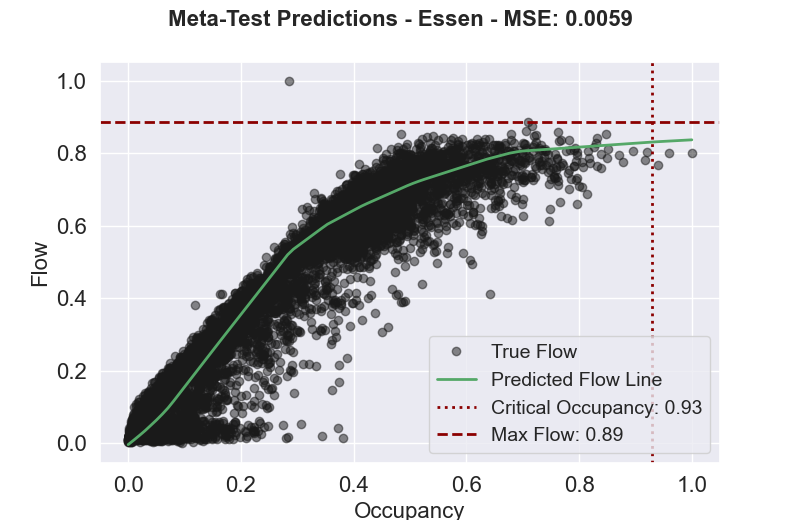}%
    \caption{MFD for the city of Essen, estimated through the MAML+MTPINN framework and the weights calculated with the selected dataset of 20 cities}
    \label{fig:essen_mfd}
\end{figure*}

While the proposed framework has been applied to multiple permutations of the 20 selected cities, the experiments reported so far have been designed to prove how the framework works. Here we instead replicate a transport problem that the MAML framework solves: the estimation of the MFD for the city of Essen. As we detail in section~\ref{sec:data}, only cities with more than 100 LDs have been selected to test the proposed architecture. Essen, with its 36 LDs, has thus been excluded. Still, Essen is the perfect case study for the proposed framework: a medium-sized city with only a handful of LDs available for MFD estimation. By applying the MAML framework over MTPINN, it is possible to initialize the MTPINN with weights $\theta$ that reflect patterns from data-rich cities (i.e., the 20 selected cities) and transfer these patterns to the city of Essen and the measurements from the 36 LDs. The architecture as described in Section~\ref{sec:methodology} is replicated, with only one modification: instead of using 3 cities for the test set, we plug-in Essen and the recording from the 36 LDs. Only 5 iterations (i.e., gradient steps) are carried out by the MTPINN on the Essen data, as it uses weights that have been calculated in Section~\ref{sec:methodology}. These 5 steps allow for the quick adaptation of the MTPINN to the new task and produce an average flow that quite faithfully frame the MFD shape, as shown in Fig.~\ref{fig:essen_mfd}.  

\begin{figure*}[h!]
    \centering
    \includegraphics[width=0.48\linewidth]{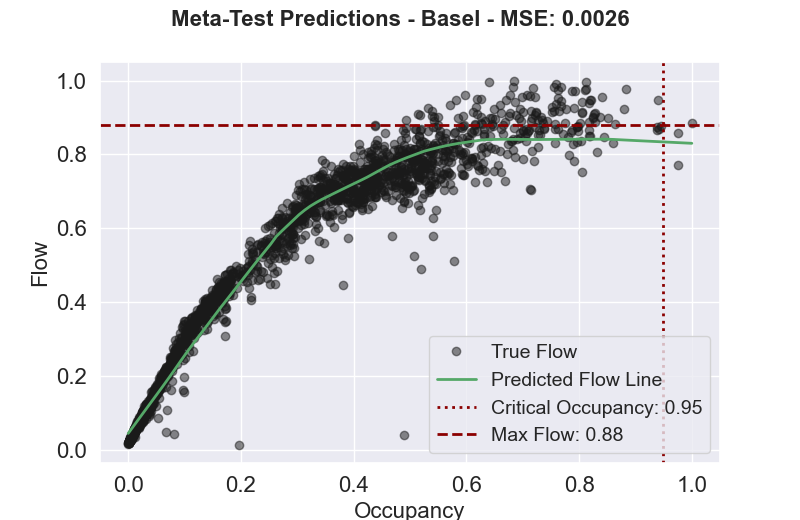} \includegraphics[width=0.48\linewidth]{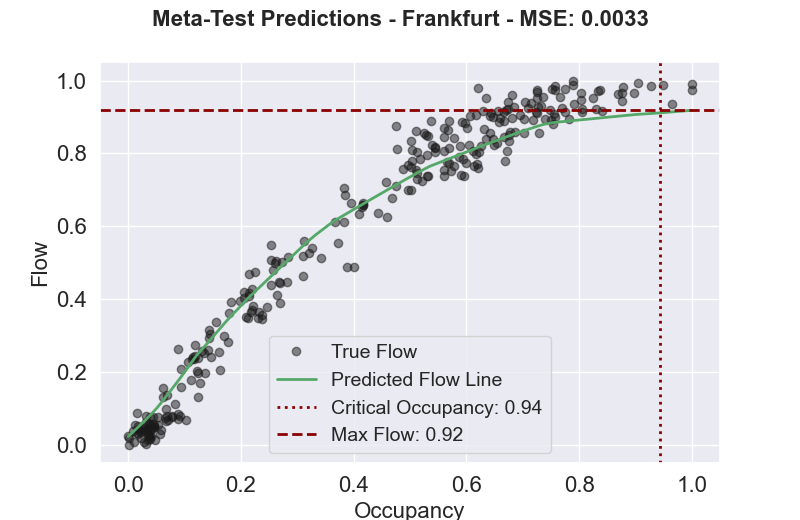}%
    \caption{MFD for the cities of Basel (left) and Frankfurt (right)}
    \label{fig:basel&frankfurt}
\end{figure*}

As it can be noticed, in this case we use all the available data from the 39 LDs to simulate a real world application. The (normalized) MSE of 0.0059 is barely outside the distribution in Fig.~\ref{fig:msecomparison} which is to be expected, as the number of detectors in Essen is smaller than 75. In Fig.~\ref{fig:basel&frankfurt} the same experiment and weights $\theta$ are applied on Basel and Frankfurt, which have respectively 50 and 73 LDs. It is worth noting that Frankfurt has a very short recording interval (one single day), which was another reason why it was kept out of the main experiment with 20 cities. This makes Frankfurt especially challenging, even if theoretically it has a number of LDs comparable to the 75 LD threshold. 

With these cases, we are hypothesizing a situation in which the municipality applies the weights resulting from Meta-Learning to their data to enhance its predicting power. Theoretically, the municipality would not even need to re-train the MTPINN model, as the weights computed in Section~\ref{sec:biasedLD} can be found at https://github.com/s184227/MAML\_for\_MFD.

\section{Model Validation}\label{sec:validation}

In the previous section, we described the results of MAML and MTPINN from a quantitative and qualitative point of view. We show that the physics-informed structure of the model allows for learning the critical occupancy, as well as learn both the congested and uncongested branch of the model, even with limited information. Results further show the capability of MAML to exploit multiple data souces to better generalize. However, other studies proposed fitting methods for the MFD. Hence, this section is divided into two parts. First, we show that MAML can be combined with other statistical models, and that the generalization properties are not limited to MTPINN. Then, we analyse the results using Transfer Learning - an alternative method to MAML for dealing with data scarcity and multiple data sources - and the impact the physics-informed domain knowledge has on model outputs and interpretability. In the rest of this section, unless specified, models are runned over 75, 50, 25 and 10 LDs, with the same hyperparameters described in Table~\ref{tab:hyperparamsMAML}.

\subsection{MAML validation with other models: FitFun.}\label{sec:FunFit}

MAML is tested now in combination with FitFun \citep{bramich2023fitfun}, a model proposed to fit the FD based on the family of statistical models called Generalized Additive Models for Location, Scale, and Shape (GAMLSS). While many models (including MTPINN) use some variation of the least square fitting method, which assumes a symmetric noise distribution, FitFun breaks this assumption - often not reflected in empirical observations \citep{bramich2022fitting, bramich2023fitfun} - by assuming asymmetric noise. In \cite{bramich2023fitfun}, more than 200 different models were tested. The author concluded that the functional form proposed in \cite{sun2014data} combined with a Skew Exponential Power III noise component provided the best results. In this paper we refer to this model as SN2014SEP3, defined as follow:

\begin{align}
    f(x)=_{i} &\stackrel{\text{ind}}{\sim} \mathcal{SEP}_3 \big(\mu(x_i), \sigma(x_i), v(x_i), \tau(x_i)\big) \\
    \mu(x_i) &= q(x_i) \\
    \ln\sigma(x_i)&=S_{nat,\sigma}(x_i) \\
    \ln v(x_i)&=S_{nat,v}(x_i) \\
    \ln\sigma(x_i)&= a_1 + a_2 x_i
\end{align}

with $f(x)$ is a univariate random variable corresponding to the \textit{i}th observation of the response variable (in our case, the average flow). $\mu(x)$ represents the mean of the Skew Exponential Power III ($\mathcal{SEP}_3$) distribution for a given density value $x$, and is modeled as in \cite{sun2014data}. $\sigma$, $v$, and $\tau$ are the 2nd, 3rd, and 4th parameter of the distribution, and indicate the standard deviation, the skewness, and the kurtosis, respectively. $S_{nat}$ is a cubic spline, and it is implemented with 5 parameters ($S_{nat,\sigma}$) and 3 parameters ($S_{nat,v}$) \citep{bramich2023fitfun}. $a_1$ and $a_2$ are parameters to be estimated. For a detailed review, we refer the interested reader to \cite{bramich2023fitfun}. It should also be noted that SN2014SEP3 is one of the most complex models tested in \cite{bramich2023fitfun}, so MAML could be tested with any of the other 200 models, and similar but worse results should be expected. However, the goal of this section is not to compare SN2014SEP3 and MTPINN, nor to discuss the assumption on the symmetric noise distribution, but rather to show that the proposed MAML, being model agnostic, can be combined with other fitting models, including those that have different functional form or assumptions on the error terms, and still have similar positive effects on the learned solutions. 
\\

\begin{figure*}[h!]
    \centering
    \includegraphics[width=0.33\linewidth]{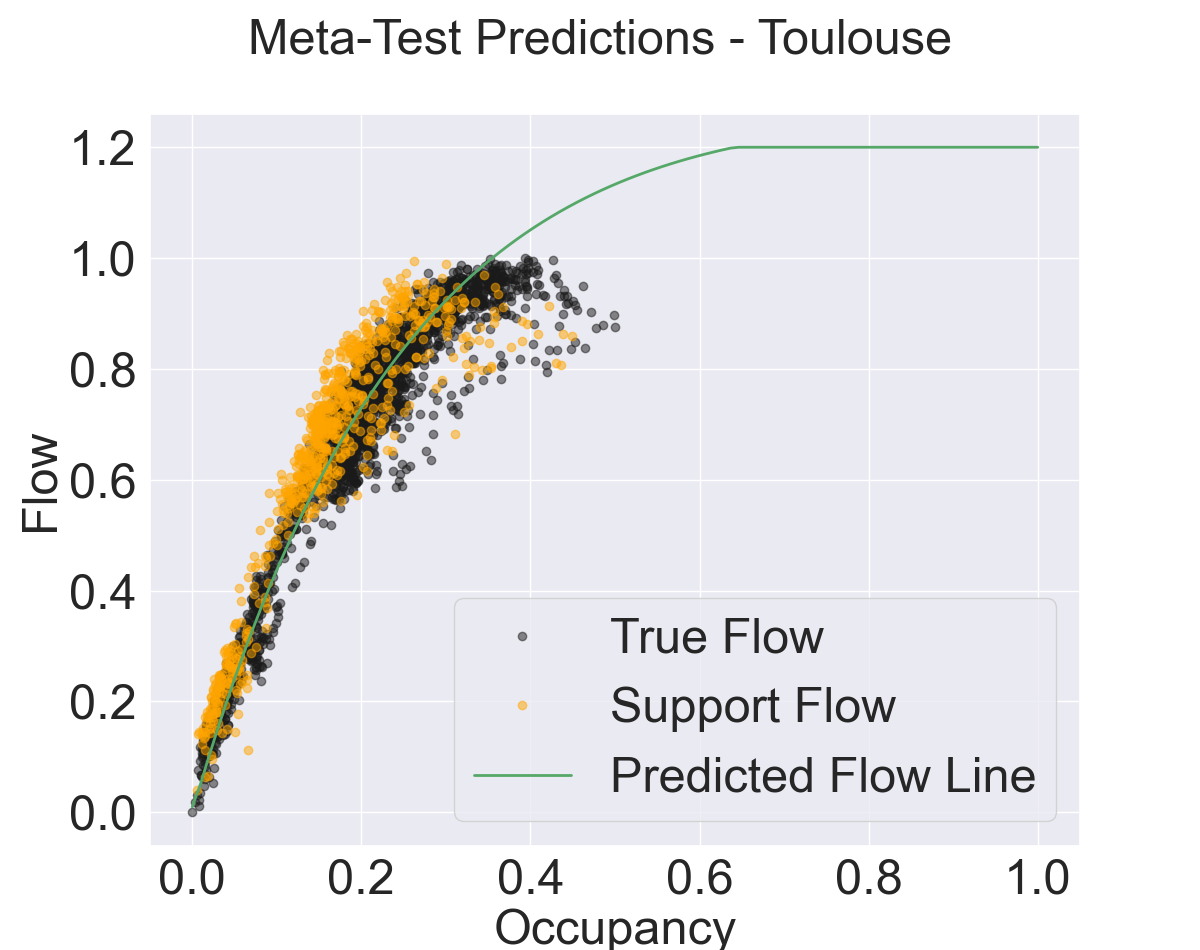}%
    \includegraphics[width=0.33\linewidth]{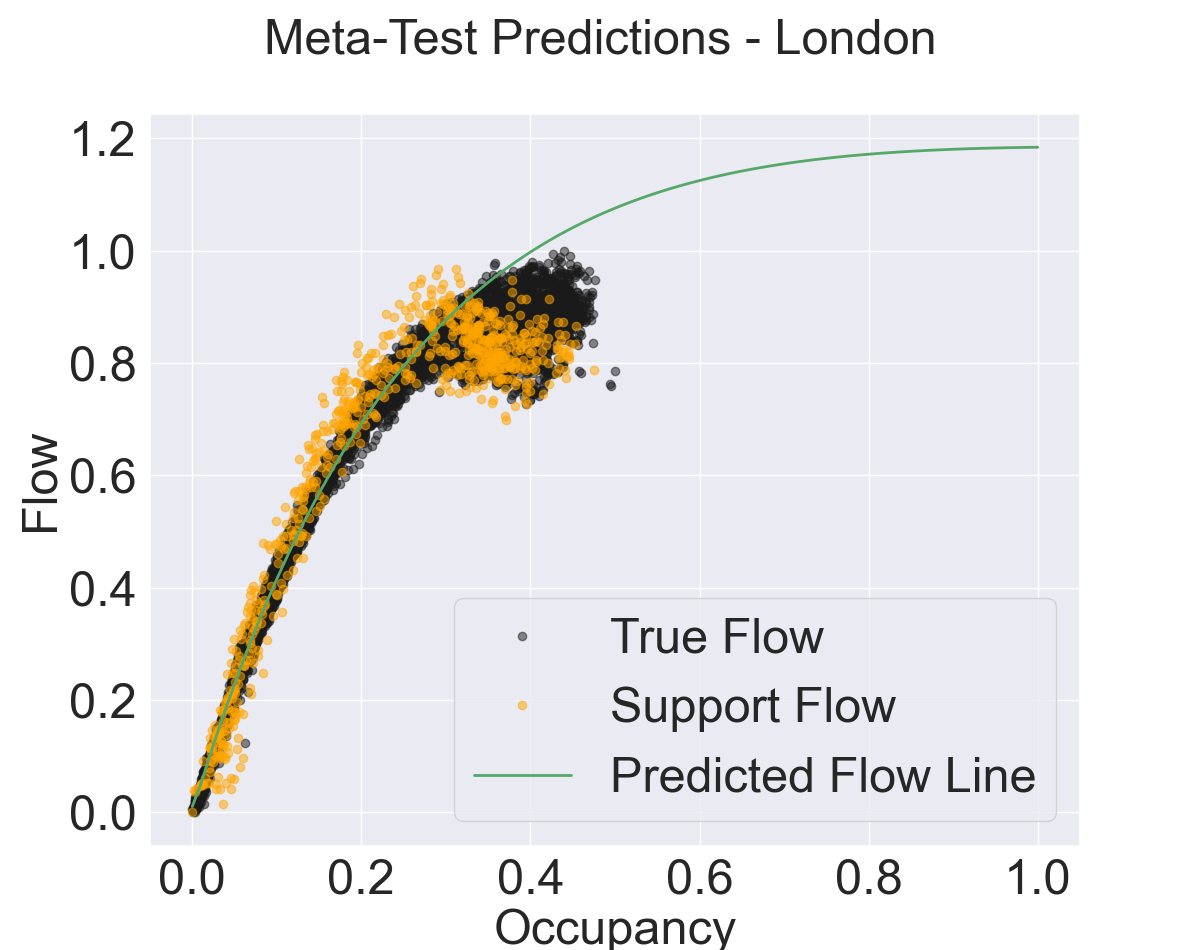}%
    \includegraphics[width=0.33\linewidth]{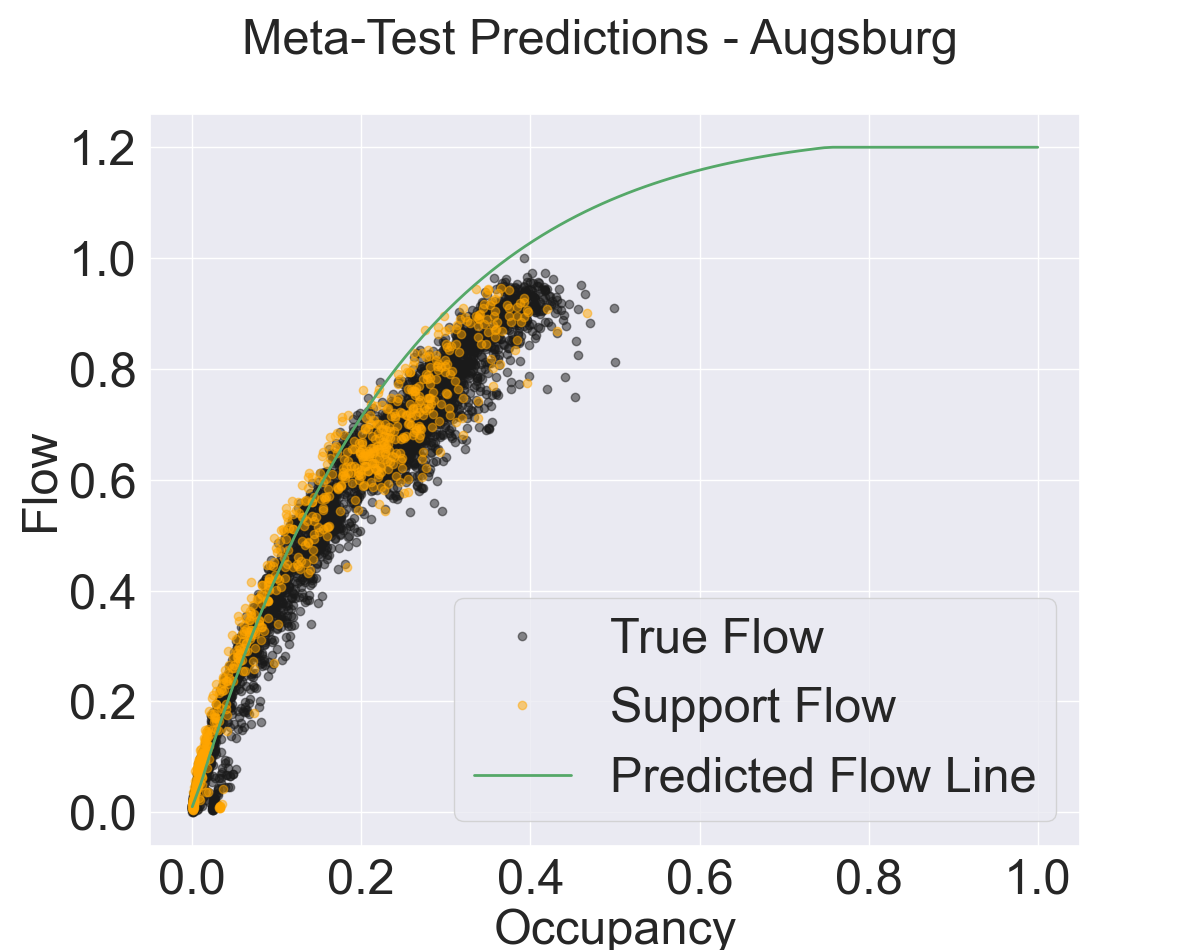}\\
    \includegraphics[width=0.33\linewidth]{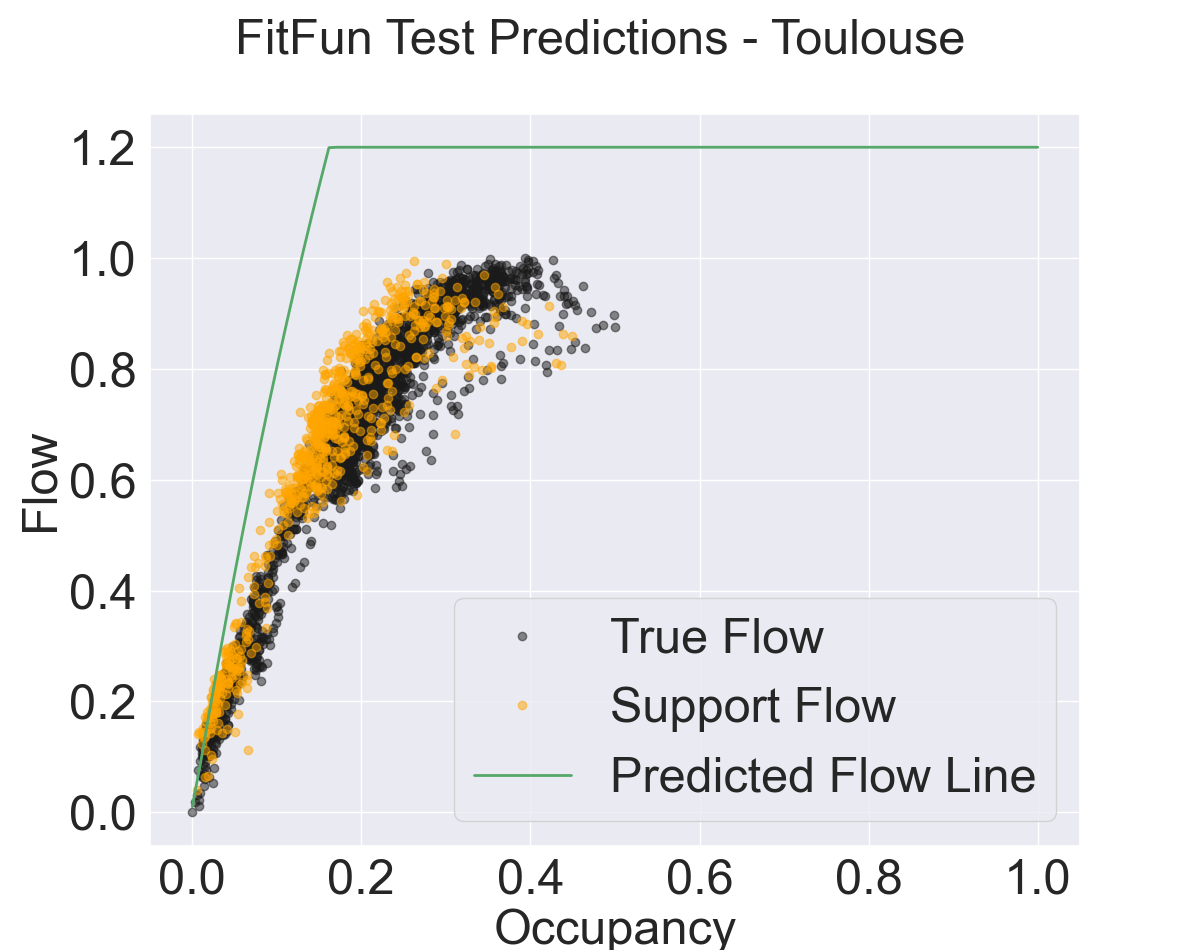}%
    \includegraphics[width=0.33\linewidth]{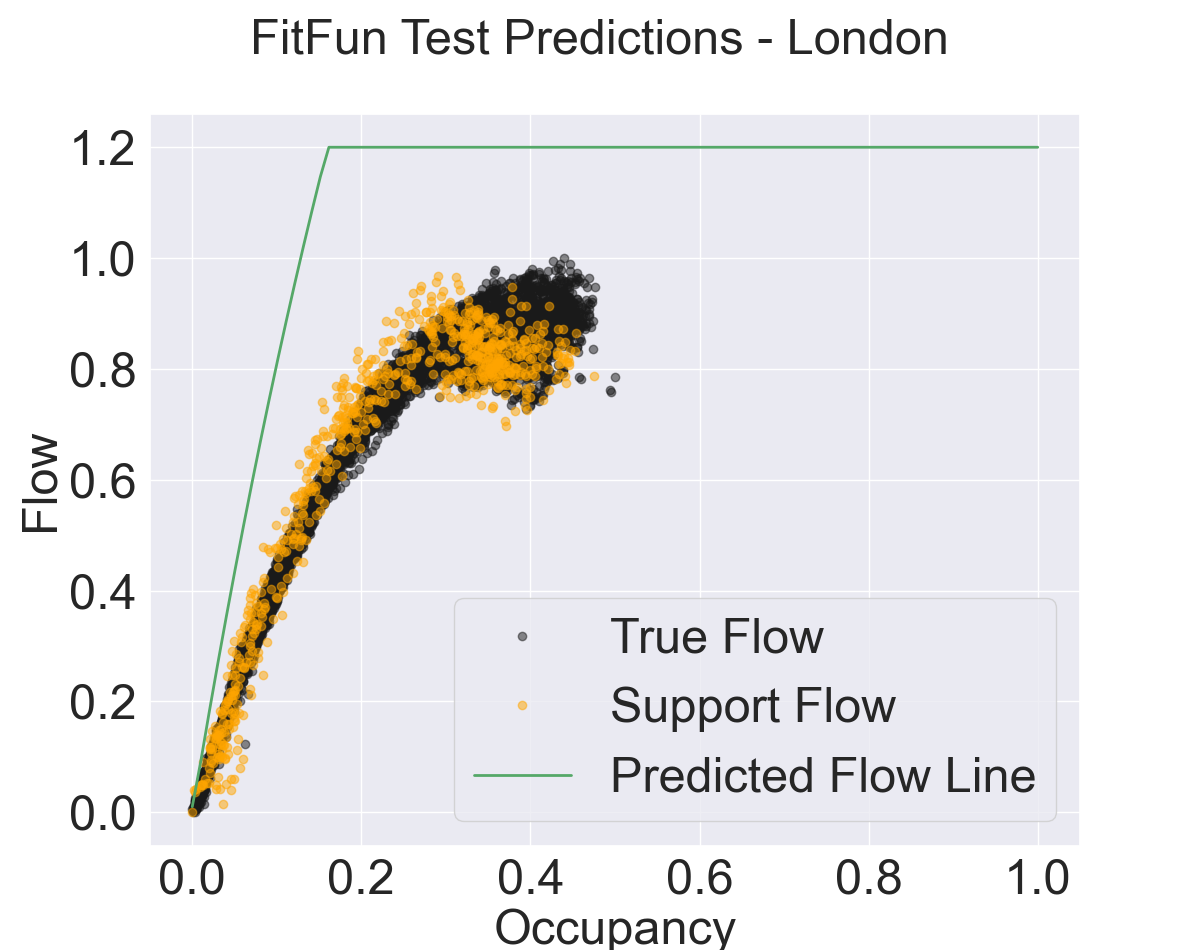}%
    \includegraphics[width=0.33\linewidth]{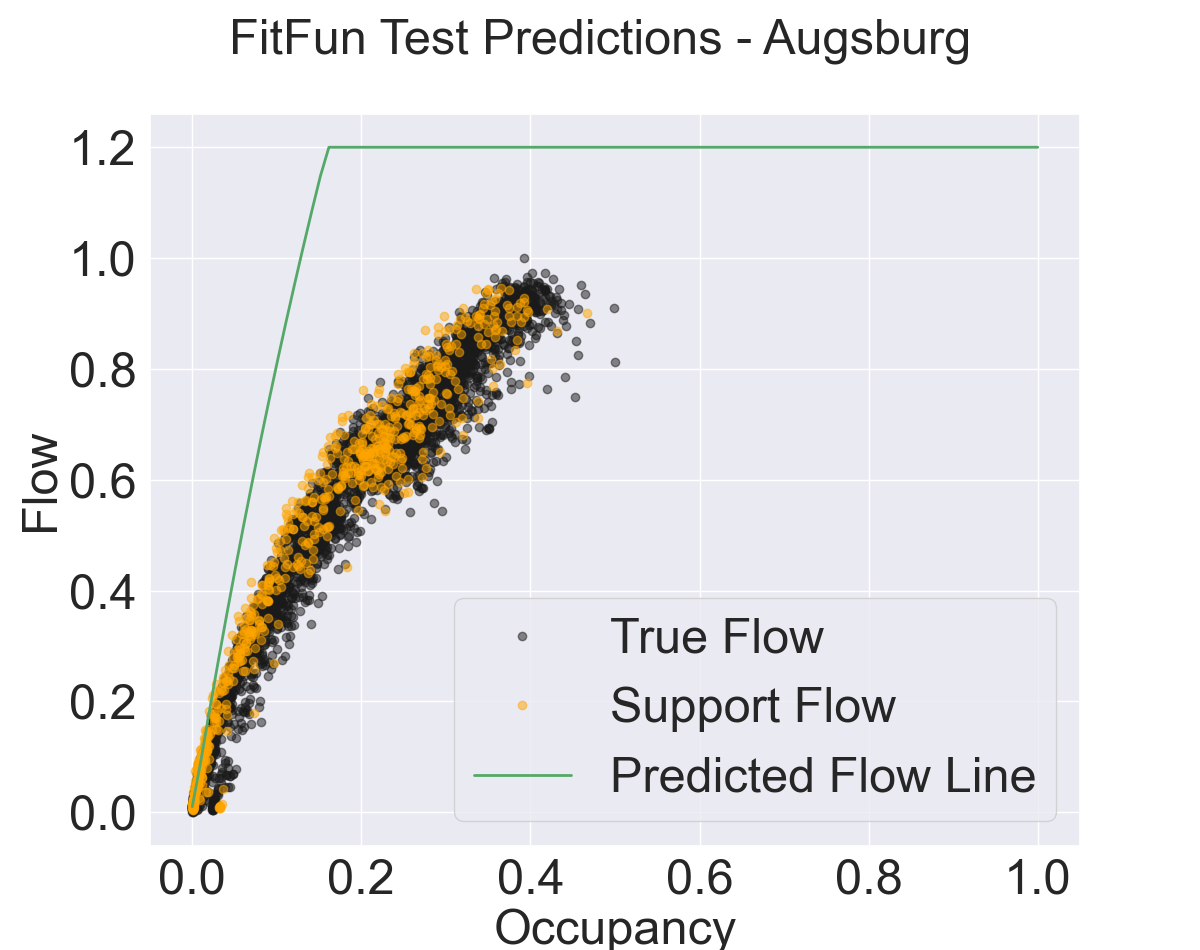}
    \caption{Extract of the results from MFD estimation with 75 LDs. Results with MAML (top row) and FitFun as stand alone (bottom row)}
    \label{fig:ff75lds_samehp}
\end{figure*}

\begin{figure*}[h!]
    \centering
    \includegraphics[width=0.5\linewidth]{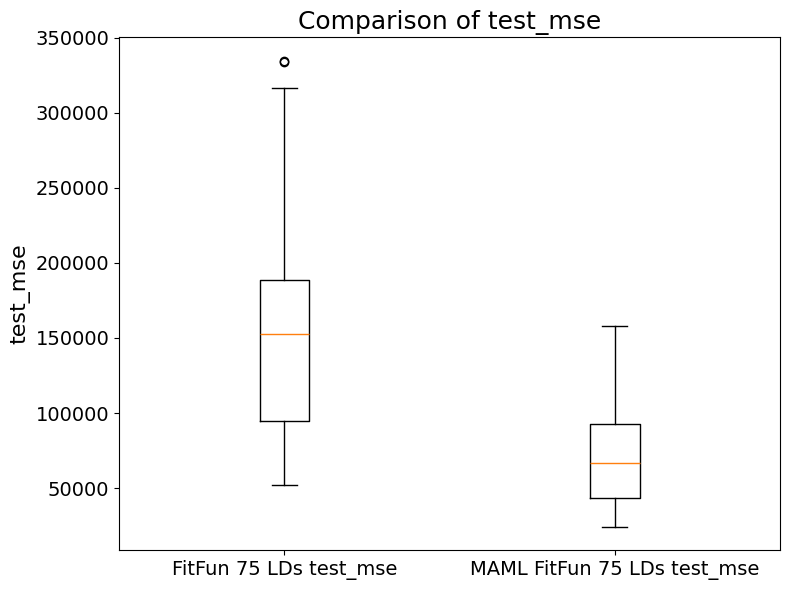}%
    \includegraphics[width=0.5\linewidth]{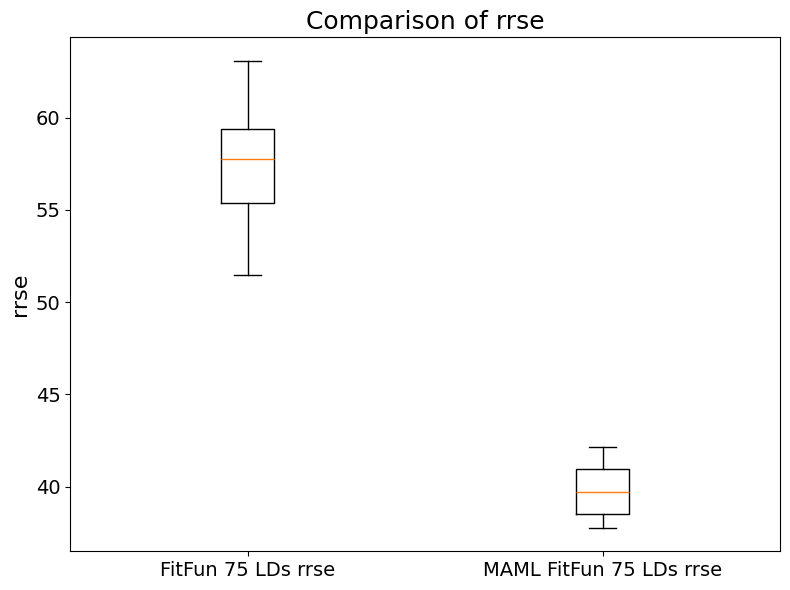}
    \caption{Comparison of the denormalized MSE and RRSE across experiments with 75 LDs}
    \label{fig:mse75lds_samehp}
\end{figure*}

\underline{\textbf{FitFun with starting hyperparameters:}} Fig.~\ref{fig:ff75lds_samehp} and Fig.~\ref{fig:mse75lds_samehp} shows the results for FitFun, with and without MAML. From the visual inspection, we can observe that FitFun faces challenges in capturing the congested branch of the MFD, a problem already observed in the previous section and mostly due to the fact that very few data points are available for the congested branch. Additionally, FitFun without MAML also faces challenges in fitting the uncongested branch in 75 epochs (the number needed when all the LDs are available). This problem disappears when FitFun is embedded within MAML, showing improvements in capturing both the congested and uncongested branch. While results in Fig.~\ref{fig:ff75lds_samehp} only represents three cities, Fig.~\ref{fig:mse75lds_samehp} shows through the MSE and the RRSE that MAML offers significant improvements across the 20 considered cities. 

\underline{\textbf{FitFun with hyperparameter tuning:}} FitFun clearly could not fit the provided data, which is however highly unrealistic considering the results presented in \cite{bramich2023fitfun}. Therefore, we increased the number of epochs to 1000. To match this kind of hyperparameter tuning to the case with 75 LDs, the MAML results reported in this section are run with 300 meta-iteration rather than 150.

\begin{figure*}[h!]
    \centering
    \includegraphics[width=0.33\linewidth]{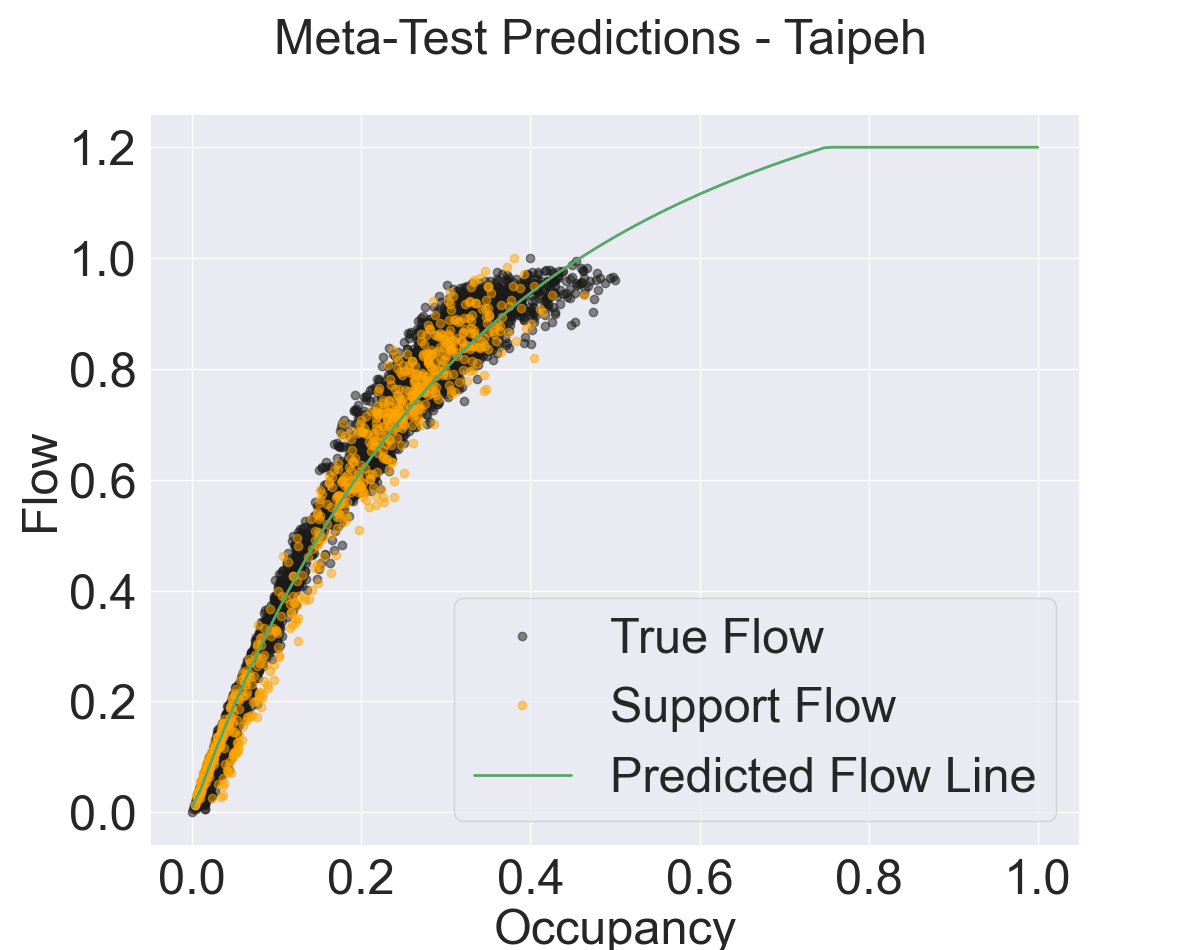}%
    \includegraphics[width=0.33\linewidth]{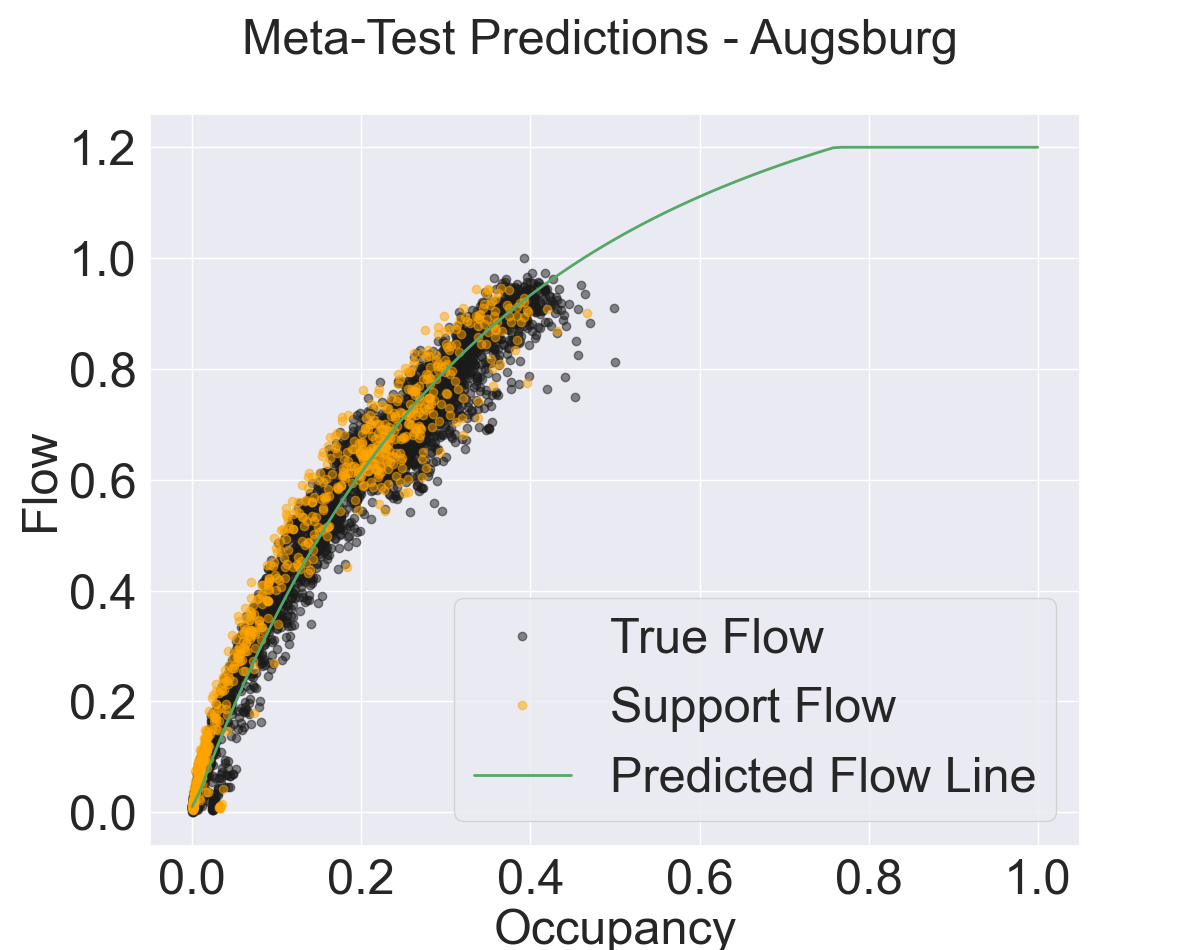}%
    \includegraphics[width=0.33\linewidth]{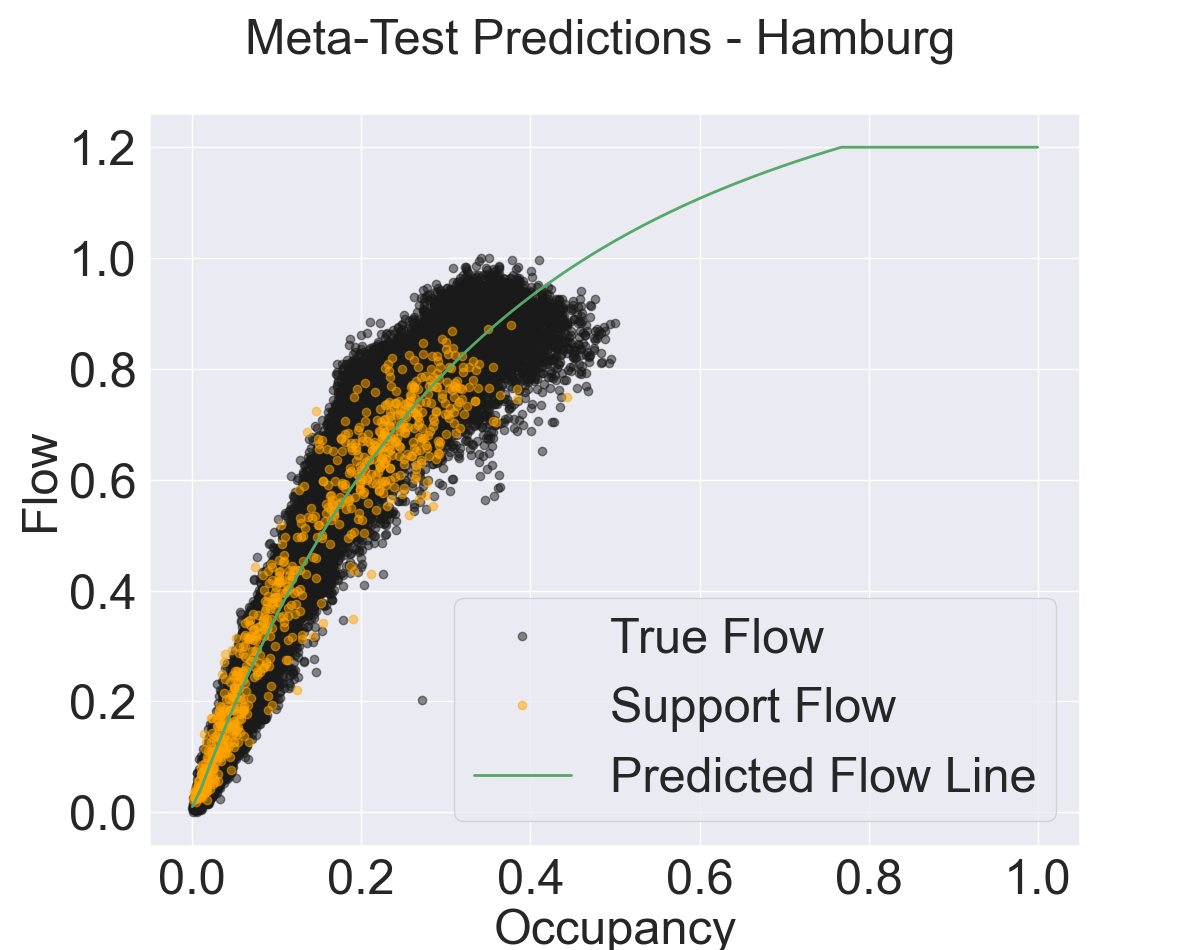}\\
    \includegraphics[width=0.33\linewidth]{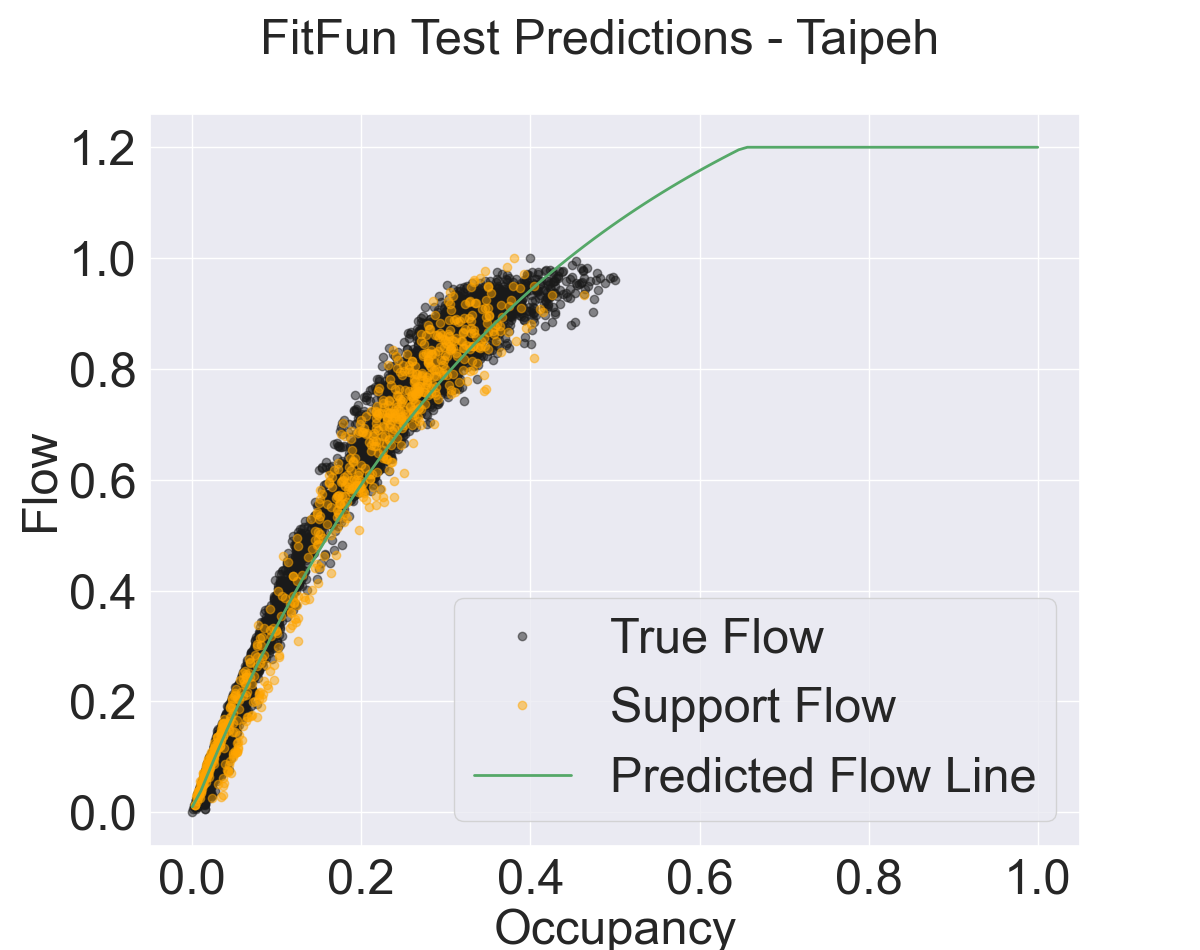}%
    \includegraphics[width=0.33\linewidth]{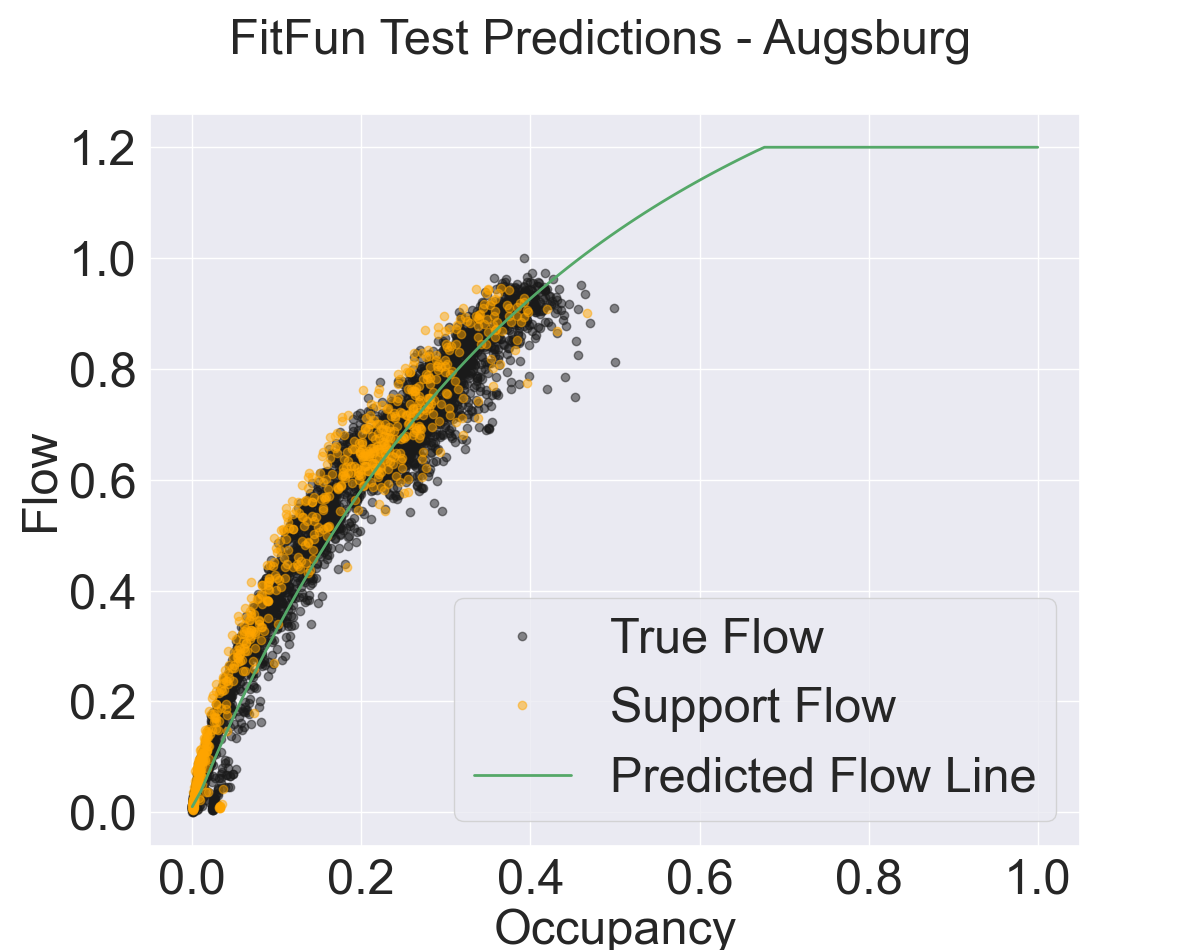}%
    \includegraphics[width=0.33\linewidth]{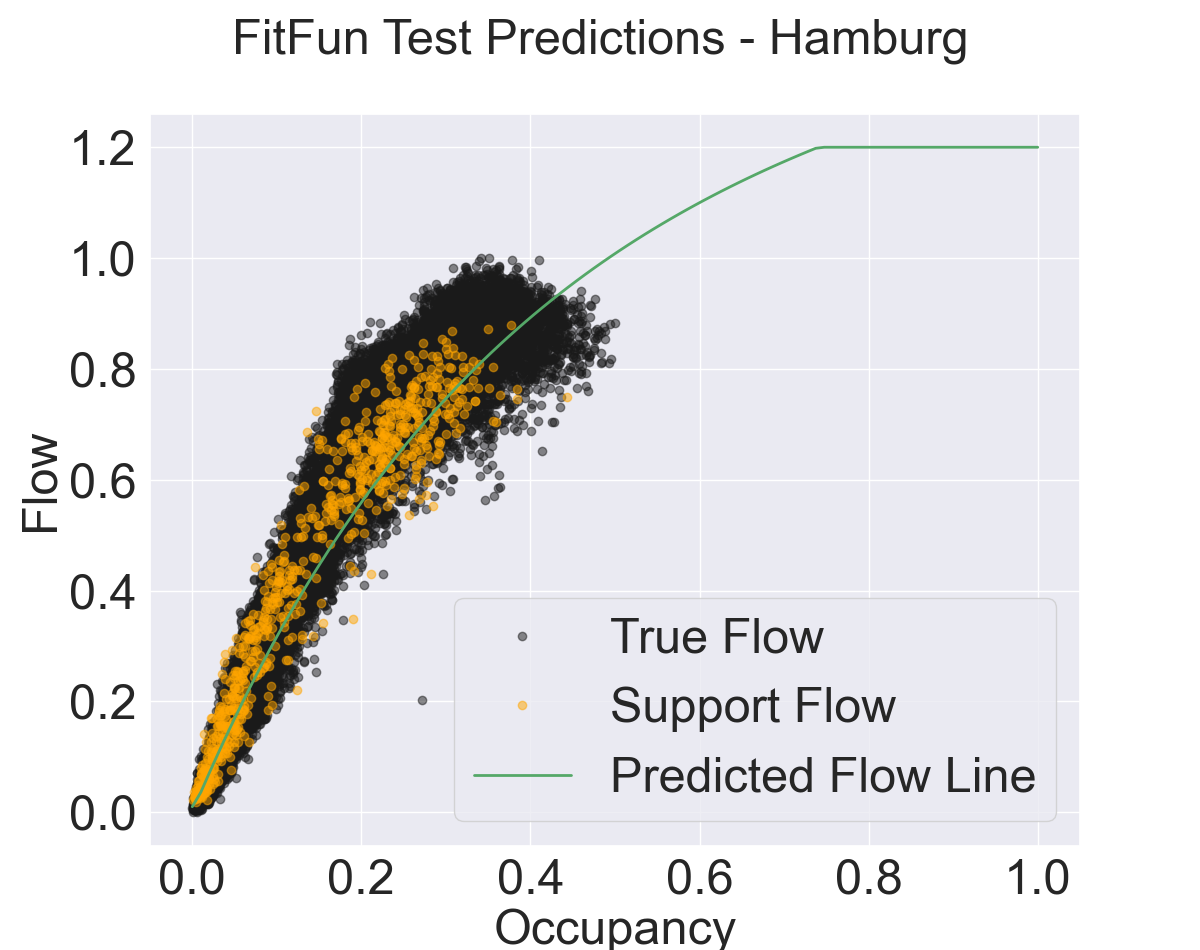}
    \caption{Extract of the results from MFD estimation with 75 LDs and tuned hyperparameters. Results with MAML (top row) and FitFun as stand alone (bottom row)}
    \label{fig:ff75lds_1000epochs}
\end{figure*}

\begin{figure*}[h!]
    \centering
    \includegraphics[width=0.5\linewidth]{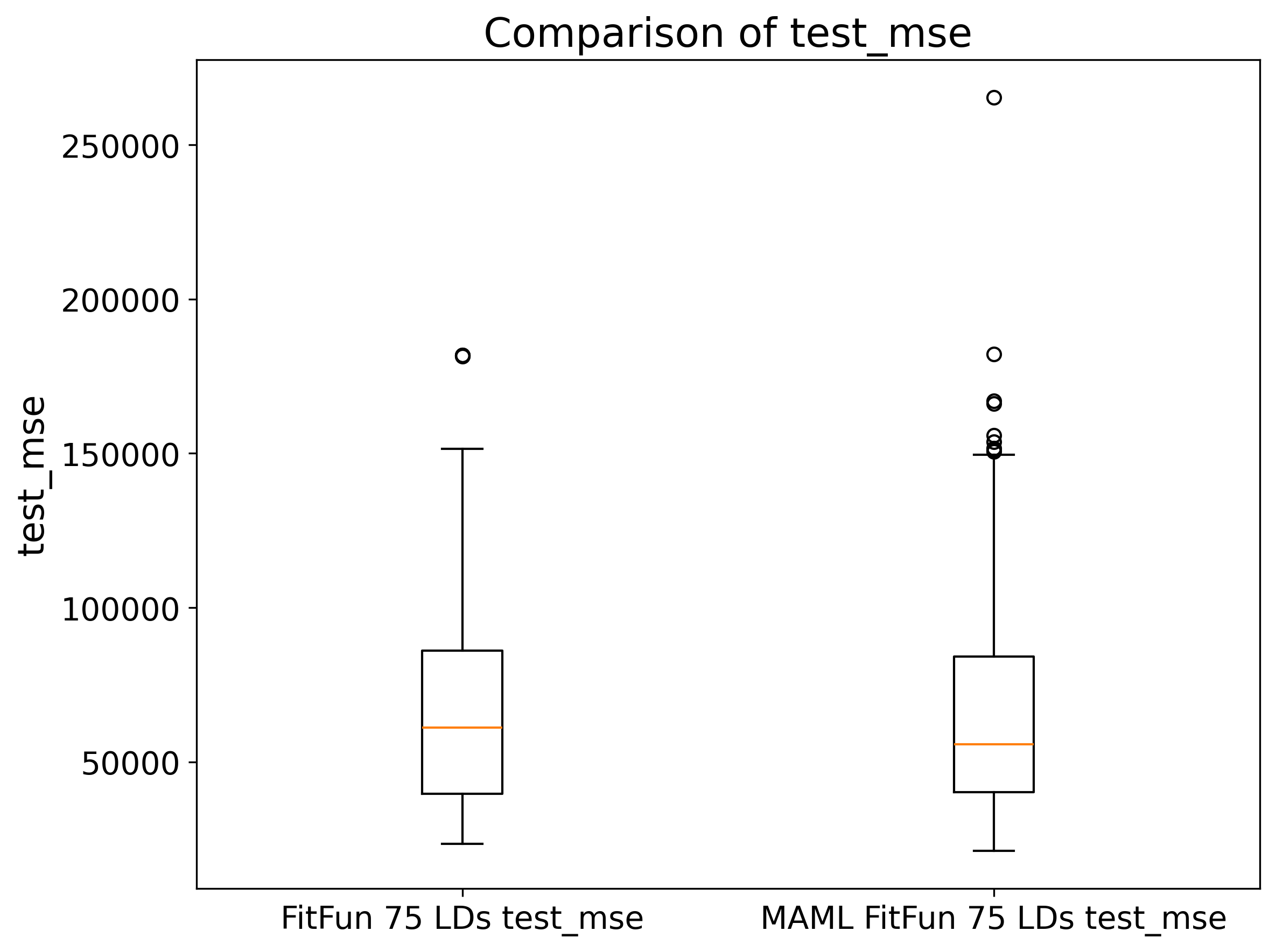}%
    \includegraphics[width=0.5\linewidth]{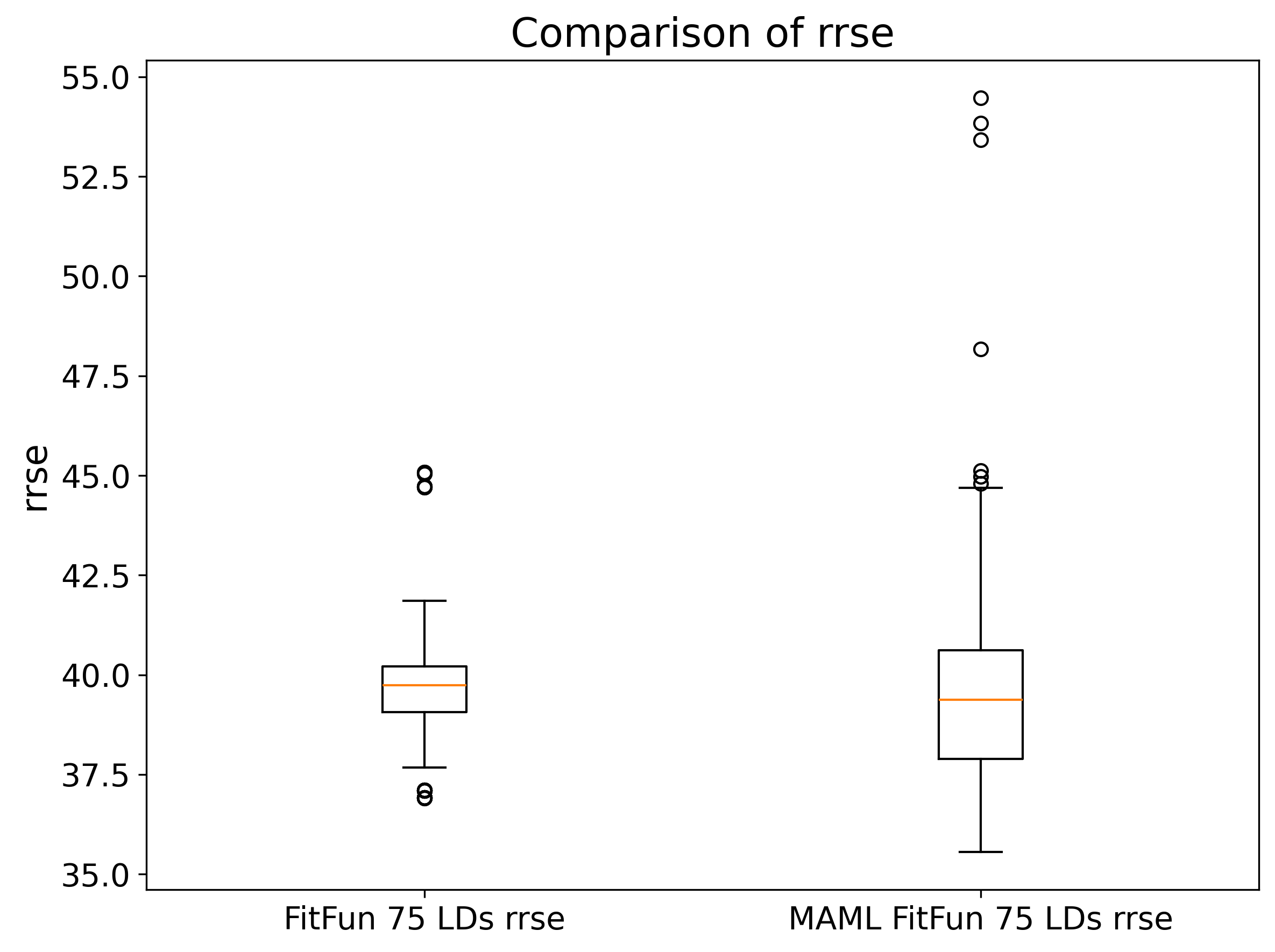}
    \caption{Comparison of the denormalized MSE and RRSE across experiments with 75 LDs and tuned hyperparameters}
    \label{fig:mse75lds_300_1000}
\end{figure*}

Fig.~\ref{fig:ff75lds_1000epochs} shows that after parameter tuning, FitFun is able to fit a more sensible parabolic behavior, closely resembling the one obtained with MAML. Indeed, as shown in Fig.~\ref{fig:mse75lds_300_1000}, the standalone FitFun reaches a similar performance as the MAML FitFun, showing comparable MSE distribution across cities but fewer outliers. Similarly, the mean RRSE is comparable between the two cases but with standalone FitFun showing a lower variance. Still, it is worth stressing how MAML+FitFun achieves these comparable results much faster, as it requires only 5 iterations. Overall, while results confirm that FitFun is a strong model for MFD fitting, they also show how MAML can easily be integrated with different architectures, providing comparable if not better and more robust results in the case of limited data. 

\begin{figure*}[h!]
    \centering
    \includegraphics[width=0.415\linewidth]{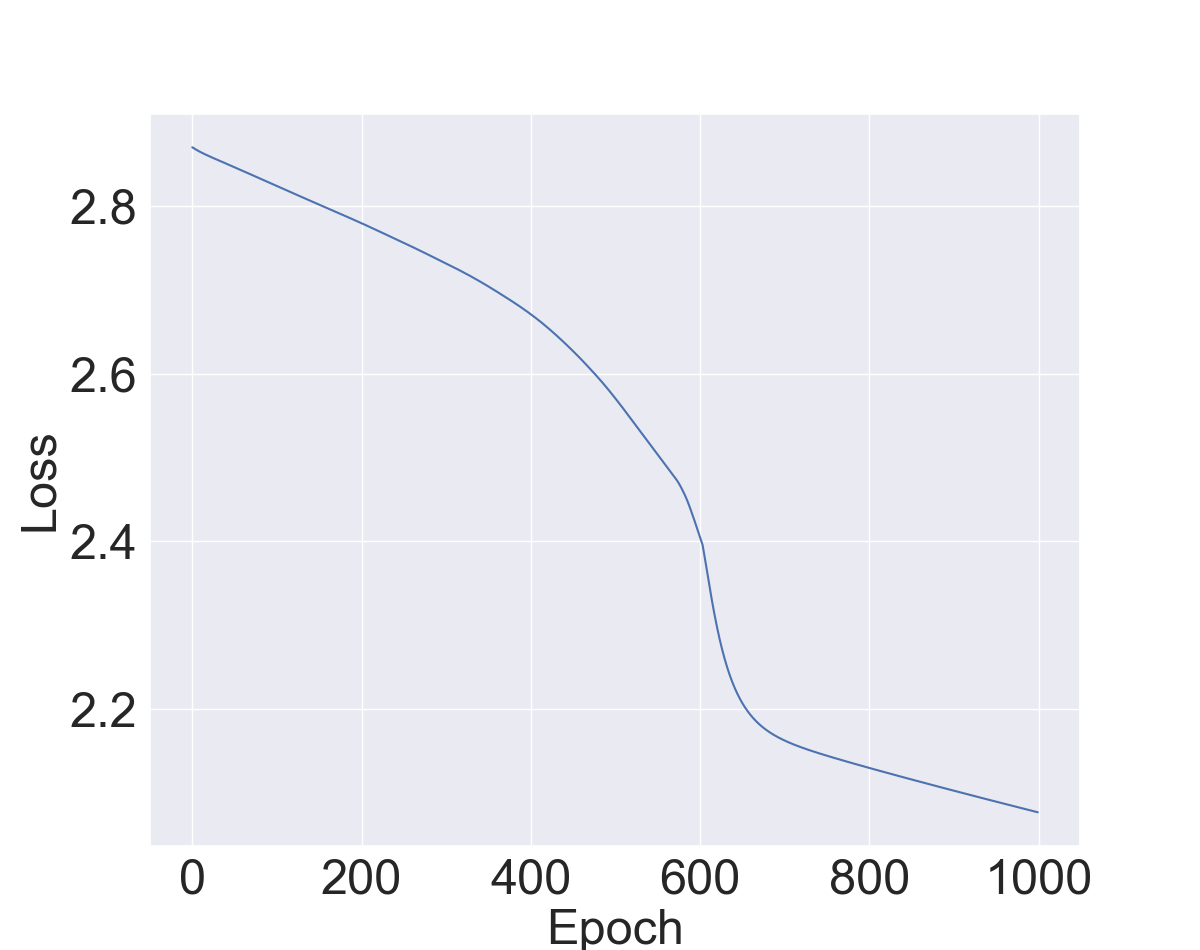} \includegraphics[width=0.4\linewidth]{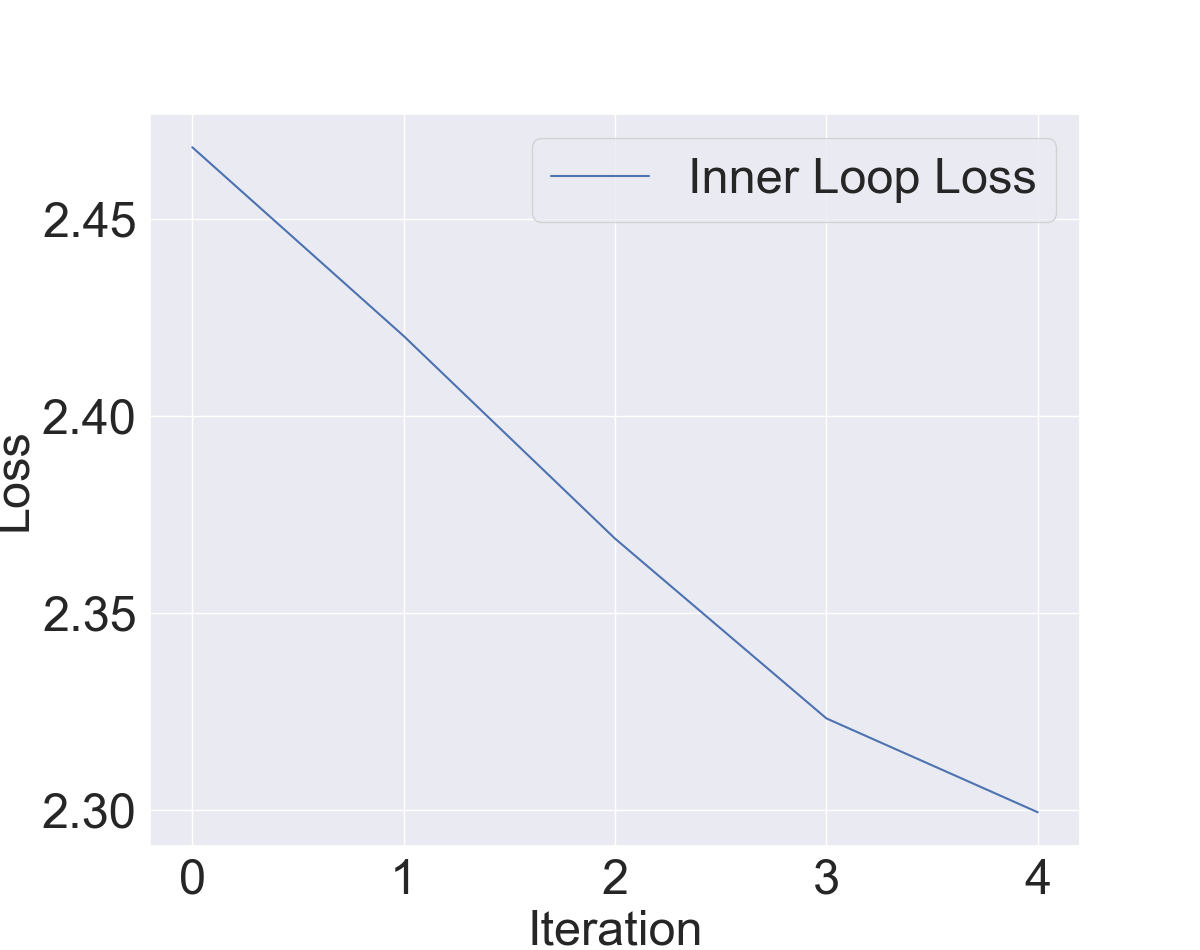}
    \caption{Comparison of the loss gain in the FitFun training (left) and in the MAML inner loop (right)}
    \label{fig:losscomparison_300_1000}
\end{figure*}

Fig.~\ref{fig:losscomparison_300_1000} clearly shows that the inner iteration of the test city start from a much more advantageous starting point but also decrease with a much faster ratio (the standalone FitFun requiring  more than 600 epochs to reach the value of 2.3 loss that MAML FitFun reaches in 5 meta-iterations). While the two hyperparameters have different notation across the paper (epoch versus meta-iteration) they are conceptually analogous as they represent the steps the models take in adjusting their loss and improve the MFD estimation. What emerges from the experiments is that, while given enough steps standalone FitFun and MAML + FitFun can perform similarly, MAML allows the model to start from a much more advantageous starting point. 
The MFD estimated plots from Fig.~\ref{fig:ff75lds_samehp} and Fig.~\ref{fig:ff75lds_1000epochs} for all cities tested in the experiments can be found at https://github.com/s184227/MAML\_for\_MFD.

\subsection{Comparison with Trasfer Learning}

This section compares MAML-MTPINN with a series of benchmark models inspired by the literature. First, the main idea of MAML is to find a good starting point by essentially averaging solutions across tasks. Therefore, we tried to simply train MTPINN on different tasks and obtain the weights by averaging them. This naive approach (which included several re-training) lead to poor performances, which are not reported in this paper. Instead, we show results using Transfer Learning, which was successfully applied in \cite{mahajan2023predicting} in the context of traffic flow prediction, with results pointing out that the model is appropriate when the distribution of data allows for observing the MFD of the network. Specifically, we use a model-based Transfer Learning for domain adaptation \citep{redko2019domain, pan2009survey}, where the model is pre-trained using all data from the support and query set to obtain good initial values. After the training, the shared layers are frozen, while the weights of the task-specific layers are re-trained on the data available in the target city. We tested the following settings: 

\begin{itemize}
    \item Transfer Learning with Cold Start (TC): In this case, weights in the task-specific branches are re-initialized to zero before retraining
    \item Transfer Learning with Warm Start (TW): In this case, weights in the task-specific branches are updated starting the values obtained during training phase on the support set. 
\end{itemize}

Cold Start is often preferred in practice, as pre-estimated parameters may be close to a local optimum, meaning that warm start is not optimal for generalization.  In addition, each method is tested using 5 epochs (TC 5/TW 5) and 1000 epochs (TC 1000/TW 1000). This is to offer a fair comparison with MAML, which only requires 5 iterations of the inner loop. However, Transfer Learning is not necessarily designed for quick adaptation, so we anticipate that more epochs may be necessary. For this reason, TC 5 results are not reported below, as the results systematically showed that  5 epochs are not sufficient when Cold Start is used. Finally, to evaluate the impact of the proposed domain-informed MTPINN method, we also test these methods with a deep Neural Network. The Neural Network (NN) has the same architecture of MTPINN, but it has a single output layer and does not have the physics-informed loss. Therefore, the only output is the average link flow. The results are depicted in Tables \ref{tab:Trasf10-25} and \ref{tab:Trasf50-75}

First, it is observable that MAML-MTPINN is the best performing model for the case with 10 detectors, with 11\% improvement in MSE compared to the second best model (TW-1000-MTPINN) and 17\% improvement compared to MAML-NN. For a larger number of detectors, Transfer Learning becomes more competitive and outperforms MAML, with TC-1000-MTPINN being the best model for the case with 50 and 25 detectors. While the error metrics for most models are close in the case of 25, 50, and 75 detectors, as mentioned, Transfer Learning should be used with a cold start to ensure good generalisation. While TC-1000-NN provides good performances for 50 and 75 detectors, the improvements become marginal for 25 and 10 detectors, with TW-5-NN being the best performing model for the NN for 10 and 25 detectors. This suggests that transfer learning does not work properly in the case of a simple Neural Network. It should also be recalled that solely looking at the MSE is not ideal, as NN may fit the data properly, but completely overlook the congested branch, as previously discussed. In addition, MTPINN provides more interpretable results, including the critical density, which makes it preferable in practice for comparable performances. Nevertheless, we focus in this section only on MSE and normalised RRSE. 

\begin{table}[ht]
\centering
\small
\resizebox{\textwidth}{!}{%
\begin{tabular}{lcccc}
\hline
& \multicolumn{2}{c}{10 Detectors} & \multicolumn{2}{c}{25 Detectors} \\
\cmidrule(lr){2-3} \cmidrule(lr){4-5}
Model & MSE (mean (std)) & RRSE (mean (std)) & MSE (mean (std)) & RRSE (mean (std)) \\
\hline
Neural Network     & 6925.96 (5378.18) & 0.4017 (0.1318) & 2879.58 (2217.52) & 0.2739 (0.0832) \\
MAML NN            & 6089.97 (4776.80)  & 0.3667 (0.1173) & 2519.77 (2584.68) & 0.2450 (0.0696) \\
TC 1000 NN         & 5831.43 (3956.05)  & 0.3738 (0.1127) & 2072.22 (1707.54)  & 0.2353 (0.0777) \\
TW 5 NN            & 5693.06 (3948.61)  & 0.3671 (0.1144) & 2027.68 (1641.25)  & 0.2338 (0.0767) \\
TW 1000 NN         & 5753.61 (3833.67)  & 0.3714 (0.1104) & 2064,21 (1672,25)  & 0,2358 (0,0800) \\
MTPINN             & 18728.16 (14149.30) & 0.6591 (0.2211) & 8337,29 (4269,23) & 0.4926 (0.1357) \\
MAML MTPINN        & \textbf{5053.14} (4463.30)  & \textbf{0.3310} (0.1144) & 2634.95 (2428.46)   & 0.2566 (0.0746) \\
TC 1000 MTPINN     & 5693.60 (3770.01)  & 0.3706 (0.1081) & \textbf{2020.63} (1625.81)  & \textbf{0.2335} (0.0766) \\
TW 5 MTPINN        & 6437.32 (4817.00)  & 0.3894 (0.1277) & 2284.85 (1721.35)  & 0.2504 (0.0855) \\
TW 1000 MTPINN     & 5665.90 (3758.90)  & 0.3697 (0.1083) & 2052.85 (1667.85)  & 0.2356 (0.0788) \\
\hline
\end{tabular}%
}
\caption{Performance metrics at 10 and 25 detectors (MSE and RRSE mean (std)).}
\label{tab:Trasf10-25}
\end{table}

\begin{table}[ht]
\centering
\small
\resizebox{\textwidth}{!}{%
\begin{tabular}{lcccc}
\hline
& \multicolumn{2}{c}{50 Detectors} & \multicolumn{2}{c}{75 Detectors} \\
\cmidrule(lr){2-3} \cmidrule(lr){4-5}
Model & MSE (mean (std)) & RRSE (mean (std)) & MSE (mean (std)) & RRSE (mean (std)) \\
\hline
Neural Network     & 2237.6 (3294.03)   & 0.2381 (0.0825) & 2208.46 (1678.9)  & 0.2532 (0.0747) \\
MAML NN            & 1730.62 (2334.57)   & 0.2240 (0.0616) & 1292.10 (814.64)  & 0.1956 (0.0580) \\
TC 1000 NN         & 1460.56 (1657.92)   & \textbf{0.2141} (0.0646) & \textbf{973.30} (664.65)  & \textbf{0.1718} (0.0545) \\
TW 5 NN            & 1520.67 (1644.14)   & 0.2187 (0.0707) & 1035.02 (721.94)  & 0.1754 (0.0547) \\
TW 1000 NN         & 1529.43 (1641.20)   & 0.2204 (0.0713) & 983.71 (672.13)  & 0.1723 (0.0547) \\
MTPINN             & 5452.38 (6982.01) & 0.3896 (0.1247) & 5972.85 (4503.78)& 0.4131 (0.1227) \\
MAML MTPINN        & 1868.77 (2064.24)   & 0.2467 (0.0593) & 1177.63 (682.23)  & 0.1911 (0.0500) \\
TC 1000 MTPINN     & \textbf{1449.13} (1552.93)   & 0.2147 (0.0638) & 1029.29 (652.52)  & 0.1773 (0.0504) \\
TW 5 MTPINN        & 1690.66 (1646.32)   & 0.2298 (0.0657) & 1314.63 (881.45)  & 0.1967 (0.0559) \\
TW 1000 MTPINN     & 1489.03 (1585.07)   & 0.2173 (0.0695) & 1028.41 (649.07)  & 0.1776 (0.0505) \\
\hline
\end{tabular}%
}
\caption{Performance metrics at 50 and 75 detectors (MSE and RRSE mean (std)).}
\label{tab:Trasf50-75}
\end{table}

Results also clearly point out that MTPINN, compared to its simple NN counterpart, is a model with relatively low variance and high bias. Results show that training MTPINN with MAML brings a reduction of the MSE metrics between 65\% and 80 \%. This shows that the structure with three output branches and the physics-informed loss makes it very difficult to overfit the data. A criticism could be that they make it too difficult, as the model alone has a relatively high error even in the case with 75 detectors. On the other hand, MTPINN exhibits a consistent and reliable behaviour with both MAML and Transfer Learning, with these models both reducing the error more and more when the number of detectors decreases. Although not as performing as MAML, TC-1000-MTPINN shows reliable performances in all experiments and a fairly predictable behaviour. The same cannot be observed in NN.

\section{Limitations and future research directions.}\label{sec:limitations}
The performance of both the MTPINN (with full access to LDs recordings) and of the MAML (with restricted access) are satisfactory for application purposes, as detailed in the previous section. Still, some limitations can be listed. First, only the very early stages of congestion are observed in the UTD19 data. Thus, evaluating the quality of the fit for the second parabola describing the congested regime is very difficult for any non-parametric model, as so little of it is observed. Future work may focus on refining the MFD predictions in the congested half of the MFD, for example by adopting a probabilistic approach to modeling the congested branch of the MFD. 
Besides, from the calibration of the hyperparameters for the MTPINN and the MAML, it became clear that finding a combination that fits all the cities accurately and equally well might not be possible. It is possible (and worth further study) that the
model could achieve better performance if calibrated to the individual cities or possibly if calibrated to fit a group of cities with specific MFD characteristics. Moreover, in this work we use only one data source (LDs) but it is likely that some cities may have different sources at their disposal (e.g., Floating Car Data). Future work may focus on data fusion techniques to merge the various sources as representative input for the data-driven model that is then embedded in the MAML framework. Another approach would be to keep different data sources separate and use one (LDs) for meta-training and the other (e.g., Floating Car Data) for validation.   Finally it should also be noted that, in validating the proposed framework, FitFun was purposely not compared to MTPINN, as the models have different assumptions, as well as different training methods (a specific likelihood was used during training as described in \cite{bramich2023fitfun}. For example, it can be seen that the MSE of FitFun (Fig. \ref{fig:mse75lds_300_1000}) with and without MAML is far higher to the one of both MAML-NN and MAML-MTPINN (Table \ref{tab:Trasf50-75}). This, however, does not automatically mean that MAML-MTPINN is superior to FitFun, especially when a large number of detectors is available. Without MAML, FitFun shows an error comparable to MTPINN, suggesting that the model is robust and does not overfit the data, hence depicting a good fit but a higher MSE. More research must be done to compare the two models and to investigate how to better exploit MAML with FitFun.

\section{Conclusions}\label{sec:conclusions}

The paper investigates the problem of MFD estimation under data scarcity. More specifically, we consider the case in which only a few road segments in the network are equipped with loop detectors, which is likely to introduce bias during the estimation of the MFD. This paper has two main methodological contributions. 

First, we investigate the use of a new Meta Machine Learning Model (MAML) based on a model-agnostic Meta-Learning framework, in the context of MFD estimation. The framework is designed to estimate the MFD parameters by harnessing patterns learned from other cities to quickly adapt to new cities when limited data is available. The methodology is designed to tackle the problem of data scarcity in urban settings where the number of available LDs alone is not sufficient for a robust MFD estimation, which is in itself a topical but not trivial problem, as its theoretical formulation makes the estimation particularly susceptible to noise, time variance and bifurcation phenomena. 
The MAML design, tailored to address these challenges, is thoroughly described in its algorithmic process and modular structure.

Second, we introduce a Multi-Task Physics-Informed Neural Network (MTPINN) for non-parametric estimation of the MFD. The MTPINN is explicitly designed to take advantage of MAML as well as domain knowledge. The model can predict average flow, critical accumlation and critical occupancy. MTPINN is composed of shared layers and output branches. The shared layers serve two purposes. First, they learn latent features that are shared across the different outputs. Second, they allow for better generalisation, which is crucial to exploit multiple data-sets with MAML. In addition, the model is trained using a physics-informed loss that encourages the predictions to follow a traditional bi-parabolic shape. The resulting model exhibits high bias but low variance, as tends to be very robust compared to simple non parametric Neural Networks. 

We tested our models with the UTD19 dataset, an open dataset with traffic data from 39 cities, and shared the code for replicability. MTPINN is tested with and without the Meta-Learning component. Both methods (MAML and MTPINN) are benchmarked against a bi-parabolic model. Overall, results show that MAML-MTPINN is a robust framework for MFD estimation. However, while they do provide a better fit, a problem emerges with the uncongested branch. As this branch is not visible in many cities, MTPINN - as well as other non parametric models - seems to often fail in capturing the congested branch. The conventional bi-parabolic is more stable when only a few data points are available for the congestion branch.

The MAML model learns the general MFD properties and is able to exploit them, to better generalize. On average, the proposed MAML boasts sensibly smaller errors in any tested configuration (e.g., mean MSE ranging between 1768 and 3082 versus values of 6140 to 11491 for the MTPINN). The lower values and variance of the performance indicators suggests that the Meta-Learning component not only improves the estimation of the MFD, but makes it more robust against further drops in available LDs (i.e., the reduce variance).

Finally, MAML and MTPINN were validated with other methods from the literature. MAML was tested on FitFun, another non-parametric model presented in the literature, which has a different functional form compared to MTPINN, as well as different assumptions on the error and a different loss function. MAML was successfully combined with this model, providing either better results or, at worst, faster performances. Finally, we compared MAML and MTPINN with two alternative approaches. On the one hand, a simple Neural Network with a comparable architecture. On the other hand, Transfer Learning, a training method that, similarly to MAML, can exploit different data sets to achieve better generalisation, especially when there is an existing model that needs to be adapted to a smaller data set. The combination of MAML and MTPINN outperforms all benchmarks models in the case of very limited detectors. In the case in which more detectors are available, Transfer Learning combined with MTPINN provides better results. More notably, MTPINN always performs as expected, using both Transfer Learning and MAML. On the contrary, Neural Networks tend to overfit the data, especially in the case of limited detectors. MAML, however, systematically helps reducing overfitting, and in general, the simple Neural Network also exhibits predictable behaviour with MAML. On the other hand, this is not the case when using Transfer Learning. While in some cases Transfer Learning outperforms MAML, when training a simple Neural Network, in other cases it actually provides worse results compared to simple training. 
On the other hand, the proposed MTPINN, due to its high bias, was also found suitable for Transfer Learning, especially if the dataset is not small. 
The presented work's contribution formalises a Meta-Learning approach and leverages it to improve estimations with fewer data. The described methodology improves the MFD estimation in conditions of data scarcity when compared to similar approaches without the meta-learning component. With 75 LDs, the average MSE drops from 6139 (without MAML) to 1768 vehicles per time interval (with MAML). The average MAE across cities similarly drops from 60 to 30 vehicles per time interval. With 10 LDs, these values become respectively 11491 and 3082 vehicles per interval for the MSE, 86 and 41 for the MAE.
As described in Section~\ref{sec:results} for the Essen example, our work already allows to exploit the developed MTPINN with the MAML initialized weights to estimated MFDs in cities with fewer detectors. This means that cities that have data similar to what has been used in our experiments can already exploit the developed method and/or results (weights) to either improve their MFD estimation (in the case of medium-sized cities) or make up for temporary drops in operating detectors (in the case of planned maintenance, for example). Our work opens multiple research directions, such as enhanced multi-region MFD estimation with transferability across regions and enhanced estimation of the uncongested branch suffering from data scarcity problems. The presented work will allow us to focus on estimating MFDs that can be used to recreate realistic simulation outputs. 

\section{Supplementary materials}
The code for the biparabolic whitebox model, the MTPINN and the MAML framework is available at https://github.com/s184227/MAML\_for\_MFD/.
Extended results and additional plots can also be found in the repository

\section{Acknowledgements}
The presented research is supported by Novo Nordisk Foundation grant NNF23OC0085356.

\bibliography{references}
\newpage

\section{Appendix}\label{completeresults}
\renewcommand{\thefigure}{A\arabic{figure}}
\setcounter{figure}{0}
In this section, we report additional material and results. For the full list of results, please refer to https://github.com/s184227/MAML\_for\_MFD/
\subsection{Outlier example for the bi-parabolic hybrid model}
\begin{figure*}[h]
    \centering
    \includegraphics[width=0.94\linewidth]{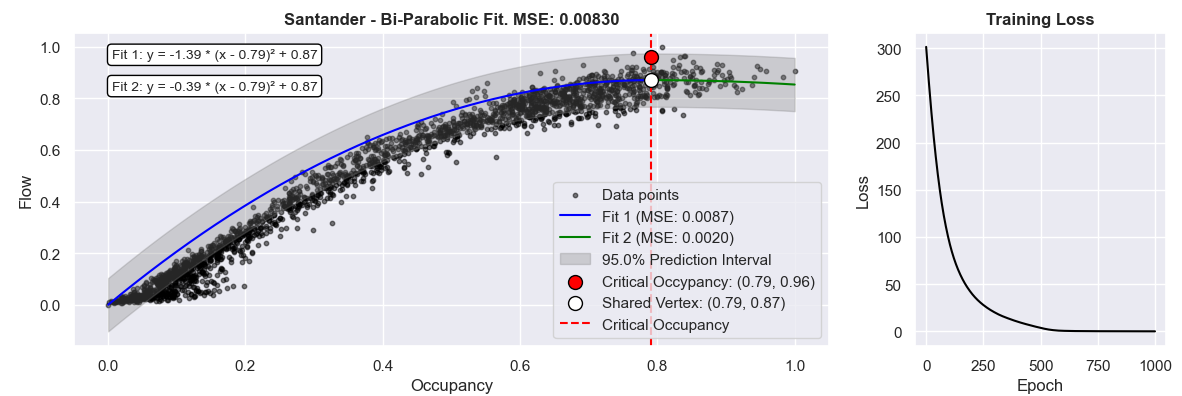}
    \caption{Results of the bi-parabolic hybrid model for Santander - The estimated MFD overestimates the flow in the first half of the fit, a behavior.}
    \label{fig:biperformance_santanders}
\end{figure*}

\subsection{MTPINN for Speyer and Santander}

\begin{figure*}[h]
    \centering
    \includegraphics[width=1.\linewidth]{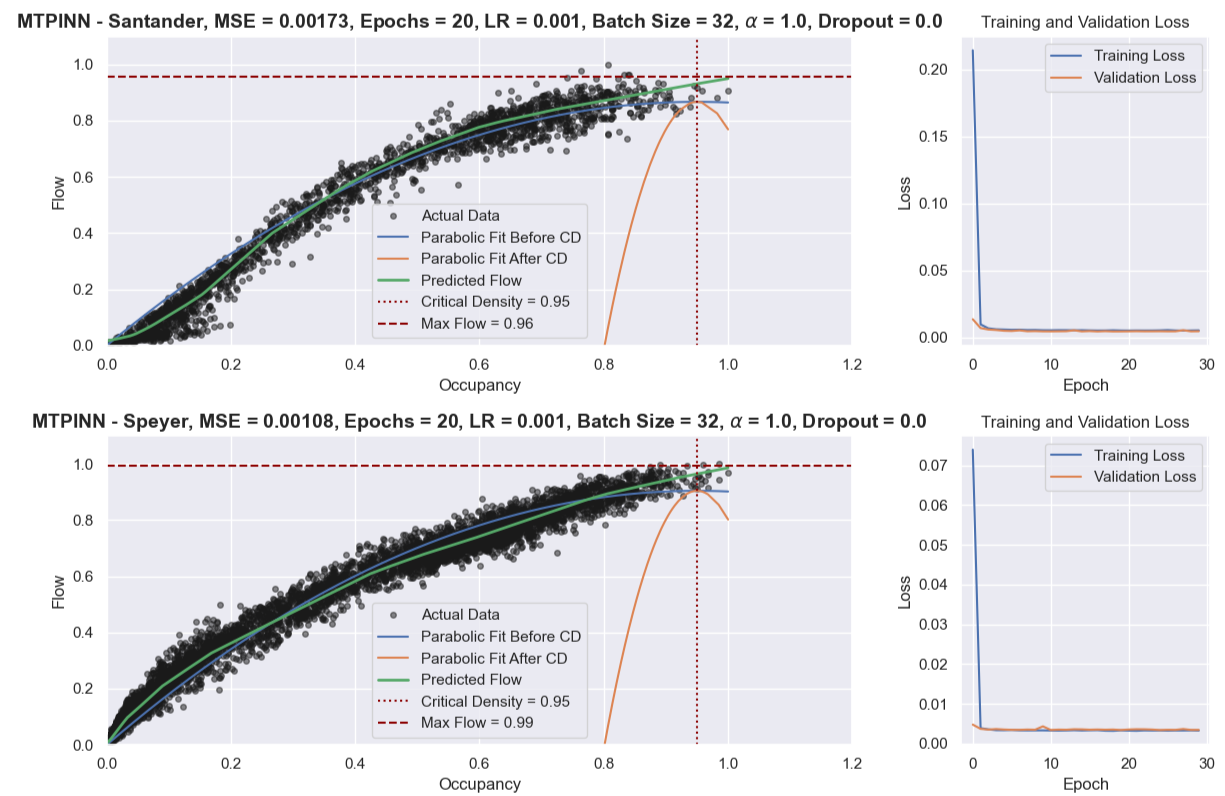}
    \caption{Results of the MTPINN model for two different cities - Santander and Speyer, - with differently shaped MFD diagrams and varying amounts of available observations.}
    \label{fig:MTPINNspeyerstdrs}
\end{figure*}
\newpage
\subsection{10 LDs MAML: Loss progression for inner- and outer-loop iterations}
\begin{figure*}[h]
    \centerline{\includegraphics[width=0.78\linewidth]{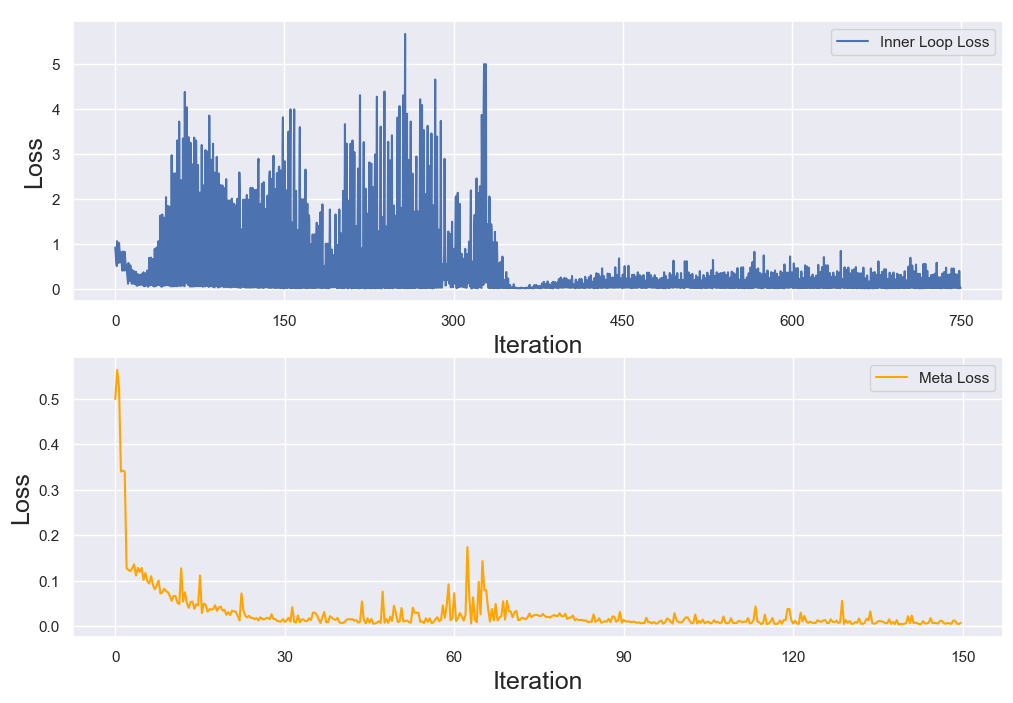}}
    \caption{Meta-losses for the inner and outer loop during meta-training  using datasets consisting of averages based on observations from 10 randomly selected detectors in a city. }
    \label{A3}
\end{figure*}

\subsection{MAML performance with 10 LDs per city}
\begin{figure*}[h]
    \centering
    \includegraphics[width=0.8\linewidth]{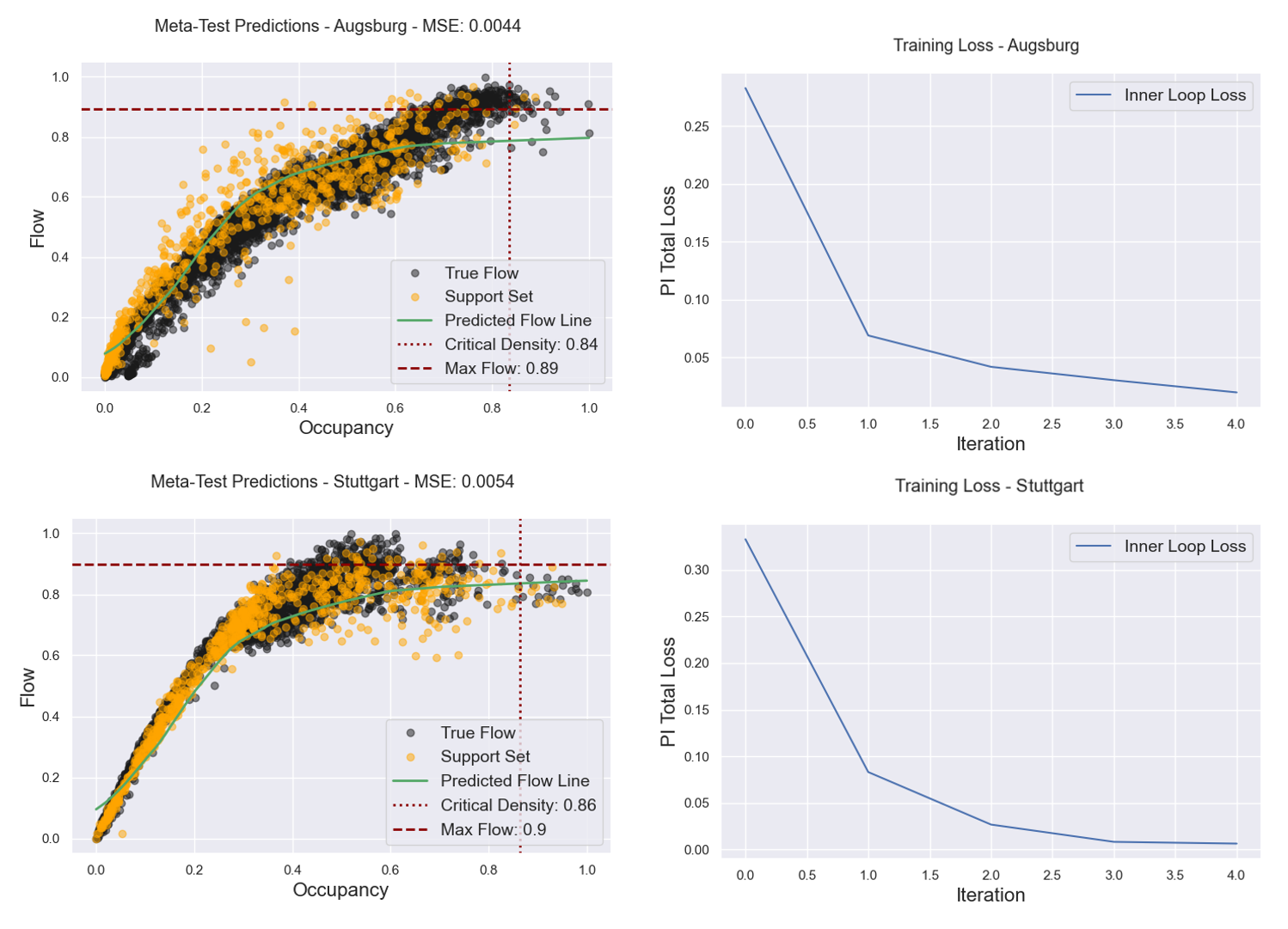}
    \caption{Meta-test sample of a meta-model trained on 10 detector datasets.}
    \label{fig:MAMLotherperf}
\end{figure*}
\pagebreak
\newpage

\subsection{MAE on the predicted flows per city, averaged across experiments}

\begin{table}[h]
\caption{MAE values calculated on the flow predicted by MAML and MTPINN with the different sets of LDs}
\hspace*{-0.9cm}
\begin{tabular}{|l|l|l|l|l|l|l|l|l|}
\hline
\textbf{City} & \footnotesize\textbf{\begin{tabular}[c]{@{}l@{}}MAML\\ 75\end{tabular}} & \footnotesize\textbf{\begin{tabular}[c]{@{}l@{}}MTPINN \\ 75\end{tabular}} & \footnotesize\textbf{\begin{tabular}[c]{@{}l@{}}MAML \\ 50\end{tabular}} & \footnotesize\textbf{\begin{tabular}[c]{@{}l@{}}MTPINN \\ 50\end{tabular}} & \footnotesize\textbf{\begin{tabular}[c]{@{}l@{}}MAML \\ 25\end{tabular}} & \footnotesize\textbf{\begin{tabular}[c]{@{}l@{}}MTPINN \\ 25\end{tabular}} & \footnotesize\textbf{\begin{tabular}[c]{@{}l@{}}MAML \\ 10\end{tabular}} & \footnotesize\textbf{\begin{tabular}[c]{@{}l@{}}MTPINN \\ 10\end{tabular}} \\ \hline
Augsburg      & 30.92                                                      & 44.26                                                         & 31.48                                                       & 48.58                                                         & 29.82                                                       & 56.27                                                         & 34.25                                                       & 49.79                                                         \\ \hline
Bern          & 32.46                                                      & 24.66                                                         & 30.46                                                       & 32.12                                                         & 31.74                                                       & 36.78                                                         & 26.06                                                       & 63.67                                                         \\ \hline
Bordeaux      & 26.18                                                      & 72.27                                                         & 27.83                                                       & 73.69                                                         & 28.09                                                       & 56.50                                                         & 30.43                                                       & 101.99                                                        \\ \hline
Bremen        & 28.55                                                      & 26.72                                                         & 31.48                                                       & 25.83                                                         & 27.97                                                       & 40.94                                                         & 26.92                                                       & 41.14                                                         \\ \hline
Darmstadt     & 27.04                                                      & 41.27                                                         & 30.21                                                       & 34.99                                                         & 49.68                                                       & 46.60                                                         & 31.83                                                       & 88.12                                                         \\ \hline
Graz          & 14.16                                                      & 45.81                                                         & 17.22                                                       & 32.76                                                         & 20.84                                                       & 48.39                                                         & 18.04                                                       & 40.58                                                         \\ \hline
Hamburg       & 19.73                                                      & 48.61                                                         & 20.57                                                       & 49.67                                                         & 28.99                                                       & 69.82                                                         & 29.27                                                       & 74.52                                                         \\ \hline
Kassel        & 21.39                                                      & 28.09                                                         & 21.81                                                       & 32.16                                                         & 23.64                                                       & 36.13                                                         & 30.97                                                       & 34.08                                                         \\ \hline
London        & 15.97                                                      & 52.53                                                         & 14.57                                                       & 51.62                                                         & 20.14                                                       & 55.09                                                         & 38.69                                                       & 68.78                                                         \\ \hline
Losangeles    & 71.87                                                      & 137.14                                                        & 82.66                                                       & 134.72                                                        & 80.16                                                       & 117.16                                                        & 68.58                                                       & 157.72                                                        \\ \hline
Madrid        & 33.37                                                      & 112.98                                                        & 28.64                                                       & 89.36                                                         & 35.59                                                       & 90.82                                                         & 94.71                                                       & 137.74                                                        \\ \hline
Marseille     & 31.40                                                      & 82.36                                                         & 43.88                                                       & 102.98                                                        & 38.91                                                       & 101.79                                                        & 56.46                                                       & 150.72                                                        \\ \hline
Santander     & 44.35                                                      & 59.02                                                         & 43.18                                                       & 84.41                                                         & 50.13                                                       & 57.66                                                         & 70.05                                                       & 99.96                                                         \\ \hline
Speyer        & 19.08                                                      & 33.49                                                         & 19.82                                                       & 31.11                                                         & 21.30                                                       & 27.47                                                         & 21.21                                                       & 46.77                                                         \\ \hline
Strasbourg    & 42.22                                                      & 77.32                                                         & 39.96                                                       & 72.05                                                         & 43.11                                                       & 105.03                                                        & 56.42                                                       & 103.88                                                        \\ \hline
Stuttgart     & 34.53                                                      & 71.54                                                         & 34.21                                                       & 78.85                                                         & 53.41                                                       & 82.97                                                         & 45.42                                                       & 95.33                                                         \\ \hline
Taipeh        & 32.97                                                      & 73.56                                                         & 36.34                                                       & 78.46                                                         & 39.57                                                       & 68.45                                                         & 40.19                                                       & 85.25                                                         \\ \hline
Toronto       & 38.23                                                      & 63.12                                                         & 39.93                                                       & 75.54                                                         & 43.96                                                       & 77.40                                                         & 42.78                                                       & 110.94                                                        \\ \hline
Toulouse      & 31.97                                                      & 83.29                                                         & 32.29                                                       & 99.20                                                         & 44.14                                                       & 88.27                                                         & 43.84                                                       & 105.35                                                        \\ \hline
Zurich        & 10.41                                                      & 28.75                                                         & 11.67                                                       & 35.05                                                         & 14.19                                                       & 39.91                                                         & 13.99                                                       & 58.55                                                         \\ \hline
\end{tabular} \label{tab:maepercity}
\end{table}

\end{document}